\documentclass[twoside,11pt]{article}

\usepackage{jmlr2e}
\usepackage{amsmath,amssymb}
\usepackage{graphicx,verbatim}
\usepackage{subfigure}
\usepackage[ruled,vlined]{algorithm2e}
\usepackage{pslatex}

\usepackage{fancybox}
\usepackage{tikz}
\usepackage{xcolor} 
\usepackage{color}
\usepackage{relsize}
\usepackage{url}

\def\trace {{\rm trace}}

\newcommand{\Minimize}{\displaystyle \operatornamewithlimits{minimize}}
\newcommand{\Maximize}{\displaystyle \operatornamewithlimits{maximize}}
\newcommand{\rcon}{row-column overlap norm}

\newcommand{\npjgl}{perturbed-node joint graphical lasso}
\newcommand{\cnjgl}{\coactive node joint graphical lasso}

\newcommand{\coactive}{cohub }

\newcommand{\RCON}{RCON }
\newcommand{\RCONnospace}{RCON}

\newcommand{\NPJGL}{PNJGL }
\newcommand{\PNJGL}{PNJGL }

\newcommand{\PNJGLnospace}{PNJGL}
\newcommand{\CNJGL}{CNJGL }
\newcommand{\CNJGLnospace}{CNJGL}

\newcommand{\auglag}{\mathcal{L}_{\rho}} 
\newcommand{\Vonec}{\tilde{V}^1}
\newcommand{\Vtwoc}{\tilde{V}^2}

\newcommand{\Vic}{\tilde{V}^i}
\newcommand{\VKc}{\tilde{V}^K}

\usepackage[colorlinks=false,allbordercolors={1 1 1}]{hyperref}

\newcommand{\beq}{\begin{eqnarray}}
\newcommand{\eeq}{\end{eqnarray}}
\newcommand{\beqs}{\begin{eqnarray*}}
\newcommand{\eeqs}{\end{eqnarray*}}
\newcommand{\barr}{\begin{array}}
\newcommand{\earr}{\end{array}}
\newcommand{\beqa}{\begin{eqnarray} \begin{array}}
\newcommand{\eeqa}{\end{array} \end{eqnarray}}
\newcommand{\beqas}{\begin{eqnarray*} \begin{array}}
\newcommand{\eeqas}{\end{array} \end{eqnarray*}}

\newcommand{\meq}{&=&}

\newcommand{\bthm}{\begin{theo}}
\newcommand{\ethm}{\end{theo}}
\newcommand{\blem}{\begin{lem}}
\newcommand{\elem}{\end{lem}}
\newcommand{\bdefn}{\begin{defn}}
\newcommand{\edefn}{\end{defn}}

\newcommand{\SUM}{\displaystyle \sum}
\newcommand{\MAX}{\displaystyle \max}
\newcommand{\MIN}{\displaystyle \min}

\newcommand{\mnorm}[1]{{\|#1 \|}}
\newcommand{\mnormfull}[1]{{\left\|#1 \right\|}}
\newcommand{\matrixnorm}{{\|.\|}}
\newcommand{\RCONmatrix}{\Omega}

\newcommand{\PD}{\mathcal{S}_{++}^p}

\newcommand{\tA}{\boldsymbol{A}}

\newcommand{\tD}{\boldsymbol{D}}
\newcommand{\tF}{\boldsymbol{F}}
\newcommand{\tG}{\boldsymbol{G}}
\newcommand{\tI}{\boldsymbol{I}}
\newcommand{\tQ}{\boldsymbol{Q}}
\newcommand{\tS}{\boldsymbol{S}}
\newcommand{\tT}{\boldsymbol{T}}
\newcommand{\tU}{\boldsymbol{U}}
\newcommand{\tV}{\boldsymbol{V}}
\newcommand{\tW}{\boldsymbol{W}}
\newcommand{\tX}{\boldsymbol{X}}

\newcommand{\tZ}{\boldsymbol{Z}}

\newcommand{\R}{\mathbb{R}}

\newcommand{\Xcal}{\mathcal{X}}

\newcommand{\diag}{\mbox{diag}}

\newcommand{\bTheta}{\boldsymbol{\Theta}}
\newcommand{\bThetaone}{{\boldsymbol{\Theta}}^1}
\newcommand{\bThetatwo}{{\boldsymbol{\Theta}}^2}
\newcommand{\bThetai}{{\boldsymbol{\Theta}}^i}

\newcommand{\bThetak}{{\boldsymbol{\Theta}}^k}
\newcommand{\bThetaK}{{\boldsymbol{\Theta}}^K}
\newcommand{\var}{\boldsymbol{\Theta}}
\newcommand{\varone}{{\boldsymbol{\Theta}}^1}
\newcommand{\vartwo}{\boldsymbol{\Theta}^2}
\newcommand{\varK}{{\boldsymbol{\Theta}}^K}
\newcommand{\vari}{{\boldsymbol{\Theta}}^i}

\newcommand{\vark}{{\boldsymbol{\Theta}}^k}
\newcommand{\hatvarone}{{\hat{\boldsymbol{\Theta}}}^1}
\newcommand{\hatvartwo}{\hat{\boldsymbol{\Theta}}^2}
\newcommand{\Lambdaone}{\Lambda^1}

\newcommand{\Lambdai}{\Lambda^i}
\newcommand{\Lambdak}{\Lambda^k}
\newcommand{\LambdaK}{\Lambda^K}
\newcommand{\bSigma}{\boldsymbol{\Sigma}}

\newcommand{\Cone}{{\boldsymbol{C}}^1}
\newcommand{\Ctwo}{{\boldsymbol{C}}^2}
\newcommand{\Ci}{{\boldsymbol{C}}^i}
\newcommand{\CK}{{\boldsymbol{C}}^K}

\newcommand{\Fi}{{\boldsymbol{F}}^i}

\newcommand{\Gi}{{\boldsymbol{G}}^i}

\newcommand{\Qone}{{\boldsymbol{Q}}^1}
\newcommand{\Qtwo}{{\boldsymbol{Q}}^2}
\newcommand{\Qi}{\boldsymbol{Q}^i}

\newcommand{\Sii}{{\boldsymbol{S}}^i}
\newcommand{\Vone}{{\boldsymbol{V}}^1}
\newcommand{\Vtwo}{{\boldsymbol{V}}^2}

\newcommand{\Vi}{{\boldsymbol{V}}^i}
\newcommand{\Vk}{{\boldsymbol{V}}^k}
\newcommand{\VK}{{\boldsymbol{V}}^K}

\newcommand{\Wi}{{\boldsymbol{W}}^i}

\newcommand{\Zone}{{\boldsymbol{Z}}^1}
\newcommand{\Ztwo}{{\boldsymbol{Z}}^2}
\newcommand{\Zi}{{\boldsymbol{Z}}^i}
\newcommand{\Zk}{{\boldsymbol{Z}}^k}
\newcommand{\ZK}{{\boldsymbol{Z}}^K}

\newcommand{\expand}{\mbox{Expand}}
\newcommand{\ADMM}{\mbox{ADMM }}

\newcommand{\argmini}{\displaystyle \mathop{\rm argmin}}

\newcommand{\thresh}{\mathcal{T}}

\newcommand{\Vard}{{\Theta}_{\bigtriangleup} }

\newcommand{\psd}{\mathbb{S}_{++}^{p}}
\newcommand{\regone}{\lambda_1}
\newcommand{\regtwo}{\lambda_2}

\newcommand{\sgn}{\operatorname{sgn}}

\definecolor{color1}{rgb}{0.11764705882,0.31372549019,1}
\definecolor{color2}{rgb}{0.25490196078,0.41176470588,0.882353}
\definecolor{color3}{rgb}{0,0,1}
\definecolor{color4}{rgb}{0.11764705882,0.56470588235,1}
\definecolor{color5}{rgb}{1,0.49803921568,0.14117647058}
\definecolor{color6}{rgb}{1,0.07843137254,0.57647058823}
\definecolor{color7}{rgb}{1,0,0} 
\definecolor{color8}{rgb}{0.64705882352,0.16470588235,0.16470588235}
\definecolor{color9}{rgb}{0, 0.882353, 0}
\definecolor{color10}{rgb} {1,0.756862745,0.145098039}
\definecolor{color11}{rgb} {0.784313, 0.58823, 0.19608}
\definecolor{color12}{rgb} {0.80392, 0.39216, 0.39216}
\definecolor{color13}{rgb} {0.784313, 0, 0.980392}

\jmlrheading{15}{2014}{1-44}{3/13; Revised 8/13}{1/14}{Karthik Mohan, Palma London, Maryam Fazel,  Daniela Witten and Su-In Lee}
\ShortHeadings{Node-based learning of multiple GGMs}{Mohan, London, Fazel, Witten and Lee}
\firstpageno{1}
\title{Node-Based Learning of Multiple Gaussian Graphical Models}

\begin{document}
\author{\name Karthik Mohan \email karna@uw.edu \\
           \name Palma London \email palondon@uw.edu \\
         \name Maryam Fazel \email mfazel@uw.edu \\
         \addr Department of  Electrical Engineering \\
               University of Washington \\
               Seattle WA, 98195 \AND
            \name Daniela Witten \email dwitten@uw.edu \\
         \addr Department of Biostatistics  \\
               University of Washington \\
               Seattle WA, 98195 \AND
         \name Su-In Lee \email suinlee@cs.washington.edu \\
         \addr Departments of  Computer Science and Engineering, Genome Sciences \\
         \addr University of Washington \\
               Seattle WA, 98195
         }
\editor{Saharon Rosset}
\maketitle

\begin{abstract}%
We consider the problem of estimating high-dimensional Gaussian
graphical models corresponding to a single set of variables under
several distinct conditions. This problem is motivated by the task of recovering transcriptional regulatory networks on the basis of gene expression data {containing heterogeneous samples, such as different disease states, multiple species, or different developmental stages}.
We assume that most aspects of the conditional dependence
networks are shared, but that there are some structured differences
between them. Rather than assuming that similarities and differences between networks are driven by
individual edges, we take a \emph{node-based} approach, which in many cases provides a more intuitive interpretation of the network differences. We consider estimation under two distinct assumptions: (1)  differences between the $K$ networks are due to individual nodes that are \emph{perturbed} across conditions, or  (2)  similarities among the $K$ networks are due to the presence of \emph{common hub nodes} that are shared across all $K$ networks.
Using a \emph{row-column overlap norm} penalty function, we formulate two convex optimization problems  that correspond to these two assumptions.
We solve these problems using  an alternating direction method of multipliers algorithm, and we derive a set of necessary and sufficient conditions that allows us to decompose the problem into independent subproblems so that our algorithm can be scaled to high-dimensional settings. Our proposal is illustrated on synthetic data, a webpage data set, and  a brain cancer gene expression data set.
\end{abstract}


\section{Introduction}
\label{sec:intro}

\emph{Graphical models} encode the conditional dependence relationships among a set of $p$ variables, or features \citep{Lauritzen1996}. They are a tool of growing importance in a number of fields, including finance, biology, and computer vision.
A graphical model is often referred to as a  conditional dependence \emph{network}, or simply as a \emph{network}. Motivated by network terminology, we can refer to the $p$ variables in a graphical model as \emph{nodes}. If a pair of variables (or features) are conditionally dependent, then there is an \emph{edge} between the corresponding pair of nodes; otherwise, no edge is present. 

Suppose that we have $n$ observations that are independently drawn from a   {multivariate normal} distribution with covariance matrix ${\bSigma}$.
Then the  corresponding \emph{Gaussian graphical model} (GGM) that describes the conditional dependence relationships among the features
is encoded by the sparsity pattern of the inverse covariance matrix, ${\bSigma}^{-1}$ \citep[see, e.g.,][]{MKB79,Lauritzen1996}.
That is, the $j$th and $j'$th variables are conditionally independent if and only if $({\bSigma}^{-1})_{jj'}=0$.
Unfortunately, when $p > n$,  obtaining an accurate estimate of ${\bSigma}^{-1}$ is challenging.
In such a scenario, we can use prior information---such as the knowledge that many of the pairs of variables are conditionally independent---in order to more accurately estimate ${\bSigma}^{-1}$ \citep[see, e.g.,][]{YuanLin07,SparseInv,Banerjee}.

In this paper, we consider the task of estimating $K$ GGMs on a single set of $p$ variables under the assumption that the GGMs are similar, with certain structured differences.
As a motivating example, suppose that we have access to gene expression
measurements for $n_1$ lung cancer samples and $n_2$ {normal} lung  samples, and that we would like to estimate the gene regulatory networks underlying the normal and cancer lung tissue. We can model each of these regulatory networks using a GGM. We have two obvious options. 
\begin{enumerate}
\item We can estimate a single network on the basis of all $n_1+n_2$ tissue samples. But
this approach overlooks fundamental differences between the true lung cancer and normal gene regulatory networks.
\item  We can estimate separate networks based on the $n_1$ cancer and $n_2$ normal samples. However, this approach fails to exploit substantial commonality of the two networks, such as lung-specific pathways.
\end{enumerate}
In order to effectively make use of the available data, we need a principled approach for jointly estimating the two networks in such a way that the two estimates are encouraged to be quite similar to each other, while allowing for certain structured differences.  In fact, these differences may be of scientific interest.

Another example of estimating multiple GGMs  arises in the analysis of the conditional dependence relationships among $p$ stocks at two distinct points in time.  We might be
interested in detecting stocks that have differential connectivity with
all other stocks across the two time points, as these likely correspond to companies that have undergone significant changes. Yet another example occurs in the field of
neuroscience, in which it is of interest to learn how the connectivity
of neurons changes over time.

Past work on joint estimation of multiple GGMs has assumed that  individual \emph{edges}  are shared or differ across conditions \citep[see, e.g.,][]{KolarXing2010,Zhang-10,Guo2011,Danaher2012}; here we refer to such approaches as  \emph{edge-based}. In this paper, we instead take a \emph{node-based} approach:
 we seek to estimate $K$ GGMs under the assumption that similarities and differences between networks are driven by individual \emph{nodes} whose patterns of connectivity to other nodes are shared across networks, or differ between networks. As we will see, node-based learning is more powerful than edge-based learning,  since it more fully exploits our prior assumptions about the similarities and differences between networks.

\begin{figure}[h]
\begin{center}
\subfigure[]{\includegraphics[width=0.25\linewidth,clip]{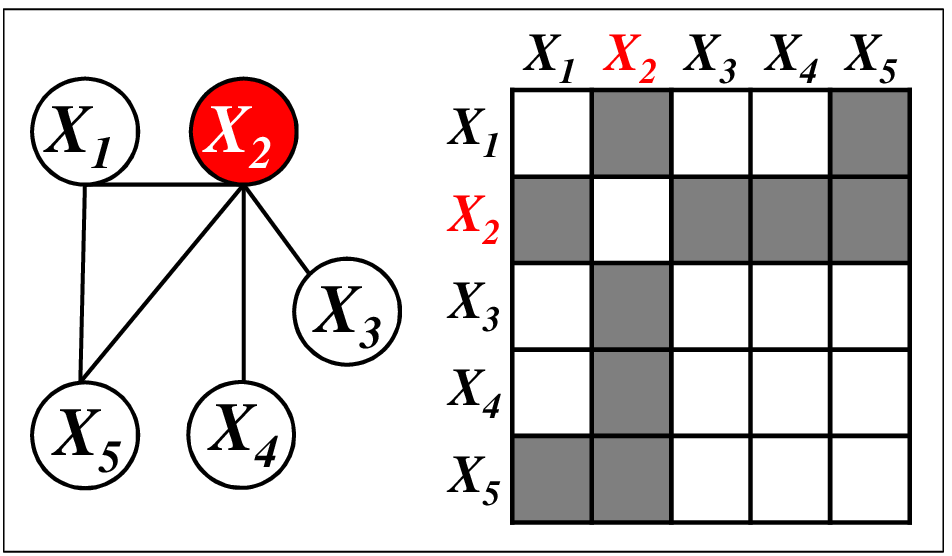}} \qquad
\subfigure[]{\includegraphics[width=0.25\linewidth,clip]{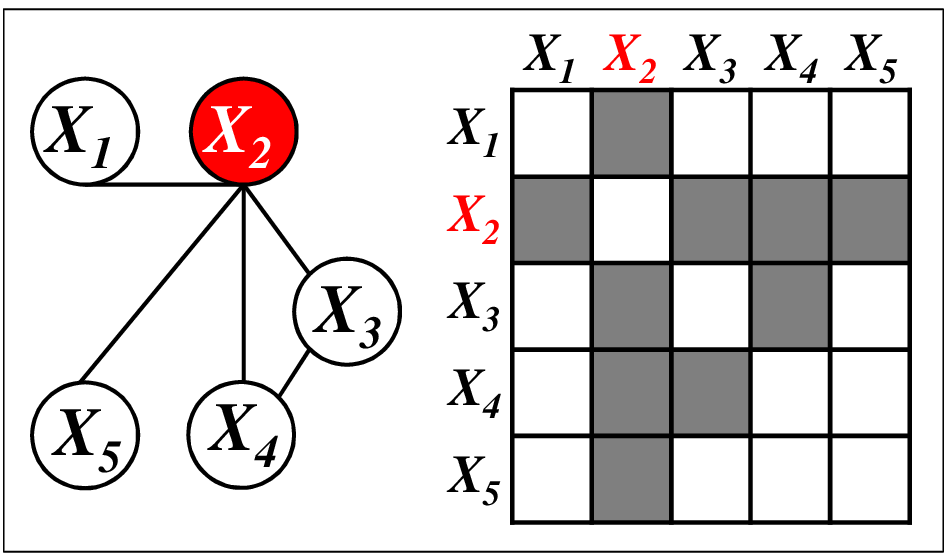}}
\end{center}
\vspace{-8mm}
\caption{Two networks  share a \emph{common hub} (co-hub) node. $X_2$ serves as a hub node in both networks. \emph{(a):} Network 1 and its adjacency matrix. \emph{(b):} Network 2 and its adjacency  matrix.  \label{fig:node-cohub}}
\vspace{-2mm}
\end{figure}

\begin{figure}[h]
\subfigure[]{\includegraphics[width=0.295\linewidth,clip]{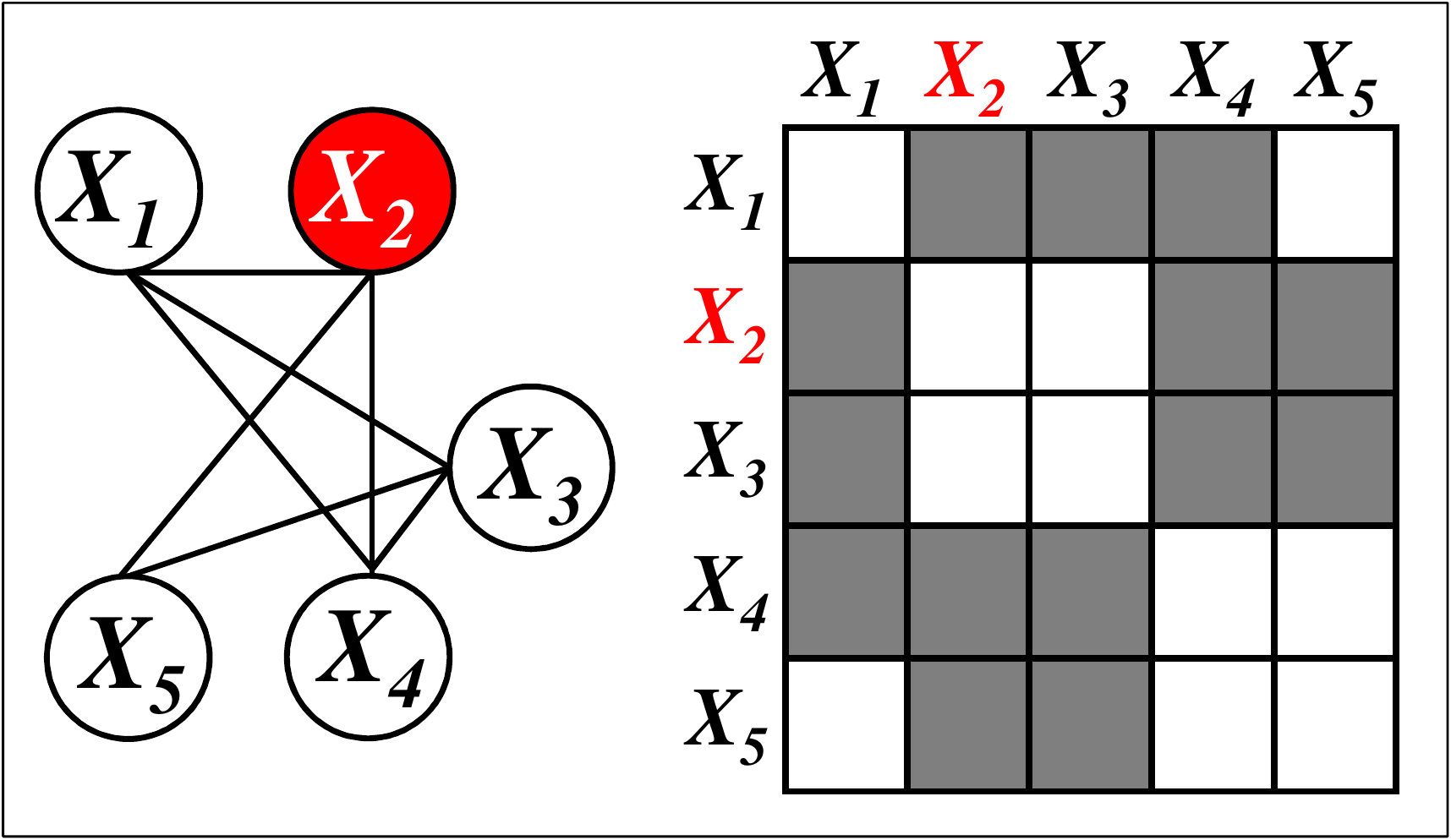}} \qquad
\subfigure[]{\includegraphics[width=0.295\linewidth,clip]{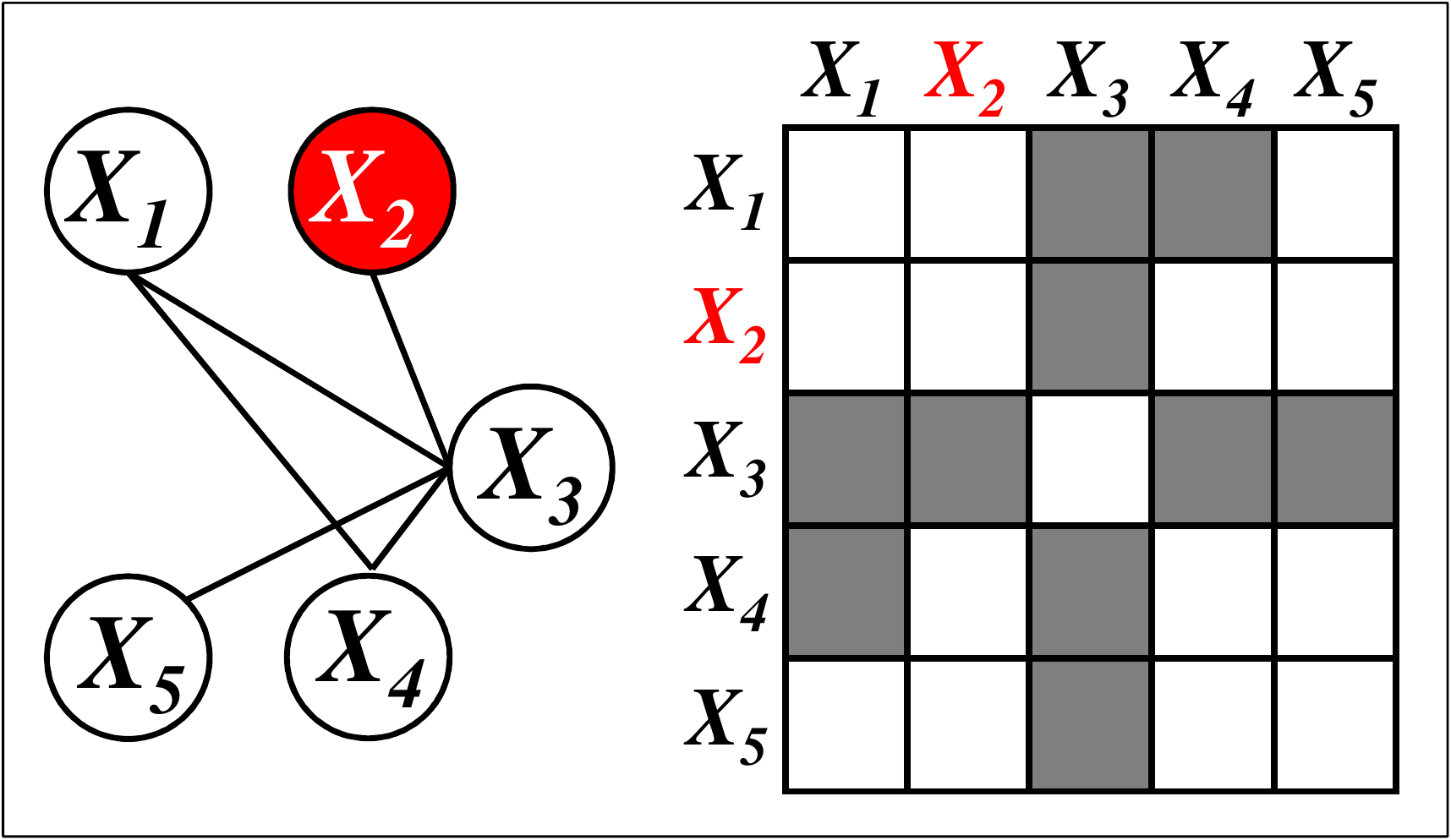}} \qquad
\subfigure[]{\includegraphics[width=0.295\linewidth,clip]{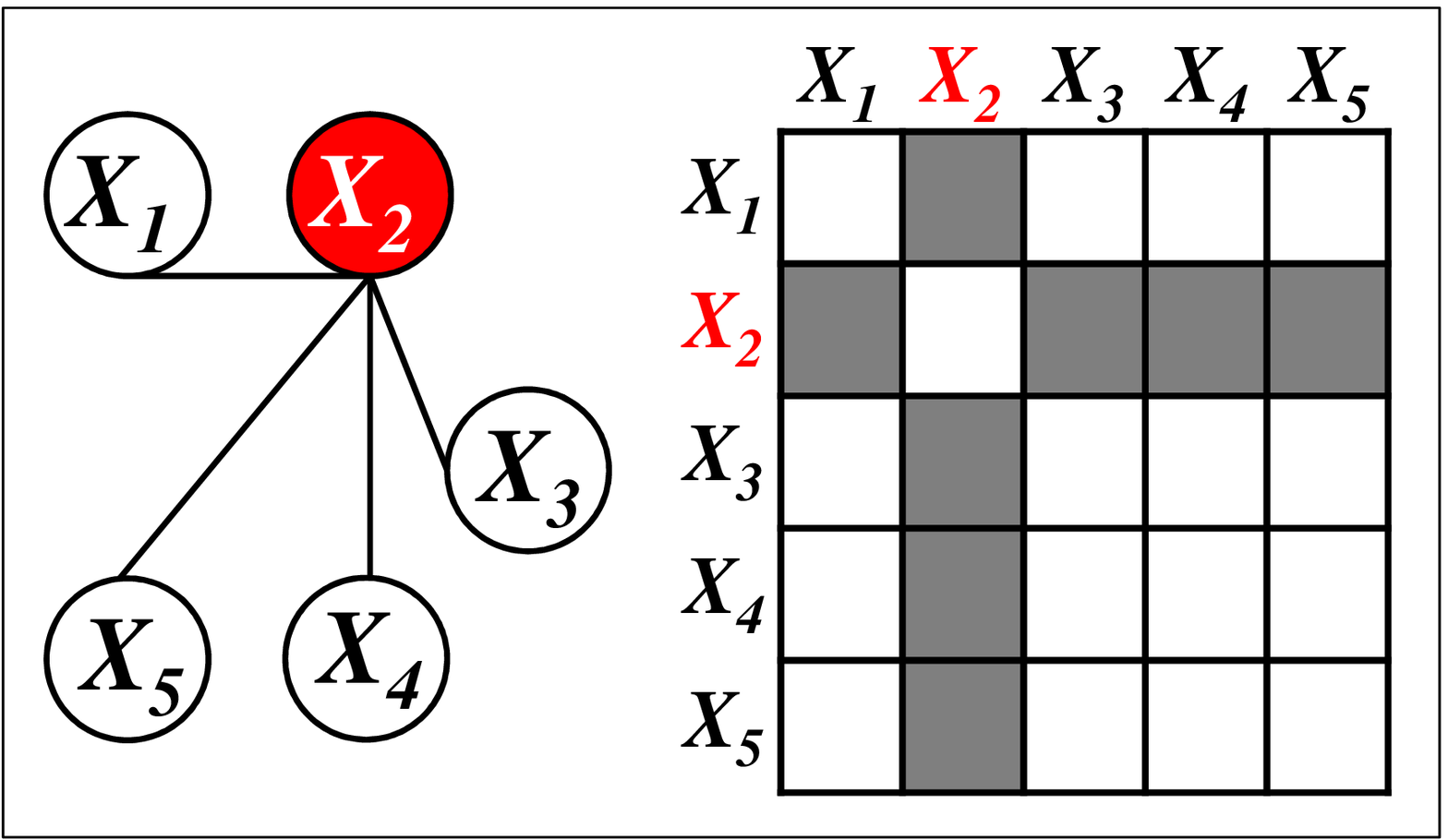}}
\vspace{-8mm}
\caption{Two networks that differ due
to \emph{node perturbation} of $X_2$. \emph{(a):} Network 1 and its
adjacency matrix. \emph{(b):} Network 2 and its adjacency matrix. \emph{(c):}
\emph{Left:} Edges that differ between the two networks.
\emph{Right:}
Shaded cells indicate edges that differ between Networks 1 and 2. \label{fig:node-pert}}
\vspace{-2mm}
\end{figure}

  More specifically, in this paper we consider two types of shared network structure.
  \begin{enumerate}
  \item Certain nodes serve as highly-connected \emph{hub} nodes. We assume that the same nodes serve as hubs in each of the $K$ networks. 
  Figure \ref{fig:node-cohub} illustrates a toy example of this setting, with $p=5$ nodes and $K=2$ networks.  In this example, the second variable, $X_2$, serves as a hub node in each network.
  In the context of transcriptional regulatory networks, $X_2$ might represent a gene that encodes a \emph{transcription factor} that regulates a large number of downstream genes {in all $K$ contexts}.
  We propose the \emph{common hub (co-hub) node  joint graphical lasso} (\CNJGLnospace), a convex optimization problem for estimating GGMs in this setting.
  \item The  networks differ due to particular nodes that are \emph{perturbed} across conditions, and therefore have a completely different connectivity pattern to other nodes in the $K$ networks.
  Figure~\ref{fig:node-pert} displays a toy example, with $p=5$ nodes and $K=2$ networks; here we see that all of the network differences are driven by perturbation in the second variable, $X_2$.
  In the context of transcriptional regulatory networks, $X_2$ might represent a gene that is mutated in a particular condition, effectively disrupting its conditional dependence relationships with other genes.
  We propose the \emph{perturbed-node joint graphical lasso} (\PNJGLnospace), a convex optimization problem for estimating GGMs in this context.
  \end{enumerate}
  Node-based learning of multiple GGMs is challenging, due to  complications resulting from symmetry of the precision matrices. In this paper, we overcome this problem through the use of a new convex regularizer. 

The rest of this paper is organized as follows. We introduce some relevant background material in Section 2. In Section 3,
we present the  \emph{row-column overlap norm} (\RCONnospace), a regularizer that  encourages a matrix
  to have a support that is the \emph{union of a set of rows and columns}.
 We apply the \RCON penalty to  a pair of inverse covariance matrices, or to the difference between a pair of inverse covariance matrices, in order to obtain the \CNJGL and \PNJGL formulations just described.
In Section 4, we propose an \emph{alternating direction method of multipliers} (ADMM) algorithm in order to solve these two convex formulations.
In order to scale this algorithm to problems with many features, in Section 5 we
 introduce a set of simple conditions on the regularization parameters that indicate that the problem can be broken down into many independent subproblems,  leading to substantial algorithm speed-ups.
In Section 6, we apply   \CNJGL and \PNJGL to synthetic data, and in Section 7 we apply them to gene expression data and to webpage data. The Discussion is in Section 8. Proofs are in the Appendix.

A preliminary version of some of the ideas in this paper appear in \cite{Mohan-12}. There the \PNJGL formulation was proposed, along with an ADMM algorithm. Here we expand upon that formulation and present the \CNJGL formulation, an ADMM algorithm for solving it, as well as comprehensive results on both real and simulated data. Furthermore, in this paper we discuss  theoretical conditions for computational speed-ups, which  are critical to application of both \PNJGL and \CNJGL to data sets with many {variables}.

\section{Background on High-Dimensional GGM Estimation}
\label{sec:background}
 
In this section, we review the literature on learning Gaussian graphical models.

\subsection{The Graphical Lasso for Estimating a Single GGM} 
As was mentioned in Section~\ref{sec:intro},  estimating a single GGM on the basis of $n$ independent and identically distributed observations from a $N_p({\bf 0}, \bSigma)$ distribution amounts to learning the sparsity structure of ${\bSigma}^{-1}$ \citep{MKB79,Lauritzen1996}.
When $n>p$, one can estimate ${\bSigma}^{-1}$ by maximum
likelihood. But in high dimensions when $p$ is large relative to $n$, this  is not possible because the
empirical covariance matrix is singular. Consequently, a number of
authors
\citep[among others,][]{YuanLin07,SparseInv,Ravikumar08,Banerjee,Scheinberg10,Hsieh11}
 have considered maximizing the penalized log likelihood
\begin{equation}
\Maximize_{\bTheta \in \mathbb{S}_{++}^p} \left\{ \log \det \bTheta - \trace(\tS \bTheta) - \lambda \|\bTheta\|_1 \right\},
\label{eqn:glasso}
\end{equation}
where ${\tS}$ is the empirical covariance matrix, $\lambda$ is a nonnegative tuning parameter, $\mathbb{S}_{++}^p$
denotes the set of  positive definite matrices of size $p$, and
 {$\|\bTheta\|_1 = \sum_{i, j} |\Theta_{ij}|$.} 
The solution to  (\ref{eqn:glasso}) serves as an estimate of
${\bSigma}^{-1}$, and a zero element in the solution
corresponds to a pair of features that are estimated to be conditionally independent. Due to the  $\ell_1$ penalty \citep{Ti96} in
(\ref{eqn:glasso}), this estimate will be positive definite for any
$\lambda>0$, and sparse  when $\lambda$ is sufficiently large. We refer to (\ref{eqn:glasso}) as the
\emph{graphical lasso}. Problem (\ref{eqn:glasso}) is convex,
and efficient algorithms for solving it are available
\citep[among others,][]{SparseInv,Banerjee,Rothman08,Aspremont08,Scheinberg10,WittenFriedman11}.

\subsection{The Joint Graphical Lasso for Estimating Multiple GGMs}
\label{sec:jgl}
Several formulations have recently been proposed for extending the graphical lasso (\ref{eqn:glasso}) to the setting in which one has access to a number of observations
from $K$ distinct conditions, each with measurements on the same set of $p$ features. The goal is to estimate a graphical model for each condition under the assumption that the $K$ networks share certain characteristics but are allowed to differ in certain structured ways. \citet{Guo2011} take a non-convex approach to solving this problem. \citet{Zhang-10}  take a convex approach, but use a least squares loss function rather than the negative Gaussian log likelihood. Here we review the convex formulation of \cite{Danaher2012}, which forms the starting point for the proposal in this paper.

Suppose that
$X^k_1,\ldots,X^k_{n_k} \in \mathbb{R}^p$ are independent and identically distributed from a $N_p({\bf 0}, \bSigma^k)$ distribution, for $k=1,\ldots,K$. Here $n_k$ is the number of observations in the $k$th condition, or class. 
Letting ${\tS}^k$ denote the empirical covariance matrix for the $k$th class, we can maximize the penalized log likelihood
\begin{equation}
\Maximize_{\bThetaone \in \mathbb{S}_{++}^p,\ldots, \bThetaK  \in \mathbb{S}_{++}^p} \left\{ L(\bThetaone,\ldots, \bThetaK)  - \lambda_1 \sum_{k=1}^K \|\bThetak\|_1 - \lambda_2
\sum_{i \neq j} P(\bThetaone_{ij}, \ldots, \bThetaK_{ij}) \right\},
\label{eqn:fgl}
\end{equation}
where $L(\bThetaone,\ldots,\bThetaK)=\sum_{k=1}^K n_k \left( \log \det \bThetak - \trace({\tS}^k \bThetak)\right)$,
 $\lambda_1$ and $\lambda_2$ are nonnegative tuning parameters,
 and $P(\bThetaone_{ij}, \ldots, \bThetaK_{ij})$ is a convex penalty function applied to each off-diagonal element of ${\bTheta}^1,\ldots,{\bTheta}^K$
in order to encourage similarity among them.
Then  the $\hat{\bTheta}^1,\ldots,\hat{\bTheta}^K$ that solve (\ref{eqn:fgl}) serve as estimates for  $({\bSigma}^1)^{-1},\ldots,({\bSigma}^K)^{-1}$. \cite{Danaher2012} refer to (\ref{eqn:fgl})  as the \emph{joint graphical lasso} (JGL).
In particular, they consider the use of a \emph{fused lasso penalty} \citep{TSRZ2005},
\begin{equation}
P(\bThetaone_{ij}, \ldots, \bThetaK_{ij}) = \sum_{k<k'}  |{\bTheta}^k_{ij} - {\bTheta}^{k'}_{ij}|,
\label{eqn:fused}
\end{equation}
 on the  differences between pairs of network edges, as well as a \emph{group lasso penalty} \citep{grouplasso},

\begin{equation}
P(\bThetaone_{ij},\bThetatwo_{ij},\ldots,\bThetaK_{ij}) =  \sqrt{\sum_{k=1}^K (\bThetak_{ij})^2 },
\label{eqn:GGL_penalty}
\end{equation}
on the edges themselves.
\cite{Danaher2012} refer to problem (\ref{eqn:fgl}) combined with (\ref{eqn:fused}) as the \emph{fused graphical lasso} (FGL), and to  (\ref{eqn:fgl}) combined with (\ref{eqn:GGL_penalty}) as the \emph{group graphical lasso} (GGL). 

FGL  encourages the $K$ network estimates to have identical edge values, whereas GGL encourages the $K$ network estimates to have a shared pattern of sparsity. 
Both the FGL and GGL optimization problems are convex. An approach related to FGL and GGL is proposed in \citet{Hara-13}.

Because FGL and GGL borrow strength across all available observations in estimating each network, they can lead to much more accurate inference than simply learning each of the $K$ networks separately. 

But both  FGL and GGL take an \emph{edge-based} approach: they assume that differences between and similarities among the networks arise from individual edges.
In this paper, we propose a \emph{node-based} formulation that allows for more powerful estimation of multiple GGMs, under the assumption that network similarities and differences arise from \emph{nodes} whose connectivity patterns to other nodes are shared or disrupted across conditions.

\section{Node-Based Joint Graphical Lasso}
\label{sec:NJGL}

In this section, we  first discuss the failure of a naive approach for node-based learning of multiple GGMs.
We then present a norm that will play a critical role
in our formulations for this task. Finally, we discuss two approaches
for node-based learning of multiple GGMs.

\label{sec:us}
\subsection{Why is Node-Based Learning Challenging?}
\label{sec:hard}
At first glance, node-based learning of multiple GGMs seems straightforward. For instance, consider the task of estimating $K=2$ networks under the assumption that the connectivity patterns of individual nodes differ across the networks. It seems that we could simply modify (\ref{eqn:fgl}) combined with (\ref{eqn:fused}) as follows,
\begin{equation}
\Maximize_{\bThetaone \in \mathbb{S}_{++}^p, \bThetatwo \in \mathbb{S}_{++}^p} \left\{ L(\bThetaone, \bThetatwo)  - \lambda_1 {\|\bThetaone\|}_1  - \lambda_1 {\|\bThetatwo\|}_1 - \lambda_2
\sum_{j=1}^p \|{\bTheta}^1_{j} - {\bTheta}^2_{j} \|_2 \right\},
\label{eqn:fgl-grouplasso}
\end{equation}
where ${\bTheta}^k_{j}$ is the $j$th column of the matrix ${\bTheta}^k$. This amounts to applying a \emph{group lasso} \citep{grouplasso} penalty to the columns of
${\bTheta}^1-{\bTheta}^2$.  Equation (\ref{eqn:fgl-grouplasso}) seems to accomplish our goal of encouraging $\bThetaone_j=\bThetatwo_j$. We will refer to this as the \emph{naive group lasso} approach.

\begin{figure}
\begin{center}
\subfigure[Naive group lasso]{\includegraphics[width=0.2\linewidth,clip]
{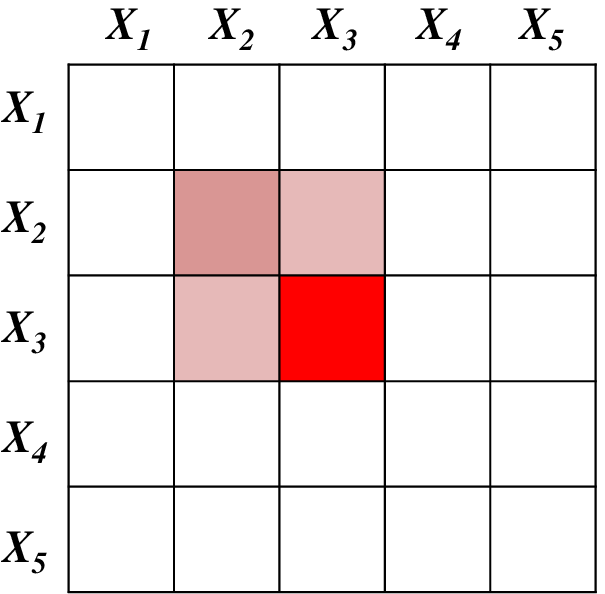}} \qquad
\subfigure[RCON: $\ell_1/\ell_1$]{\includegraphics[width=0.2\linewidth,clip]
{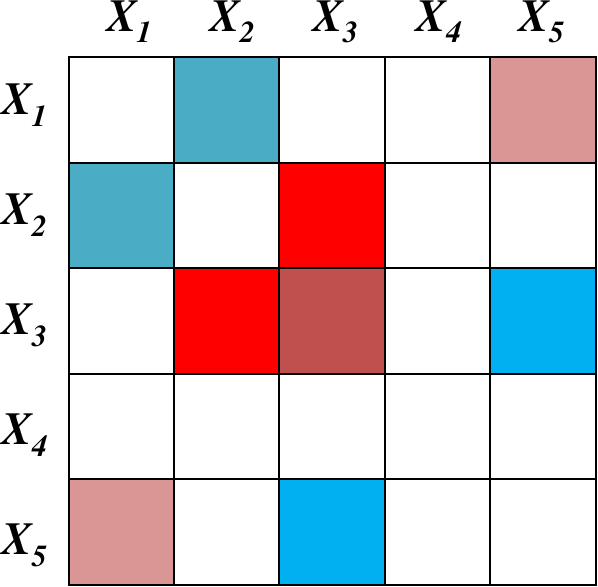}} \qquad
\subfigure[RCON: $\ell_1/\ell_2$]{\includegraphics[width=0.2\linewidth,clip]
{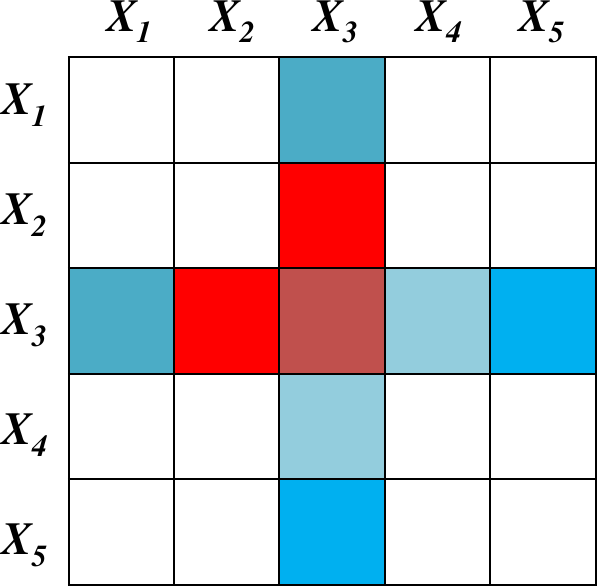}} \qquad
\subfigure[RCON:  $\ell_1/\ell_{\infty}$]{\includegraphics[width=0.2\linewidth,clip]
{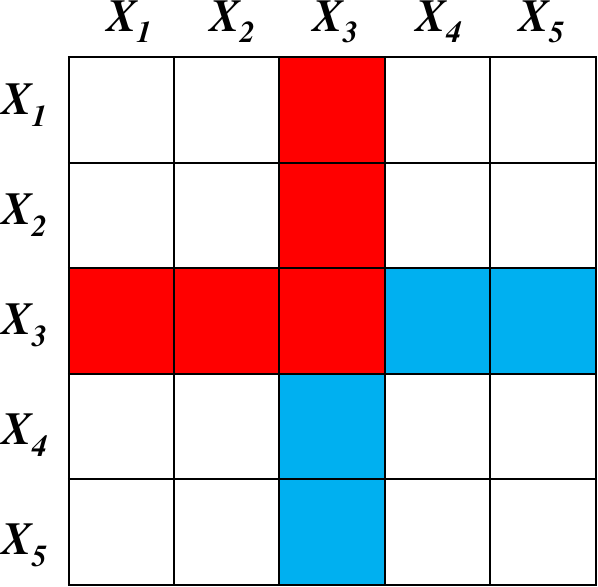}}
\end{center}
\vspace{-6mm}
\caption{\label{fig:RCON} Toy example of the results from applying various penalties in order to estimate a $5 \times 5$ matrix, under a symmetry constraint. Zero elements are shown in white; non-zero elements are shown in shades of red (positive elements) and blue (negative elements).
\emph{(a):} The naive group lasso applied to the columns of the matrix  yields non-zero elements that are the \emph{intersection}, rather than the \emph{union}, of a set of rows and columns.
\emph{(b):} The RCON penalty using an $\ell_1/\ell_1$ norm results in unstructured sparsity in the estimated matrix.
\emph{(c):} The  RCON penalty  using an $\ell_1/\ell_2$ norm results in entire rows and columns of non-zero elements.
\emph{(d):} The RCON penalty using an $\ell_1/\ell_{\infty}$ norm results in  entire rows and columns of non-zero elements; many  take on a single maximal (absolute) value.}
\end{figure}

In (\ref{eqn:fgl-grouplasso}), we have applied the group lasso using $p$ groups; the $j$th group is the $j$th column
of $\bThetaone-\bThetatwo$.
Due to the symmetry of $\bThetaone$ and $\bThetatwo$, there is substantial overlap among the $p$ groups: the $(i,j)$th element of $\bThetaone-\bThetatwo$ is contained in both the $i$th and $j$th  groups.
In the presence of overlapping groups, the group lasso penalty yields estimates whose \emph{support is the complement of the union of groups} \citep{OverlapGroupLasso,Obozinski-11}. Figure~\ref{fig:RCON}(a)  displays a simple example of the results obtained if we attempt to estimate $({\bSigma}^1)^{-1}-({\bSigma}^2)^{-1}$ using  (\ref{eqn:fgl-grouplasso}). The figure reveals that (\ref{eqn:fgl-grouplasso}) cannot be used to detect node perturbation.

A naive approach to co-hub detection is challenging for a similar reason. Recall that the $j$th node is a co-hub if the $j$th columns of both $\varone$ and  $\vartwo$ contain predominantly non-zero elements, and let $\diag(\bTheta)$ denote a matrix consisting of the diagonal elements
of $\bTheta$. It is tempting to formulate the optimization problem
\begin{equation*}
\Maximize_{\bThetaone \in \mathbb{S}_{++}^p, \bThetatwo \in \mathbb{S}_{++}^p}  \left\{ L(\bThetaone, \bThetatwo)  - \lambda_1 {\|\bThetaone\|}_1  - \lambda_1 {\|\bThetatwo\|}_1 -
  \lambda_2 \sum_{j=1}^p {\left\| \left[ \begin{array}{c} \bThetaone - \diag(\bThetaone) \\ \bThetatwo - \diag(\bThetatwo) \end{array} \right]_j \right\|}_2 \right\},
\label{eqn:cnjgl-grouplasso}
\end{equation*}
where  the group  lasso penalty  encourages the off-diagonal elements of many of the columns  to be simultaneously zero in $\bThetaone$ and $\bThetatwo$. Unfortunately, once again, the presence of overlapping groups encourages the support of the matrices $\bThetaone$ and $\bThetatwo$ to
be the intersection of a set of rows and columns, as in Figure~\ref{fig:RCON}(a), rather than the union of a set of rows and columns.

\subsection{Row-Column Overlap Norm}
Detection of perturbed nodes or co-hub nodes requires a penalty function that, when applied  to a matrix, yields a support given by the union of a set of rows and columns. We now propose
the \emph{\rcon} (\RCONnospace) for this task.

\begin{definition}
The \rcon \;(RCON)  induced by a matrix norm $\matrixnorm$ is defined as
\begin{eqnarray*} \label{eqn:rcon}
\begin{array}{rcll}
\RCONmatrix(\bThetaone,\bThetatwo,\ldots,\bThetaK) &=& \MIN_{\Vone,\Vtwo,\ldots,\VK} & \mnormfull{\left[\begin{array}{c} \Vone \\ \Vtwo \\ \vdots \\ \VK \end{array} \right]}   \\
&& \mathrm{subject \; to} & \bThetak = \Vk + (\Vk)^T \;\; \mathrm{ for } \; k=1,\ldots,K.
\end{array}
\end{eqnarray*}
\end{definition}

It is easy to check that $\RCONmatrix$ is indeed a norm for all matrix norms $\matrixnorm$. Also, when
$\matrixnorm$ is symmetric in its argument, that is, $\mnorm{V} = \mnorm{V^T}$, then
\begin{equation*}
\RCONmatrix(\bThetaone,\bThetatwo,\ldots,\bThetaK) = \frac{1}{2}\mnormfull{\left[\begin{array}{c} \bThetaone \\ \bThetatwo \\ \vdots \\ \bThetaK \end{array} \right]}.
\end{equation*}

Thus if $\| \cdot \|$ is an $\ell_1/\ell_1$ norm, then $\Omega(\bThetaone,\bThetatwo,\ldots,\bThetaK) = \frac{1}{2} \sum_{k=1}^K \sum_{i,j} |\Theta^k_{ij}|$.

{We now discuss the motivation behind Definition 1. Any symmetric matrix $\bThetak$ can be (non-uniquely) decomposed as $\Vk + (\Vk)^T$; note that $\Vk$ need not be symmetric.
This amounts to interpreting $\bThetak$ as a set of columns (the columns of $\Vk$) \emph{plus} a set of rows (the columns of $\Vk$, transposed).
In this paper, we are interested in the particular case of \RCON penalties where  $\matrixnorm$ is an $\ell_1/\ell_q$ norm, given by
$\| V \|= \sum_{j=1}^p \|\tV_j\|_q$, where $1 \leq q \leq \infty$.
With a little abuse of notation, we will let $\Omega_q$ denote $\RCONmatrix$ when $\matrixnorm$ is given by the $\ell_1/\ell_q$ norm.
 Then
 $\Omega_q$ encourages  $\bThetaone,\bThetatwo,\ldots,\bThetaK$ to decompose into $\Vk$ and $(\Vk)^T$
 such that the summed $\ell_q$ norms of all of the columns (concatenated over $\Vone,\ldots,\VK$) is small.
 This encourages structures of interest on the columns \emph{and} rows of $\bThetaone,\bThetatwo,\ldots,\bThetaK$. }

{To illustrate this point, in Figure \ref{fig:RCON} we display schematic results obtained from estimating a $5 \times 5$ matrix subject to the RCON penalty $\Omega_q$,  for $q = 1$, $2$, and $\infty$.
We see from Figure~\ref{fig:RCON}(b) that when $q=1$, the RCON penalty yields a matrix estimate with unstructured sparsity; recall that $\Omega_1$ amounts to an $\ell_1$ penalty applied to the matrix entries.
When $q=2$ or $q=\infty$, we see from Figures~\ref{fig:RCON}(c)-(d) that the RCON penalty yields a sparse matrix estimate for which the non-zero elements are a set of rows \emph{plus} a set of columns---that is, the union of a set of rows and columns.}

We note that $\Omega_2$ {can be derived from} the \emph{overlap norm} \citep{Obozinski-11,OverlapGroupLasso}
applied to groups given by rows and columns of $\bThetaone,\ldots,\bThetaK$.
Details are described in Appendix  \ref{app:obozinski}.
Additional properties of RCON  are discussed in Appendix~\ref{sec:appendix_overlap}.

\subsection{Node-Based Approaches for Learning GGMs}
We discuss two approaches for node-based learning of GGMs.
The first  promotes networks whose differences are attributable to  perturbed nodes.
The second encourages the networks to share co-hub nodes.

\subsubsection{Perturbed-node Joint Graphical Lasso}
Consider the task of jointly estimating {$K$} precision matrices by solving
\begin{equation}
\Maximize_{\bThetaone, \bThetatwo, \ldots ,\bThetaK \in \mathbb{S}_{++}^p} \left\{
L(\bThetaone,\bThetatwo,\ldots,\bThetaK) - \lambda_1 \sum_{k=1}^K  \|\bThetak\|_1 - \lambda_2 \sum_{k<k'} \Omega_q(\bThetak- \bTheta^{k'}) \right\}.
\label{eqn:sjgl}
\end{equation}
We refer to the convex optimization problem  (\ref{eqn:sjgl}) as the \emph{\npjgl} (\PNJGLnospace).
Let ${\hat\bTheta}^1,{\hat\bTheta}^2,\ldots,{\hat\bTheta}^K$ denote the solution to (\ref{eqn:sjgl}); these serve as estimates for $(\bSigma^1)^{-1},\ldots,(\bSigma^K)^{-1}$.
 In (\ref{eqn:sjgl}), $\lambda_1$ and $\lambda_2$ are nonnegative tuning parameters, and  $q \geq 1$.
When $\lambda_2=0$, (\ref{eqn:sjgl}) amounts simply to applying the graphical lasso optimization problem (\ref{eqn:glasso}) to each condition separately in order to separately estimate $K$  networks. When $\lambda_2>0$, we are encouraging similarity among the $K$ network estimates. {When $q=1$, we have the following observation.}

\begin{remark}
The FGL formulation (Equations \ref{eqn:fgl} and \ref{eqn:fused}) is a special case of PNJGL (\ref{eqn:sjgl}) with $q=1$.
\end{remark}

In other words, when $q=1$, (\ref{eqn:sjgl}) amounts to the \emph{edge-based} approach  of \cite{Danaher2012} that encourages many entries of ${\hat{\bTheta}}^k - {\hat{\bTheta}}^{k'}$ to equal zero.

However, when $q = 2$ or $q = \infty$, then (\ref{eqn:sjgl}) amounts to a  \emph{node-based approach}: the support of ${\hat{\bTheta}}^k - {\hat{\bTheta}}^{k'}$ is encouraged to be a union of a few rows and the corresponding columns. These can be interpreted as  a set of nodes that are perturbed across the conditions.
An example of the sparsity structure detected by \NPJGL with $q=2$ or $q=\infty$ is shown in Figure \ref{fig:node-pert}.

\subsubsection{Co-hub Node Joint Graphical Lasso}
We now consider jointly estimating $K$ precision matrices by solving  the convex optimization problem

{\footnotesize
\begin{equation}
 \Maximize_{\bThetaone,\bThetatwo, \ldots, \bThetaK \in \mathbb{S}_{++}^p}  \left\{ L(\bThetaone,\bThetatwo,\ldots,\bThetaK) - \lambda_1 \SUM_{k=1}^K \|\bThetak\|_1 -
                                                 \lambda_2 \Omega_q (\bThetaone - \diag(\bThetaone), \ldots, \bThetaK - \diag(\bThetaK)) \right\}.
\label{eqn:cnjgl_form}
\end{equation}
}
We refer to  (\ref{eqn:cnjgl_form}) as the \emph{co-hub node joint graphical lasso} (CNJGL) formulation. In (\ref{eqn:cnjgl_form}), $\lambda_1$ and $\lambda_2$ are nonnegative tuning parameters, and  $q \geq 1$. When $\lambda_2=0$ then this amounts to a graphical lasso optimization problem applied to each network separately; however, when $\lambda_2>0$, a shared structure is encouraged among the $K$ networks. In particular,
(\ref{eqn:cnjgl_form}) encourages network estimates that have a common set of hub nodes---that is, it encourages the supports of $\bThetaone,\bThetatwo,\ldots,\bThetaK$ to be the same, and  the union of  a  set of rows and columns.

CNJGL can be interpreted as a node-based extension of the GGL proposal \citep[given in Equations~\ref{eqn:fgl} and \ref{eqn:GGL_penalty}, and originally proposed by][]{Danaher2012}. While GGL encourages the $K$ networks to share a common edge support, CNJGL instead encourages the networks to share a common node support.

We now remark on an additional connection between CNJGL and the graphical lasso.
\begin{remark} If $q=1$, then 
CNJGL  amounts to a modified graphical lasso on each network separately, with 
a penalty of $\lambda_1$  applied to the diagonal elements, and a penalty of $\lambda_1+\lambda_2/2$  applied to the off-diagonal elements. 
\end{remark}

\section{Algorithms}
\label{sec:ADMM}

The \PNJGL and \CNJGL optimization problems (\ref{eqn:sjgl}, \ref{eqn:cnjgl_form}) are convex, and so can be directly solved in the modeling environment \verb=cvx= \citep{cvx}, which calls conic interior-point solvers such as \verb=SeDuMi= or \verb=SDPT3=. However, when applied to solve semi-definite programs, second-order methods such as the interior-point algorithm do not scale well with the problem size.

We next examine the use of existing first-order methods to solve (\ref{eqn:sjgl}) and (\ref{eqn:cnjgl_form}).
Several first-order algorithms have been proposed for minimizing a least squares objective with a group lasso penalty \citep[as in][]{grouplasso} in the presence of overlapping groups \citep{Fixedpoint_proximal2011,smooth_proximal2011,primal-dual_overlap2010}. Unfortunately, those algorithms  cannot be applied to the \NPJGL and \CNJGL formulations, which  involve the \RCON penalty  rather than simply a standard group lasso with overlapping groups.
The \RCON penalty is a variant of the overlap norm proposed in \citet{Obozinski-11}, and indeed those authors propose an algorithm for minimizing a least squares objective subject to the overlap norm. However, in the context of CNJGL and PNJGL, the objective of interest is a Gaussian log likelihood, and the algorithm of \citet{Obozinski-11} cannot be easily applied.

Another possible approach for solving (\ref{eqn:sjgl}) and (\ref{eqn:cnjgl_form}) involves the use of a standard first-order method, such as a projected subgradient approach.
Unfortunately, such an approach is not straightforward, since computing the subgradients of the \RCON penalty involves solving a non-trivial optimization problem (to be discussed in detail in Appendix~\ref{sec:appendix_overlap}). Similarly,
a proximal gradient approach for solving (\ref{eqn:sjgl}) and (\ref{eqn:cnjgl_form}) is challenging because the proximal operator of the combination of the overlap norm and the $\ell_1$ norm has no closed form.

To overcome the challenges outlined above, we propose to solve the \NPJGL and \CNJGL problems using an \emph{alternating {direction} method of multipliers} algorithm \citep[ADMM; see, e.g.,][]{ADMMBoyd}.

\subsection{The ADMM Approach}
Here we briefly outline the standard ADMM approach for a general optimization problem,
\begin{eqnarray}\label{eq:opt-simple}
\begin{array}{cc}
\Minimize_{X} & g(X) + h(X) \\
\mbox{subject to} & X \in \Xcal.
\end{array}
\end{eqnarray}
ADMM is  attractive in cases where the proximal operator of $g(X)+h(X)$ cannot be easily computed, but where the proximal operator of $g(X)$ and the proximal operator of $h(X)$ are easily obtained.
The approach  is as follows \citep{ADMMBoyd,Eckstein-Bertsekas-92, Gabay-Mercier-76}:
\begin{enumerate}
\item Rewrite the optimization problem (\ref{eq:opt-simple}) as
\begin{eqnarray} \label{eq:opt-simple-decouple}
\begin{array}{cc}
\Minimize_{X,Y} & g(X) + h(Y) \\
\mbox{subject to} & X \in \Xcal,\; X = Y,
\end{array}
\end{eqnarray}
where here we have decoupled $g$ and $h$ by introducing a new optimization variable, $Y$.
\item Form the augmented Lagrangian to (\ref{eq:opt-simple-decouple}) by first forming the Lagrangian,
$$L(X,Y,\Lambda) = g(X) + h(Y) + \langle \Lambda, X - Y \rangle,$$ and then augmenting it by a quadratic function of $X-Y$,
 $$L_{\rho}(X,Y,\Lambda) = L(X,Y,\Lambda) + \frac{\rho}{2}\|X - Y\|_F^2,$$
 where $\rho$ is a positive constant.
\item Iterate until convergence:
\begin{enumerate}
\item Update each primal variable in turn by minimizing the augmented Lagrangian with respect to that variable, while keeping all other variables fixed.
The updates in the $k$th iteration are as follows:
\begin{eqnarray*}
X^{k+1} &\leftarrow& \arg\min_{X \in \Xcal} L_{\rho} (X,Y^k,\Lambda^k), \\
Y^{k+1} &\leftarrow& \arg\min_{Y} L_{\rho} (X^{k+1},Y,\Lambda^k).
\end{eqnarray*}
\item Update the dual variable using a dual-ascent update,
\begin{eqnarray*}
\Lambda^{k+1} &\leftarrow & \Lambda^k + \rho (X^{k+1} - Y^{k+1}).
\end{eqnarray*}
\end{enumerate}
\end{enumerate}

The standard ADMM presented here involves minimization over two primal variables, $X$ and $Y$. For our problems, we will use a similar algorithm {but with \emph{more than two primal variables}}. More details about the algorithm and its convergence are discussed in Section 4.2.4.

\subsection{ADMM Algorithms for \NPJGL and \CNJGL}
\label{sec:algorithm}
Here we outline the ADMM algorithms for the \NPJGL and \CNJGL optimization problems; we refer the reader to  Appendix \ref{sec:update_derv_ADMM} for detailed derivations of the update rules.

\subsubsection{ADMM Algorithm for \NPJGL}
Here we consider solving \NPJGL with $K=2$; {the extension for $K>2$ is slightly more complicated.}
To begin, we note that (\ref{eqn:sjgl}) can be rewritten as
\begin{eqnarray}
\begin{array}{cc}
\Maximize_{\bThetaone, \bThetatwo \in \mathbb{S}_{++}^p, V \in \mathbb{R}^{p \times p}} & \left\{
L(\bThetaone,\bThetatwo) - \lambda_1  \|\bThetaone\|_1  - \lambda_1  \|\bThetatwo\|_1  - \lambda_2 \SUM_{j=1}^p \|\tV_j\|_{q} \right\} \\
 \mbox{subject  to } &
\bThetaone-\bThetatwo = V+V^T.
\end{array}
\label{eq:pnjgl-2}
\end{eqnarray}
We now reformulate (\ref{eq:pnjgl-2}) by introducing new variables, so as to decouple some of the terms in the objective function that are difficult to optimize jointly:
\begin{eqnarray} \label{eq:NPJGL_reform}
\begin{array}{cc}
\Minimize_{\bThetaone \in \PD,\bThetatwo \in \PD,\Zone,\Ztwo,\tV,\tW} &
\left\{ -L(\bThetaone,\bThetatwo)  +  \regone\|\Zone\|_1 + \regone\|\Ztwo\|_1 + \regtwo \SUM_{j=1}^p \|\tV_j\|_{q} \right\} \\
\mbox{subject to } &  \bThetaone - \bThetatwo = \tV + \tW, \tV = \tW^T,
                         \bThetaone = \Zone, \bThetatwo = \Ztwo.
\end{array}
\end{eqnarray}
The augmented Lagrangian to (\ref{eq:NPJGL_reform}) is  given by
\begin{eqnarray}
\begin{array}{lll}
&-&L(\bThetaone,\bThetatwo) +  \regone\|\Zone\|_1 + \regone\|\Ztwo\|_1  +
\regtwo\SUM_{j=1}^p \|\tV_j\|_q  +  \langle \tF, \bThetaone - \bThetatwo - (\tV + \tW) \rangle   \\
& +&  \langle \tG, \tV - \tW^T \rangle + \langle \Qone, \bThetaone - \Zone \rangle + \langle \Qtwo, \bThetatwo - \Ztwo \rangle + \frac{\rho}{2}\|\bThetaone - \bThetatwo - (\tV + \tW)\|_F^2    \\
&+ &\frac{\rho}{2}\|\tV - \tW^T\|_F^2 + \frac{\rho}{2}\|\bThetaone -
\Zone\|_F^2 + \frac{\rho}{2}\|\bThetatwo - \Ztwo\|_F^2.
\end{array}
\label{eq:PNJGL_aug_lag}
\end{eqnarray}
In (\ref{eq:PNJGL_aug_lag}) there are six primal variables and four dual variables.
Based on this augmented Lagrangian, the complete ADMM algorithm for (\ref{eqn:sjgl}) is given in Algorithm \ref{algo:PNJGL_ADMM},  in which  the operator $\expand$ is given by
{\small
$$\expand(\tA,\rho, n_k) = \argmini_{{\bTheta}\in \mathbb{S}_{++}^p} \left\{-n_k\log\det({\bTheta}) + \rho\|{\bTheta} - \tA\|_F^2 \right\}
                                = \frac{1}{2}\tU\left({\tD} + \sqrt{{\tD}^2 + \frac{2n_k}{\rho}\tI}\right)\tU^T,$$
}
where $\tU{\tD} \tU^T$ is the eigenvalue decomposition of a symmetric matrix $\tA$,
and as mentioned earlier, $n_k$ is the number of observations in the
$k$th class. The operator $\thresh_q$ is given by
$$\thresh_q(\tA,\lambda) = \argmini_{\tX} \left\{  \frac{1}{2}\|\tX - \tA\|_F^2 + \lambda \SUM_{j=1}^p \|\tX_j\|_q \right\},$$
and is also known as the proximal operator corresponding to the
$\ell_1/\ell_q$ norm. For $q=1,2,\infty$, $\thresh_q$ takes a simple
form \citep[see, e.g., Section 5 of][]{Duchi-singer-09}.

\begin{algorithm}[H]
{\bf input}: $\rho > 0, \mu > 1, t_{\max} > 0$\;
{\bf Initialize}: Primal variables to the identity matrix and dual variables to the zero matrix\;
\For{t = 1:$t_{\max}$}{ $\rho \leftarrow \mu\rho$\;
\While{Not converged}{
$\bThetaone \leftarrow \expand \left(\frac{1}{2}(\bThetatwo +
\tV + \tW + \Zone) - \frac{1}{2\rho}(\Qone + n_1 S^1 + \tF), \rho,n_1 \right)$\;
$\bThetatwo \leftarrow \expand \left(\frac{1}{2}(\bThetaone - (\tV + \tW) + \Ztwo) - \frac{1}{2\rho}
(\Qtwo + n_2 S^2 - \tF), \rho, n_2 \right)$\;
${\tZ}^i \leftarrow \thresh_1\left({\bTheta}^i + \frac{\tQ^i}{\rho}, \frac{\lambda_1}{\rho} \right) \mbox{ for } i=1,2$\;
$\tV \leftarrow \thresh_q\left(\frac{1}{2}({\tW}^T - \tW + (\bThetaone -
\bThetatwo)) + \frac{1}{2\rho}(\tF - \tG),{\frac{\regtwo}{2\rho}}
\right)$\;
$\tW \leftarrow \frac{1}{2}({\tV}^T - \tV + (\bThetaone - \bThetatwo)) + \frac{1}{2\rho}(\tF + {\tG}^T)$\;
$\tF \leftarrow \tF +
\rho(\bThetaone - \bThetatwo - (\tV + \tW))$\;
 $\tG \leftarrow \tG +\rho(\tV - {\tW}^T)$\;
${\tQ}^i \leftarrow {\tQ}^i + \rho({\bTheta}^i - {\tZ}^i)  \mbox{ for } i=1,2$
} }
\caption{\label{algo:PNJGL_ADMM} \ADMM algorithm for the \NPJGL optimization problem (\ref{eqn:sjgl})}
\end{algorithm}

\subsubsection{ADMM Algorithm for \CNJGL}

The CNJGL formulation in (\ref{eqn:cnjgl_form}) is equivalent to
\begin{eqnarray}
\begin{array}{cc}
\Minimize_{\bThetai \in \mathbb{S}_{++}^p, V^i \in \mathbb{R}^{p \times p}, i = 1 \ldots K} & 
-L(\bThetaone,\bThetatwo,\ldots,\bThetaK) + \lambda_1 \SUM_{i=1}^K \|\bThetai\|_1  + \lambda_2  \SUM_{j=1}^p \left\| \left[\begin{array}{c} V^1 \\ V^2 \\ \vdots \\ V^K \end{array}\right]_j \right\|_{q}  \\
 \mbox{subject  to } & \bThetai-\diag(\bThetai) = V^i+(V^i)^T \mbox{ for } i=1,\ldots,K.
\end{array}
\label{eq:cnjgl_form_2}
\end{eqnarray}
One can easily see that the problem (\ref{eq:cnjgl_form_2}) is equivalent to the problem
{\footnotesize
\begin{eqnarray}
\begin{array}{cc}
\Minimize_{\bThetai \in \mathbb{S}_{++}^p, \Vic \in \mathbb{R}^{p \times p}, i = 1 \ldots K} & -L(\bThetaone,\bThetatwo,\ldots,\bThetaK) + \lambda_1 \SUM_{i=1}^K \|\bThetai\|_1 + \lambda_2  \SUM_{j=1}^p \left\| \left[\begin{array}{c} \Vonec - \diag(\Vonec) \\ \Vtwoc - \diag(\Vtwoc) \\ \vdots \\ \VKc - \diag(\VKc) \end{array}\right]_j \right\|_{q}  \\
\mbox{subject to } & \bThetai = \Vic + (\Vic)^T \; \mbox{ for } i = 1,2,\ldots,K,
\end{array}
\label{eq:cnjgl_form_3}
\end{eqnarray} }
\noindent in the sense that the optimal solution $\{\tV^i\}$ to (\ref{eq:cnjgl_form_2}) and the optimal solution $\{\Vic\}$ to (\ref{eq:cnjgl_form_3}) have the following relationship: $\tV^i = \Vic - \diag(\Vic) \; \mbox{for } i = 1,2,\ldots,K$.
We now present an ADMM algorithm for solving (\ref{eq:cnjgl_form_3}).
We reformulate (\ref{eq:cnjgl_form_3}) by introducing additional variables in order  to decouple some terms of the objective that are difficult to optimize jointly:
{\footnotesize
\begin{eqnarray} \label{eq:cnjgl_reform}
\begin{array}{cc}
\Minimize_{\bThetai \in \mathbb{S}_{++}^p, \Zi, \Vic, \Wi \in \mathbb{R}^{p \times p}} & -L(\bThetaone,\bThetatwo,\ldots,\bThetaK) + \lambda_1 \SUM_{i=1}^K \|\Zi\|_1 +
\lambda_2 \SUM_{j=1}^p \left\| \left[\begin{array}{c} \Vonec - \diag(\Vonec) \\ \Vtwoc - \diag(\Vtwoc) \\ \vdots \\ \VKc - \diag(\VKc) \end{array}\right]_j \right\|_{q}  \\
\mbox{subject to } & \vari = \Vic + \Wi, \Vic = (\Wi)^T, \bThetai = \Zi\; \mbox{for } i = 1,2,\ldots,K.
\end{array}
\end{eqnarray}}
\noindent The augmented Lagrangian to (\ref{eq:cnjgl_reform}) is given by
{\footnotesize
\begin{eqnarray} \label{eq:CNJGL_aug_lag}
\begin{array}{rcl}
\SUM_{i=1}^K n_i (-\log\det(\bThetai) + \trace(\Sii\bThetai))  + \lambda_1 \SUM_{i=1}^K  \|\Zi\|_1 + \lambda_2 \SUM_{j=1}^p \left\| \left[\begin{array}{c} \Vonec - \diag(\Vonec) \\ \Vtwoc - \diag(\Vtwoc) \\ \vdots \\ \VKc - \diag(\VKc) \end{array}\right]_j \right\|_{q} &+&\\
\SUM_{i=1}^K \left\{\langle \Fi, \vari - (\Vic + \Wi) \rangle + \langle \Gi, \Vic - (\Wi)^T \rangle + \langle \Qi, \bThetai - \Zi \rangle \right\}  &+&\\
\frac{\rho}{2}\SUM_{i=1}^K \left\{ \|\vari - (\Vic + \Wi)\|_F^2 + \|\Vic - (\Wi)^T\|_F^2 + \|\bThetai - \Zi\|_F^2  \right\}.	&&
\end{array}
\end{eqnarray}}
\noindent The corresponding ADMM algorithm is given in Algorithm \ref{algo:CNJGL_ADMM}.

\begin{algorithm}[H]
{\bf input}: $\rho > 0, \mu > 1, t_{\max} > 0$\;
{\bf Initialize}: Primal variables to the identity matrix and dual variables to the zero matrix\;
\For{t = 1:$t_{\max}$}{ $\rho \leftarrow \mu\rho$\;
\While{Not converged}
{
 $\bThetai \leftarrow \expand \left(\frac{1}{2}(\Vic + \Wi + \Zi) - \frac{1}{2\rho}(\Qi + n_i\Sii + \Fi), \rho, n_i \right) \mbox{ for } i=1,\ldots, K$\;
 $\Zi \leftarrow \thresh_1\left(\bThetai + \frac{\Qi}{\rho}, \frac{\lambda_1}{\rho} \right) \mbox{ for } i=1,\ldots,K$\;
Let $\Ci = \frac{1}{2}((\Wi)^T - \Wi + \bThetai) + \frac{1}{2\rho}(\Fi - \Gi) \mbox{ for } i=1,\ldots,K$\;
$\left[\begin{array}{c}\Vonec \\ \Vtwoc \\ \vdots \\ \VKc \end{array} \right] \leftarrow \thresh_q \left( \left[\begin{array}{c} \Cone - \diag(\Cone) \\ \Ctwo - \diag(\Ctwo) \\ \vdots \\ \CK - \diag(\CK) \end{array} \right],\frac{\lambda_2}{2\rho} \right)
+ \left[\begin{array}{c} \diag(\Cone) \\ \diag(\Ctwo) \\ \vdots \\ \diag(\CK) \end{array} \right]$\;
$\Wi \leftarrow \frac{1}{2}((\Vic)^T - \Vic + \bThetai) + \frac{1}{2\rho}(\Fi + ({\Gi})^T) \mbox{ for } i=1,\ldots,K$\;
$\Fi \leftarrow \Fi + \rho(\bThetai - (\Vic + \Wi)) \mbox{ for } i=1,\ldots,K$\;
$\Gi \leftarrow \Gi + \rho(\Vic - {(\Wi)}^T) \mbox{ for } i=1,\ldots,K$\;
$\Qi \leftarrow \Qi + \rho(\bThetai - \Zi) \mbox{ for } i=1,\ldots,K$

}
}
\caption{\label{algo:CNJGL_ADMM} \ADMM algorithm for the \CNJGL optimization problem (\ref{eqn:cnjgl_form})}
\end{algorithm}

\subsubsection{Numerical Issues and Run-Time of the ADMM Algorithms}
We  set $\mu = 5$, $\rho = 0.5$ and  $t_{\max} = 1000$ in the \PNJGL and \CNJGL algorithms.
In our implementation of these algorithms, the stopping criterion for the inner loop (corresponding to a fixed $\rho$) is
\begin{eqnarray*}
\MAX_{i \in \{1,2,\ldots,K\}} \left\{\frac{\|(\vari)^{(k+1)} - {(\vari)}^{(k)}\|_F}{\|(\vari)^{(k)}\|_F} \right\} \leq \epsilon,
\end{eqnarray*}
where $(\vari)^{(k)}$ denotes the estimate of $\vari$ in the $k$th iteration of the ADMM algorithm, and $\epsilon$ is a tolerance that is chosen in our experiments to equal $10^{-4}$.

The per-iteration complexity of the ADMM algorithms for  \CNJGL and \PNJGL (with $K=2$) is $O(p^3)$; this is the complexity of computing the SVD. On the other hand, the complexity of {a general} interior point method is $O(p^6)$.
 In a small example with $p = 30$, run on an Intel Xeon X3430 2.4Ghz CPU, the interior point method (using \verb=cvx=, which calls \verb=Sedumi=) takes 7 minutes to run, while the \ADMM algorithm for PNJGL, coded in \verb=Matlab=, takes only 0.58 seconds. When $p = 50$, the times are 3.5 hours and 2.1 seconds, respectively. Let $\hat{\var}^1, \hat{\var}^2$ and $\bar{\var}^1, \bar{\var}^2$  denote the solutions obtained by \ADMM and \verb=cvx=, respectively.
We observe that on average, the error  $\MAX_{i \in \{1,2\}} \left\{ \|\hat{\var}^i - \bar{\var}^i\|_F / \|\bar{\var}^i\|_F  \right\}$ is on the order of $10^{-4}$. {Thus, the algorithm has good empirical accuracy in recovering the optimal solution}.

\begin{figure}[!htbp]
\centering
\subfigure[CNJGL]{\includegraphics[width = 0.34\linewidth,clip]{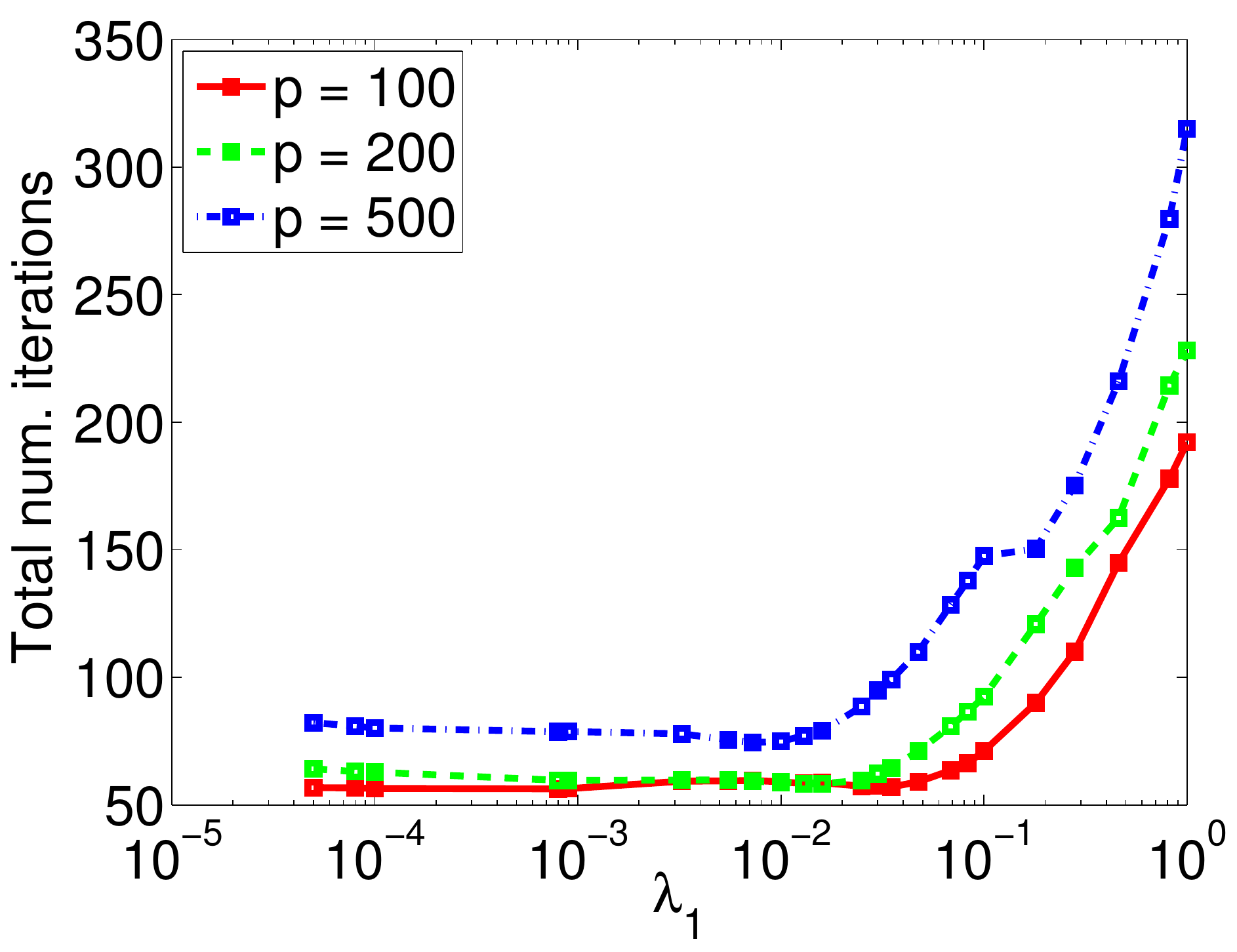}}
\subfigure[CNJGL]{\includegraphics[width = 0.34\linewidth,clip]{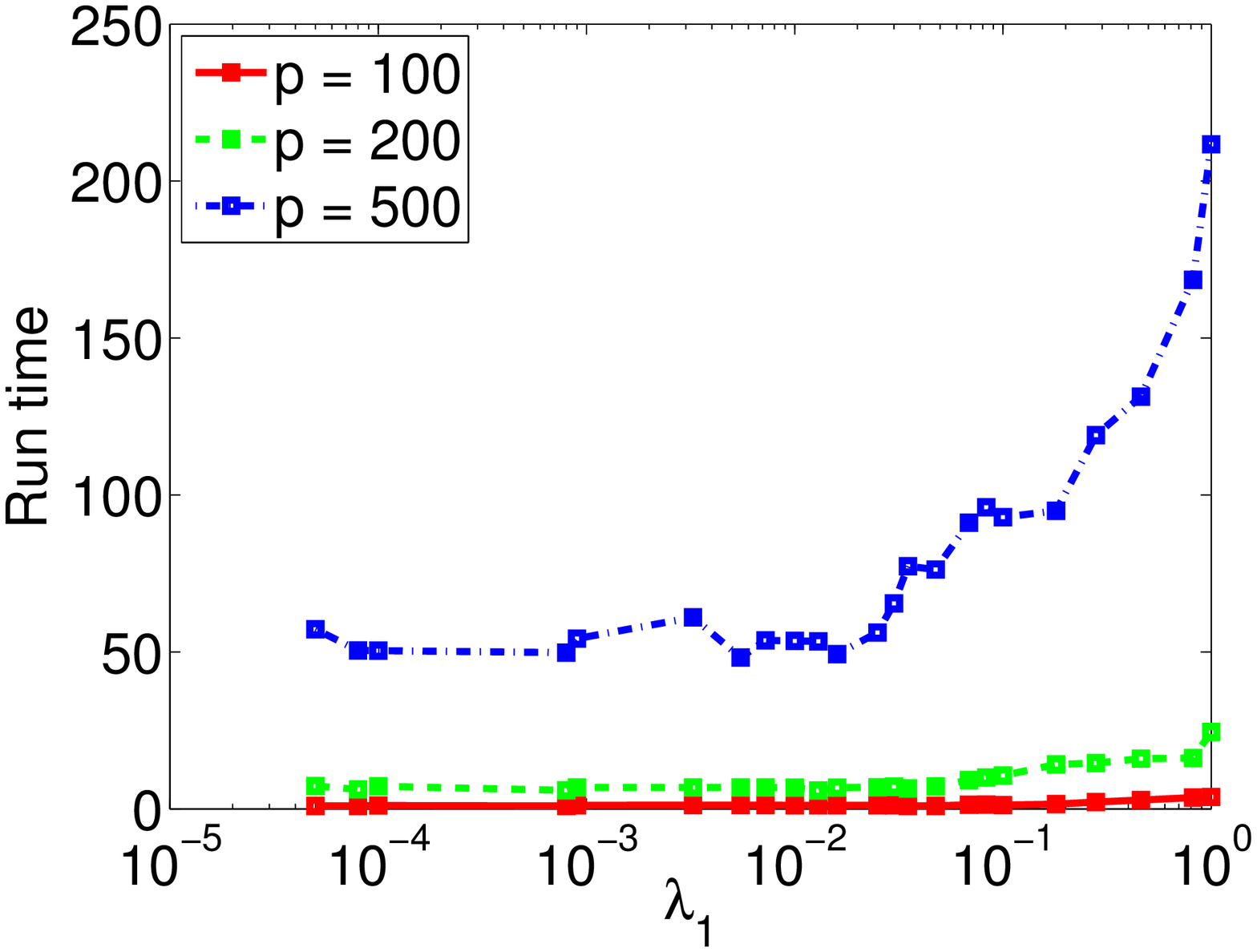}}
\subfigure[PNJGL]{\includegraphics[width = 0.34\linewidth,clip]{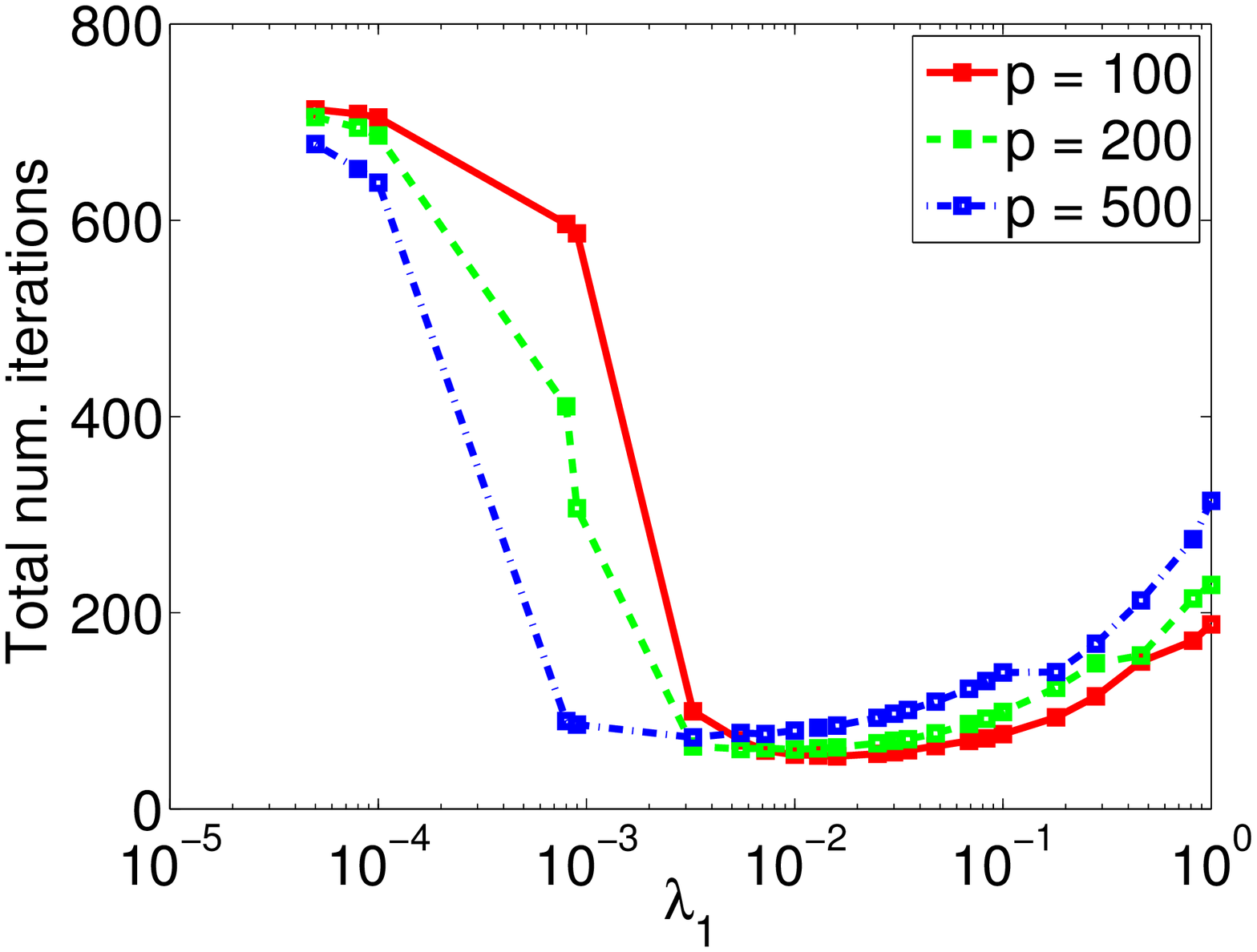}}
\subfigure[PNJGL]{\includegraphics[width = 0.34\linewidth,clip]{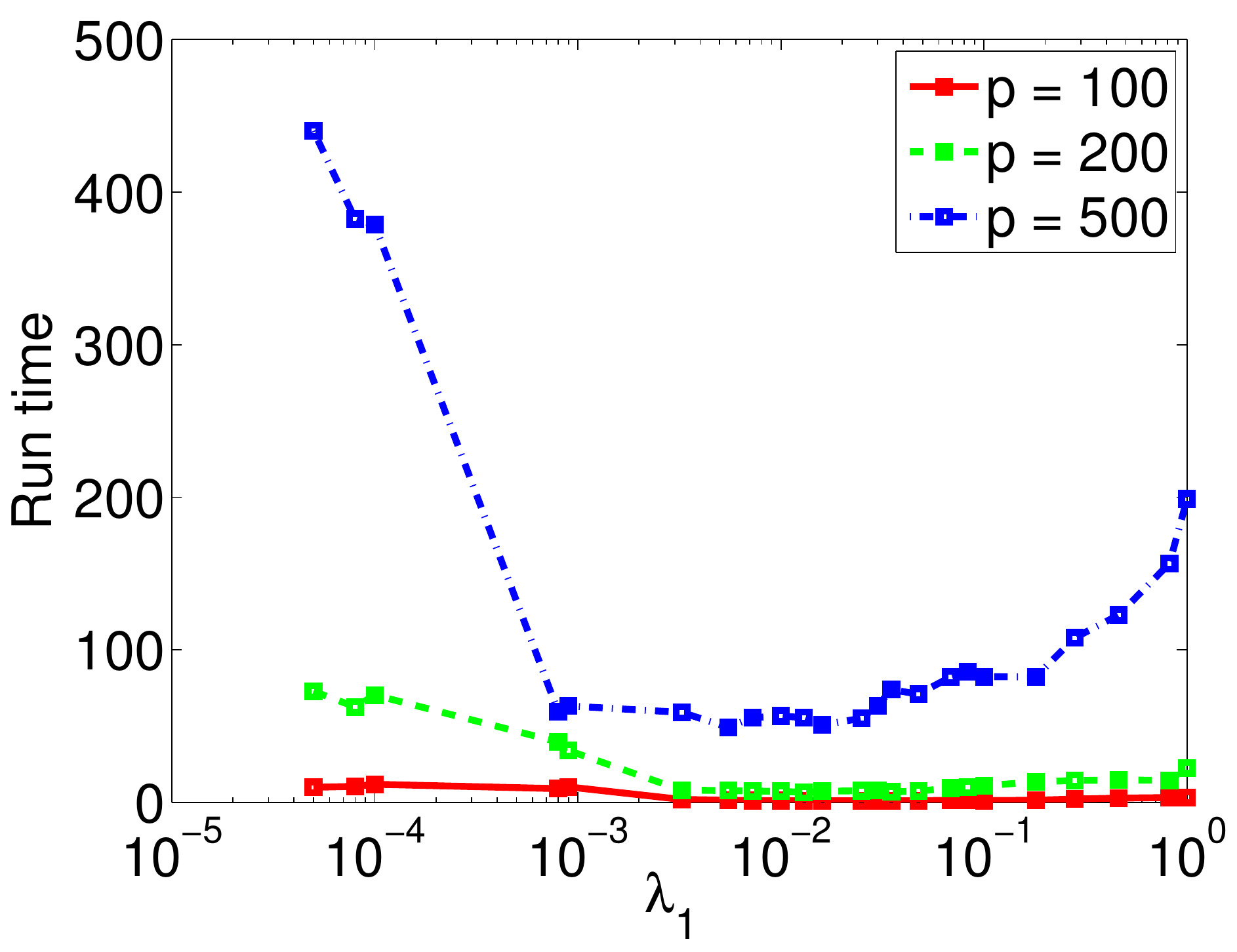}}
\caption{\label{fig:run-times}  \emph{(a):} The total number of iterations for the \CNJGL algorithm, as a function of $\lambda_1$. \emph{(b):} Run-time (in seconds) of the \CNJGL algorithm, as a function of $\lambda_1$.
\emph{(c)-(d)}: As in (a)-(b), but for the \PNJGL algorithm.
All results are averaged over 20 random generations of synthetic data.}
\end{figure}

We now present a more extensive runtime study for the ADMM algorithms for \PNJGL and \CNJGLnospace.
We ran experiments with $p = 100,200,500$ and with $n_1=n_2= p/2$.
We generated synthetic data as described in Section \ref{sec:data}.
Results are displayed in Figures \ref{fig:run-times}(a)-(d), where the panels depict the {run-time} and
{number of iterations} required for the {algorithm to terminate}, as a function of $\lambda_1$, and with $\lambda_2$ fixed. The {number of  iterations} required for {the algorithm to terminate}  is computed as the total number of inner loop iterations performed in Algorithms~\ref{algo:PNJGL_ADMM} and \ref{algo:CNJGL_ADMM}.
From Figures~\ref{fig:run-times}(b) and (d), we observe that as $p$ increases from $100$ to $500$, the run-times increase substantially, but never exceed several minutes.

Figure \ref{fig:run-times}(a) indicates that for CNJGL, the total number of iterations required for {algorithm termination} is small when $\lambda_1$ is small. In contrast, for PNJGL, Figure~\ref{fig:run-times}(c) indicates that the total number of iterations is
large when $\lambda_1$ is small. This phenomenon results from the use of the identity matrix to initialize the network estimates in the ADMM algorithms: when $\lambda_1$ is small, the identity is a poor initialization for PNJGL, but a good initialization for CNJGL (since for CNJGL, $\lambda_2$ induces sparsity  even when $\lambda_1=0$).

\subsubsection{Convergence of the ADMM Algorithm}
Problem (\ref{eq:opt-simple-decouple}) involves two (groups of) primal variables, $X$ and $Y$; in this setting, convergence of ADMM has been established \citep[see, e.g.,][]{ADMMBoyd,Mota-2011}. However, the \PNJGL and \CNJGL optimization problems involve more than two groups of primal variables, and convergence of ADMM in this setting is an ongoing area of research.
 {Indeed, as mentioned in \cite{Eckstein-12}, the standard analysis for ADMM with two groups does not extend in a straightforward way to ADMM with more than two groups of variables}.  \cite{Han-Yuan-2012} and \cite{Luo-2012} show convergence of ADMM with more than two groups of variables under assumptions that do not hold for CNJGL and PNJGL. Under very minimal assumptions, \cite{He-2012} proved that a modified ADMM algorithm (with Gauss-Seidel updates) converges to the optimal solution for problems with any number of groups. More general conditions for convergence of the ADMM algorithm with more than two groups is left as a topic for future work. We also leave for future work a reformulation of the \CNJGL and \PNJGL problems as consensus problems, for which an ADMM algorithm involving two groups of primal variables {can be obtained}, and for which convergence would be guaranteed. 
 {Finally, note that despite the lack of convergence theory, ADMM with more than two groups has been used in practice and often observed to converge faster than other variants. As an example see \cite{TaoYuan-2011}, where their ASALM algorithm (which is the same as ADMM with more than two groups) is reported to be significantly faster than a variant with theoretical convergence.}

\section{Algorithm-Independent Computational Speed-Ups}
\label{sec:comp-improvements}

The ADMM algorithms presented in the previous section work well on problems of moderate size.
In order to solve the PNJGL or CNJGL optimization problems when the number of variables is large, a faster approach  is needed.
We now describe conditions  under which  any algorithm for solving the PNJGL or CNJGL problems can be sped up substantially, for an appropriate range of tuning parameter values.
Our approach mirrors previous results for the graphical lasso
\citep{WittenFriedman11,Mazumder-12}, and FGL and GGL \citep{Danaher2012}.
The idea is simple: if the solutions to  the PNJGL or CNJGL optimization problem are block-diagonal (up to some permutation of the features) with shared support, then we can obtain the global solution to the PNJGL or CNJGL optimization problem by solving the PNJGL or CNJGL problem separately on the features within each block.  This can lead to massive speed-ups. For instance, if the solutions are block-diagonal with $L$ blocks of equal size, then the complexity of our ADMM algorithm reduces from $O(p^3)$ per iteration, to $O((p/L)^3)$ per iteration in each of $L$ independent subproblems. 
Of course, this hinges upon knowing that the  PNJGL or CNJGL solutions are block-diagonal, and knowing the partition of the features into blocks.

In Sections~\ref{sec:nec-PNJGL}-\ref{sec:gen} we derive necessary and  sufficient conditions for the solutions to the PNJGL and CNJGL problems to be block-diagonal.
 Our conditions depend only on the sample covariance matrices $\tS^1,\ldots,\tS^k$ and regularization parameters $\lambda_1, \lambda_2$.
 These conditions can be applied in at most $O(p^2)$ operations.
 In Section~\ref{sec:numer}, we demonstrate the speed-ups that can result from applying these sufficient conditions.

\begin{figure}[h!]
\centering
\includegraphics[width=0.2\linewidth,clip]{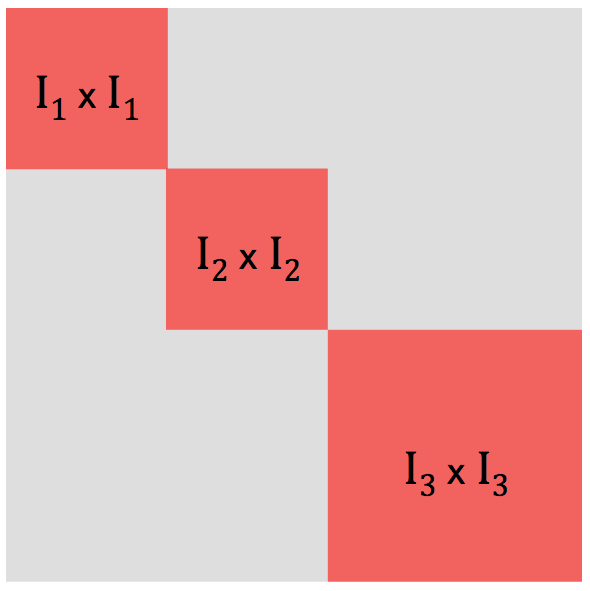}
\caption{ A $p \times p$ matrix is displayed, for which $I_1, I_2, I_3$ denote a partition of the index set $\{1,2,\ldots,p\}$. ${\tT} = \bigcup_{i=1}^L \{ I_i \times I_i \}$ is shown in red, and $\tT^c$ is shown in gray. \label{fig:new}}
\vspace{-3mm}
\end{figure}

 {Related results for the graphical lasso \citep{WittenFriedman11,Mazumder-12} and FGL and GGL \citep{Danaher2012} involve a single condition that is both necessary and sufficient for the solution to be block diagonal. In contrast, in the results derived below, there is a gap between
the necessary and sufficient conditions. Though only the sufficient conditions  are required in order to obtain the computational speed-ups discussed in Section~\ref{sec:numer}, knowing the necessary conditions allows us to get a handle on the tightness (and, consequently, the practical utility) of the sufficient conditions, for a particular value of the tuning parameters. }

 We now introduce some notation that will be used throughout this section.   Let \linebreak[4] $(\tI_1, \tI_2, \ldots, \tI_L)$ be a partition of  the index set $\{1,2,\ldots,p\}$, and
 let
 ${\tT} = \bigcup_{i=1}^L \{ I_i \times I_i \}$.
Define the \emph{support} of a matrix $\bTheta$, denoted by $\mbox{supp}(\bTheta)$, as the set of indices of the non-zero entries in $\bTheta$.
We say $\bTheta$ is supported on $T$ if $\mbox{supp}(\bTheta)\subseteq T$ .
Note that any matrix supported on $\tT$ is block-diagonal subject to some permutation of its rows and columns.
Let $|T|$ denote the cardinality of the set $T$, and let $T^c$ denote the complement of $T$. The scheme is displayed in Figure~\ref{fig:new}.
In what follows we use an $\ell_1/\ell_q$ norm in the RCON penalty, with $q \geq 1$, and let   $\frac{1}{s} + \frac{1}{q} = 1$.

\subsection{Conditions for \PNJGL Formulation to Have Block-Diagonal Solutions}
\label{sec:nec-PNJGL}
In this section, we give necessary conditions and sufficient conditions on the regularization parameters $\lambda_1, \lambda_2$ in the \PNJGL problem (\ref{eqn:sjgl})  so that the resulting precision matrix estimates $\hat{\bTheta}^1,\ldots,\hat{\bTheta}^K$ have a shared block-diagonal structure (up to a permutation of the features).

We first present a necessary condition for $\hat{\bTheta}^1$ and $\hat{\bTheta}^2$  that minimize  (\ref{eqn:sjgl}) with $K=2$ to be block-diagonal.

\begin{theorem} \label{thm:PNJGL_nec}
Suppose that the matrices $\hat{\bTheta}^1$ and $\hat{\bTheta}^2$ that minimize  (\ref{eqn:sjgl}) with $K=2$ have support $\tT$. Then, if $q \geq 1$, it must hold that
\begin{eqnarray}
n_k |S_{ij}^k| \leq \lambda_1 + \lambda_2/2 \; && \forall (i,j) \in T^c,  \;\;\; \mathrm{ \; for \; }k = 1,2, \;  \mathrm{\; and} \label{eq:one}\\
|n_1 S_{ij}^1 + n_2 S_{ij}^2 | \leq 2 \lambda_1 \; && \forall (i,j) \in T^c. \label{eq:two}
\end{eqnarray}
Furthermore, if $q>1$, then it must additionally hold that
\begin{equation}
 \frac{n_k}{|T^c|} \sum_{(i,j) \in T^c} |S_{ij}^k| \leq \lambda_1 + \frac{\lambda_2}{2} \left( \frac{p}{|T^c|} \right) ^{1/s}, \;\; \mathrm{\; for \;} k=1,2. \label{extra}
\end{equation}
\end{theorem}

\begin{remark}
If $|T^c|=O(p^r)$ with $r>1$, then as $p \rightarrow \infty$,  (\ref{extra}) simplifies to
$\frac{n_k}{|T^c|} \sum_{(i,j) \in T^c} |S_{ij}^k|$ \linebreak[4] $\leq \lambda_1$.
\end{remark}

We  now present a sufficient condition for $\hat{\bTheta}^1,\ldots,\hat{\bTheta}^K$  that minimize  (\ref{eqn:sjgl}) to be block-diagonal.
\begin{theorem}
\label{thm:PNJGL_suff}
For $q \geq 1$, a sufficient condition for the matrices $\hat{\bTheta}^1,\ldots,\hat{\bTheta}^K$ that minimize  (\ref{eqn:sjgl}) to each have support $\tT$ is that
$$n_k |S_{ij}^k| \leq \lambda_1 \;\;\; \forall (i,j) \in T^c,  \;\;\; \mathrm{ \; for \; }k = 1,\ldots,K. $$
Furthermore, if $q=1$ and $K=2$, then the necessary conditions
(\ref{eq:one}) and (\ref{eq:two})
are also sufficient.
\end{theorem}
When $q=1$ and $K=2$, then the necessary and sufficient conditions in Theorems \ref{thm:PNJGL_nec} and \ref{thm:PNJGL_suff} are identical, as was previously reported
 in \citet{Danaher2012}.
In contrast, there is a gap between the necessary and sufficient conditions in Theorems \ref{thm:PNJGL_nec} and \ref{thm:PNJGL_suff} when $q > 1$ and $\lambda_2>0$.
 When $\lambda_2=0$, the necessary and sufficient conditions in Theorems \ref{thm:PNJGL_nec} and \ref{thm:PNJGL_suff}  reduce to the results laid out in \cite{WittenFriedman11} for the graphical lasso.

\subsection{Conditions for CNJGL Formulation to Have Block-Diagonal Solutions}
In this section, we give necessary and sufficient conditions on the regularization parameters
 $\lambda_1, \lambda_2$ in the CNJGL optimization problem (\ref{eqn:cnjgl_form})
  so that the resulting precision matrix estimates $\hat{\Theta}^1,\ldots,\hat{\Theta}^K$ have a shared block-diagonal structure (up to a permutation of the features).
\begin{theorem}\label{thm:CNJGL_nec}
Suppose that the matrices $\hat{\bTheta}^1,\hat{\bTheta}^2,\ldots,\hat{\bTheta}^K$  that minimize  (\ref{eqn:cnjgl_form}) have support $T$.
Then, if $q \geq 1$, it must hold that
\begin{equation*}
n_k |S_{ij}^k| \leq \lambda_1 + \lambda_2/2 \;\;\; \forall (i,j) \in T^c,  \;\;\; \mathrm{ \; for \; }k = 1,\ldots,K.
\end{equation*}
Furthermore, if $q>1$, then it must additionally hold that
\begin{equation}
 \frac{n_k}{|T^c|} \sum_{(i,j) \in T^c} |S_{ij}^k| \leq \lambda_1 + \frac{\lambda_2}{2} \left( \frac{p}{|T^c|} \right) ^{1/s}, \;\; \mathrm{\; for \;} k=1,\ldots,K. \label{extra2}
\end{equation}
\end{theorem}

\begin{remark}
If $|T^c|=O(p^r)$ with $r>1$, then as $p \rightarrow \infty$,  (\ref{extra2}) simplifies to
$\frac{n_k}{|T^c|} \sum_{(i,j) \in T^c} |S_{ij}^k|$ \linebreak[4] $\leq \lambda_1$.
\end{remark}
We now present a sufficient condition for $\hat{\bTheta}^1,\hat{\bTheta}^2,\ldots,\hat{\bTheta}^K$  that minimize  (\ref{eqn:cnjgl_form}) to be block-diagonal.
\begin{theorem} \label{thm:CNJGL_suff}
A sufficient condition for $\hat{\bTheta}^1,\hat{\bTheta}^2,\ldots,\hat{\bTheta}^K$  that minimize (\ref{eqn:cnjgl_form}) to have support $\tT$ is that
$$ n_k | S_{ij}^k| \leq \lambda_1 \;\;\; \forall (i,j) \in T^c,  \;\;\; \mathrm{ \; for \; }k = 1,\ldots,K. $$
\end{theorem}
As was the case for the PNJGL formulation, there is a gap between the necessary and sufficient conditions for the estimated precision matrices from the  CNJGL formulation to have a common
block-diagonal support.

\subsection{General Sufficient Conditions} \label{sec:gen}
In this section, we give sufficient conditions for the solution to a general class of optimization problems that include FGL, PNJGL, and CNJGL as special cases to be block-diagonal.
Consider the  optimization problem
\begin{equation} \label{eq:generic_opt}
\Minimize_{\bThetaone,\ldots, \bThetaK \in \psd}  \left\{ \sum_{k=1}^K n_k(-\log\det(\bThetak) + \langle \bThetak, \tS^k \rangle) 
					   +	\sum_{k=1}^K \lambda_1 \|\bThetak\|_1 + \lambda_2 h(\bThetaone,\ldots,\bThetaK) \right\}.
\end{equation}
Once again, let $\tT$ be the support of a $p \times p$ block-diagonal matrix. Let $\bTheta_{\tT}$ denote the restriction of any $p \times p$ matrix $\bTheta$ to $\tT$; that is, $(\bTheta_{\tT})_{ij}= \begin{cases} \bTheta_{ij} & \mbox{if\; } {(i,j) \in T}\\ 0 & \mbox{else} \end{cases}$.
Assume that the function $h$ satisfies  $$h(\bThetaone, \ldots,\bThetaK) > h(\bThetaone_{\tU},\ldots, \bThetaK_{\tU})$$
for any matrices $\bThetaone,\ldots,\bThetaK$ whose support strictly contains $\tU$.

\begin{theorem} \label{thm:generic_opt_suff_cond}
A sufficient condition for the matrices $\hat{\Theta}^1,\ldots,\hat{\Theta}^K$ that solve  (\ref{eq:generic_opt}) to  have support $\tT$ is that
$$ n_k | S_{ij}^k| \leq \lambda_1 \;\;\; \forall (i,j) \in T^c,  \;\;\; \mathrm{ \; for \; }k = 1,\ldots,K. $$
\end{theorem}
Note that this sufficient condition applies to a broad class of regularizers $h$; indeed, the sufficient conditions for PNJGL and CNJGL given in Theorems~\ref{thm:PNJGL_suff} and \ref{thm:CNJGL_suff} are special cases
of Theorem~\ref{thm:generic_opt_suff_cond}. In contrast, the necessary conditions for PNJGL and CNJGL  in Theorems~\ref{thm:PNJGL_nec} and \ref{thm:CNJGL_nec} exploit the specific structure of the RCON penalty.

\subsection{Evaluation of Speed-Ups on Synthetic Data} \label{sec:numer}
 Theorems~\ref{thm:PNJGL_suff} and \ref{thm:CNJGL_suff} provide sufficient conditions for the precision matrix estimates from PNJGL or CNJGL to be block-diagonal with a given support. How can these be used  in order to obtain
 computational speed-ups? We construct a $p \times p$ matrix $A$ with elements 
 \begin{equation*}
 A_{ij} = \begin{cases} 1 & \mathrm{if\; }  i=j \\
  1& \mathrm{ if\; }  n_k |S_{ij}^k| > \lambda_1 \mathrm{\; for \; any \;} k=1,\ldots,K \\
 0 &  \mathrm{else }
 \end{cases}.
 \end{equation*}
 We can then check, in $O(p^2)$ operations, whether $A$ is (subject to some permutation of the rows and columns) block-diagonal, and can also determine the partition of the rows and columns corresponding to the blocks \citep[see, e.g.,][]{Tarjan72}.
 Then, by Theorems~\ref{thm:PNJGL_suff} and \ref{thm:CNJGL_suff}, we can conclude that the PNJGL or CNJGL estimates are block-diagonal, with the same partition of the features into blocks. Inspection of the
 PNJGL and CNJGL optimization problems reveals that we can then solve the problems on the features within each block separately, in order to obtain the global solution to the original PNJGL or CNJGL optimization problems.

We now investigate the speed-ups that result from applying this approach.
We consider the problem of estimating two networks of size $p = 500$.
We create two inverse covariance matrices that are block diagonal with two equally-sized blocks, and sparse within each block. We then generate $n_1=250$ observations from a multivariate normal distribution with the first covariance matrix, and $n_2=250$ observations from a multivariate normal distribution with the second covariance matrix. These observations are used to generate sample covariance matrices  $\tS^1$ and $\tS^2$. We then performed CNJGL and PNJGL with $\lambda_2=1$ and a range of $\lambda_1$ values, with and without the computational speed-ups just described.

\begin{figure}[!htbp]
\begin{center}
$
\begin{array}{ccc}

\begin{subfigure}[]{\includegraphics[width = 0.295\linewidth,clip]{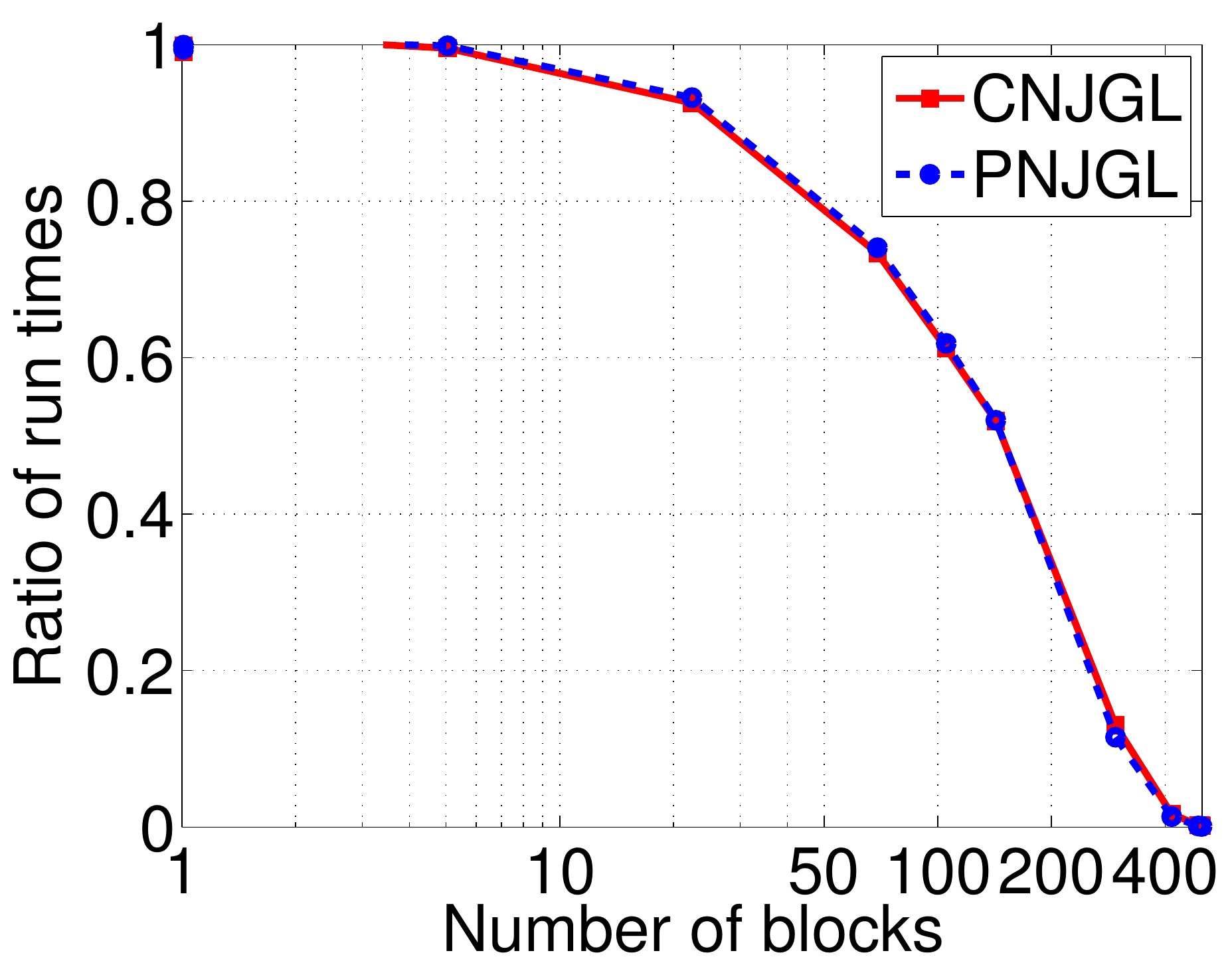}
}
\end{subfigure}

&

\begin{subfigure}[]{\includegraphics[width = 0.295\linewidth,clip]{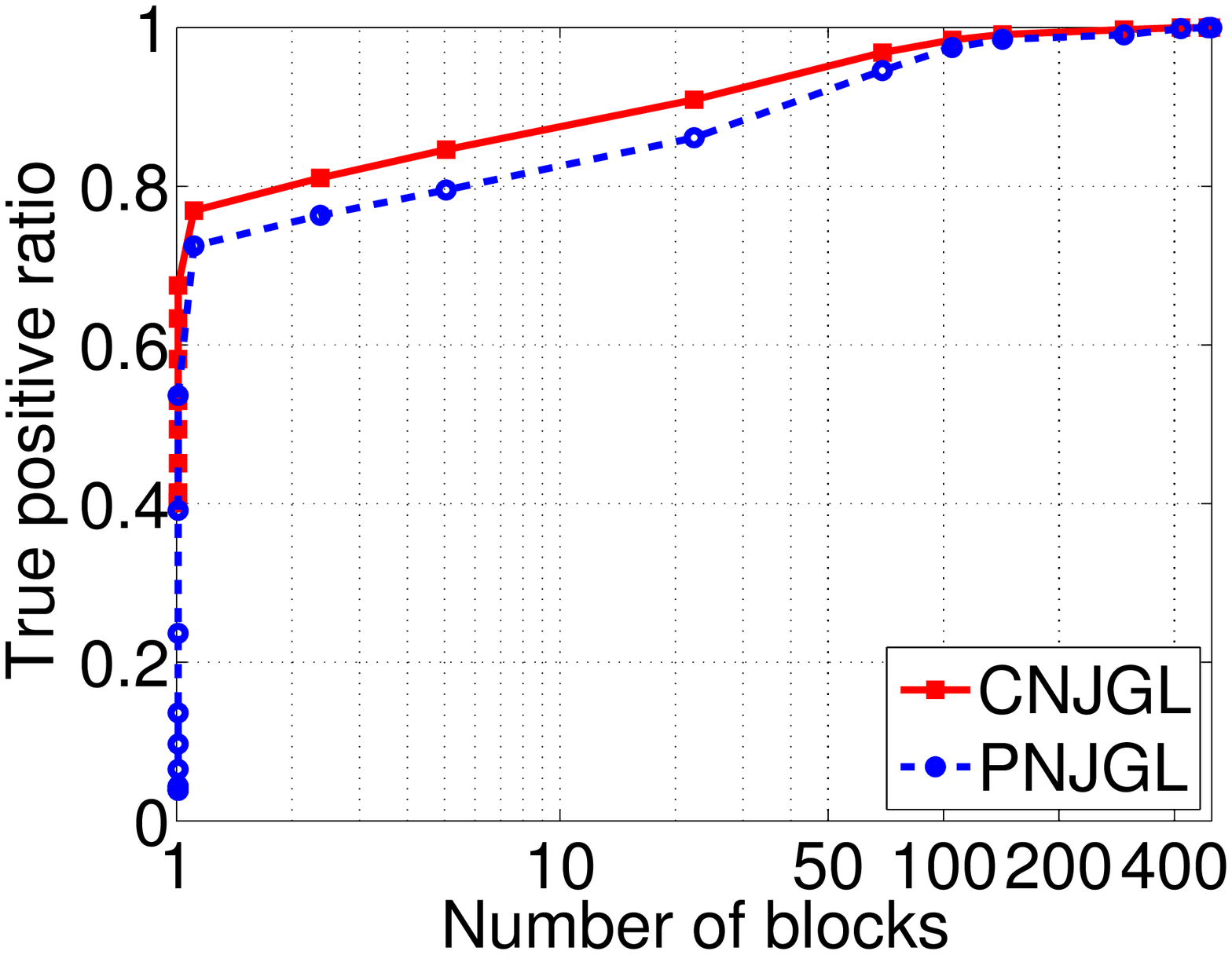}
}
\end{subfigure}

&

\begin{subfigure}[]{\includegraphics[width = 0.295\linewidth,clip]{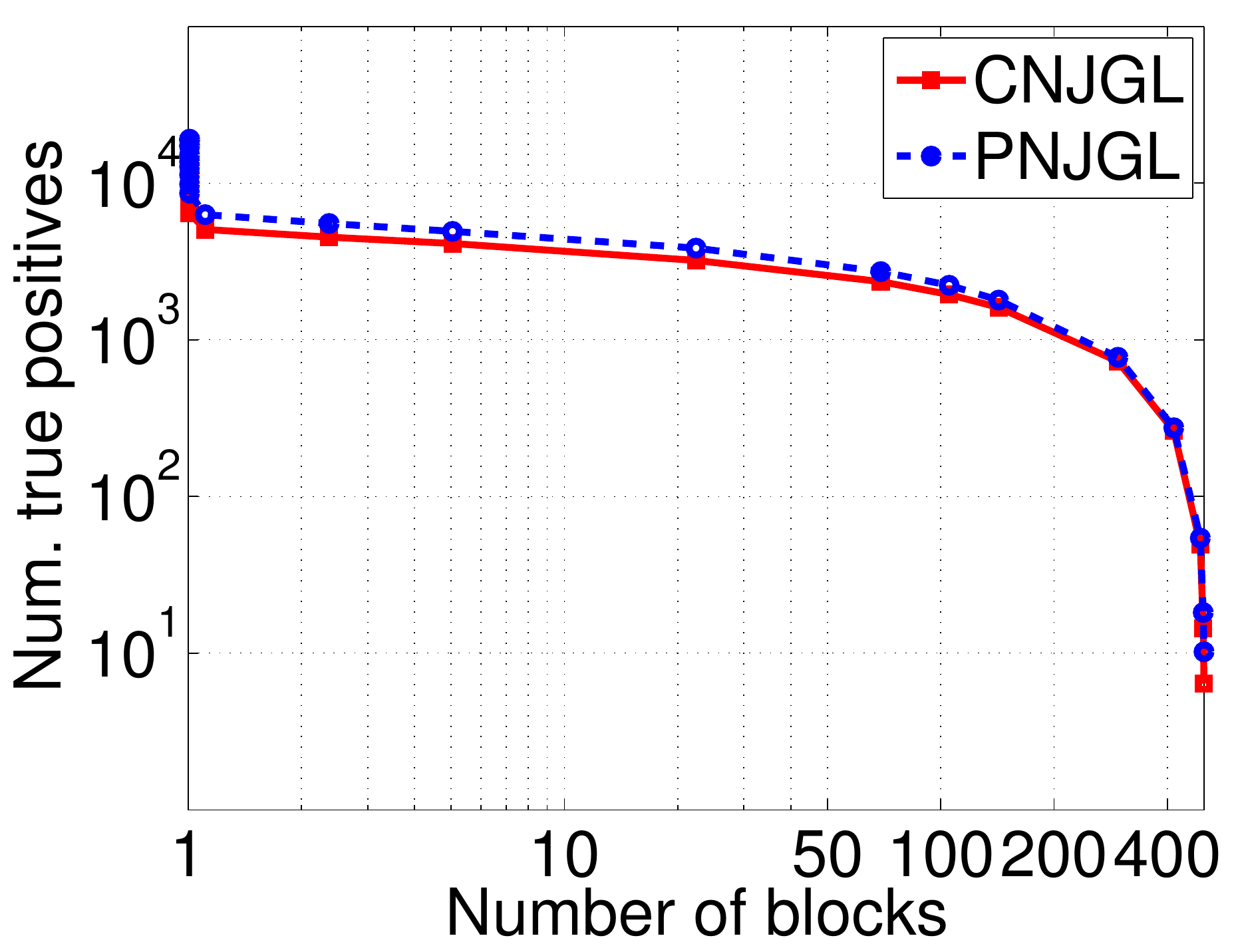}
}
\end{subfigure}

\end{array}
$
\caption{\label{fig:ratio-run-times} Speed-ups for CNJGL and PNJGL on a simulation set-up with $p = 500$ and  $n_1 =   n_2 = 250$. The true inverse covariance matrices  are block-diagonal with two equally-sized sparse blocks. 
The $x$-axis in each panel displays the number of blocks into which the CNJGL or PNJGL problems are decomposed using the sufficient conditions; this is a surrogate for $\lambda_1$. The $y$-axes display
\emph{(a)}:  the ratio of run-times   with and without the sufficient conditions;
 \emph{(b)}: the true positive ratio of the edges estimated; and \emph{(c)}:
the total number of true positive edges estimated. }
\end{center}
\end{figure}

Figure \ref{fig:ratio-run-times} displays the performance of the \CNJGL and \PNJGL formulations, averaged over 20 data sets generated in this way.
In each panel, the $x$-axis shows the number of blocks into which the optimization problems were decomposed using the sufficient conditions; note that this is a surrogate for the value of $\lambda_1$ in the CNJGL or PNJGL optimization problems.
Figure  \ref{fig:ratio-run-times}(a) displays  the  ratio of the run-time taken by the ADMM algorithm when exploiting the sufficient conditions to the run-time when not using the sufficient conditions.
Figure \ref{fig:ratio-run-times}(b) displays the {true-positive ratio}---that is, the ratio of the number of true positive edges in the precision matrix estimates to the total number of edges in the precision matrix estimates.  Figure \ref{fig:ratio-run-times}(c) displays the total number of true positives for the \CNJGL and \PNJGL estimates.   Figure~\ref{fig:ratio-run-times} indicates that the sufficient conditions detailed in this section lead to substantial computational improvements.

\section{Simulation Study}
\label{sec:data}

In this section, we present the results of a simulation study demonstrating the empirical performance of PNJGL and CNJGL.

\subsection{Data Generation}
{In the simulation study, we generated two synthetic networks (either Erdos-Renyi, scale-free, or community), each of which contains  a common set of $p$ nodes. 
 Four of the $p$ nodes were then modified in order to create two perturbed nodes and two co-hub nodes. Details are provided in Sections~\ref{one}-\ref{three}.}

\subsubsection{{Data Generation for Erdos-Renyi Network}} \label{one}

We generated the data as follows, for $p=100$, and $n \in \{25,50,100,200\}$: \\
 
\begin{list}{}{}
\item{\textbf{{Step 1:}}} To generate an Erdos-Renyi network, we  created a $p \times p$ symmetric matrix $A$ with elements  
 $${A}_{ij} \sim_{\mathrm{i.i.d.}} \left\{\begin{array}{cc} 0 & \mbox{ with probability } 0.98, \\
 \mathrm{Unif}([-0.6,-0.3] \cup [0.3,0.6]) & \mbox{ otherwise. } \end{array} \right.$$ 
\item{\textbf{Step 2:}} We  duplicated $A$ into two matrices, ${A}^1$ and ${A}^2$. We selected two nodes at random, and for each node, we set the elements of the corresponding row and column of either ${A}^1$ or ${A}^2$ (chosen at random) to be i.i.d. draws from a $\mathrm{Unif}([-0.6,-0.3] \cup [0.3,0.6])$ distribution. This results in two perturbed nodes. 
\item{\textbf{Step 3:}} We randomly selected two nodes to serve as co-hub nodes, and set each element of the corresponding rows and columns in each network to be i.i.d. draws from a $\mathrm{Unif}([-0.6,-0.3] \cup [0.3,0.6])$ distribution. 
In other words, these co-hub nodes are \emph{identical} across the two networks. 

\item{\textbf{Step 4:}}
In order to make the matrices  positive definite, we let $c = $ \linebreak[4] $\min(\lambda_{\min}({A}^1),\lambda_{\min}({A}^2))$, where $\lambda_{\min}(\cdot)$ indicates the smallest eigenvalue of the matrix. We then set ${({\boldsymbol\Sigma}^1)^{-1}}$ equal to ${A}^1 + (0.1+|c|) {I}$ and set ${({\boldsymbol\Sigma}^2)^{-1}}$ equal to ${A}^2 + (0.1+|c|) {I}$, where  $I$ is the $p \times p$ identity matrix. 

\item{\bf Step 5:} We generated $n$ independent observations each from a $N({0}, {\boldsymbol\Sigma}^1)$ and a $N({0}, {\boldsymbol\Sigma}^2)$ distribution, and used them to compute the sample covariance matrices ${S}^1$ and ${S}^2$.   \end{list} 

\subsubsection{Data Generation for Scale-free Network} \label{two}
The data generation proceeded as in Section~\ref{one}, except that Step 1 was modified:
\begin{list}{}{}
\item{\bf Step 1:} We used the \verb=SFNG= functions in \verb=Matlab= \citep{SFNG-07} with parameters \verb+mlinks=2+ and \verb+seed=1+ to generate a scale-free network with $p$ nodes. We then created a $p \times p$ symmetric matrix $A$
that has non-zero elements only for the edges in the scale-free network. These non-zero elements were generated i.i.d. from a $\mathrm{Unif}([-0.6,-0.3] \cup [0.3,0.6])$ distribution. 
\end{list}
Steps 2-5 proceeded as in Section~\ref{one}.

\subsubsection{Data Generation for Community Network} \label{three}
 
We generated data as in Section~\ref{one}, except for one modification: at the end of Step 3, we set the [1:40, 61:100] and [61:100, 1:40] submatrices of $A^1$ and $A^2$ equal to zero.

Then $A^1$ and $A^2$ have non-zero entries concentrated in the top and bottom $60\times 60$ principal
submatrices. These two submatrices correspond to two communities. Twenty nodes overlap between the two communities. 

\subsection{Results}

\begin{table}
\begin{tabular}{|p{4mm}|p{141mm}|}

\hline 
$ $ 

(1) 

&  
$ $

\noindent
\underline{{Positive edges}}:  

$\hspace{1mm}  \sum_{i<j} \left( \mathbf{1}   \{ |\hat{\Theta}_{ij}^1| > t_0   \}  + \mathbf{1}    \{  |\hat{\Theta}_{ij}^2| > t_0 \}  \right)$

\noindent
\underline{{True positive edges}}: 

$\hspace{1mm}  \sum_{i< j} \left(   \mathbf{1} \{   |\bThetaone_{ij}| > t_0 \hspace{2mm} \mbox{and} \hspace{2mm} |\hat{\Theta}_{ij}^1| > t_0 \}   
+  \mathbf{1} \{  |\bThetatwo_{ij}| > t_0 \hspace{2mm} \mbox{and} \hspace{2mm} |\hat{\Theta}_{ij}^2| > t_0 \} \right)$

\\

\hline 
$ $

 (2) 
 & 
  $ $
  
\underline{{Positive perturbed columns}} (PPC): \\
 & PNJGL: $ \sum_{i = 1}^p \mathbf{1}    \left\{   \|\hat{ V }_{-i,i}\|_2 > t_{s}  \right\} $;   FGL/GL: $\hspace{1mm}  \sum_{i = 1}^p \mathbf{1}    \{   \|{( \hatvarone - \hatvartwo)}_{-i,i}\|_2 > t_{s}  \} $\\
&\\
& \underline{True positive perturbed columns} (TPPC):\\
& PNJGL: $\hspace{1mm} \sum_{i \in I_p} \mathbf{1}   \{   \|\hat{ V }_{-i,i}\|_2 > t_{s}\}$;  FGL/GL: $\hspace{1mm} \sum_{i \in I_p} \mathbf{1}   \{   \|{( \hatvarone - \hatvartwo)}_{-i,i}\|_2 > t_{s}\},$  \\

& {where $I_P$ is the set of  perturbed column indices.}\\
 
&\\

& \underline{{Positive co-hub columns}} (PCC):  \\
& CNJGL: $\hspace{0.2mm} \sum_{i = 1}^p \mathbf{1}   \left\{   \|\hat{V}^1_{-i,i}\|_2 > t_{s} \hspace{0.4mm}\mbox{and} \hspace{0.4mm}  \|\hat{V}^2_{-i,i}\|_2  >  t_{s}    \right\} $; \\
& GGL/GL:  $\hspace{0.1mm} \sum_{i = 1}^p \mathbf{1}   \left\{   \| {\hatvarone}_{-i,i}\|_2 > t_{s} \hspace{0.4mm} \mbox{and} \hspace{0.4mm}  \|{\hatvartwo}_{-i,i}\|_2  >  t_{s}    \right\} $\\

& \\
& \underline{{True positive co-hub columns}} (TPCC): \\
&CNJGL: $\hspace{0.2mm} \sum_{i  \in I_c} \mathbf{1}   \left\{   \|\hat{V}^1_{-i,i}\|_2 > t_{s} \hspace{0.4mm} \mbox{and} \hspace{0.3mm} \|\hat{V}^2_{-i,i}\|_2  >  t_{s}  \right\}$; \\
&GGL/GL: $\hspace{0.1mm} \sum_{i  \in I_c} \mathbf{1}   \left\{    \|{\hatvarone}_{-i,i}\|_2 > t_{s} \hspace{0.4mm} \mbox{and} \hspace{0.3mm} \|{\hatvartwo}_{-i,i}\|_2  >  t_{s}\right\},$ \\
&  {where $I_C$ is the set of co-hub column indices.}

\\

\hline 

(3) &
\underline{{Error}}:
$\sqrt{  \sum_{i<j} (\bThetaone_{ij} - \hatvarone_{ij})^2  } + \sqrt{
\sum_{i<j} (\bThetatwo_{ij} - \hatvartwo_{ij})^2  }$ 
 \\

\hline
\end{tabular}
\caption{Metrics used to quantify algorithm performance. Here $\bThetaone$ and $\bThetatwo$ denote the true inverse covariance matrices, and ${\hatvarone}$ and ${\hatvartwo}$ denote the two estimated inverse covariance matrices.
Here ${\bf 1}\{A\}$ is an indicator variable that equals one if the event $A$ holds, and equals zero otherwise. (1) Metrics based on recovery of the support of $\bThetaone$ and $\bThetatwo$. Here $t_0=10^{-6}$.  (2) Metrics based on identification of perturbed nodes and co-hub nodes. The metrics PPC and TPPC quantify node perturbation, and  are applied to PNJGL, FGL, and GL. The metrics PCC and TPCC relate  to co-hub detection, and are applied to CNJGL, GGL, and GL. We let $t_{s} = \mu + 5.5\sigma$, where $\mu$ is the mean and $\sigma$ is the standard deviation of  $\{ \| \hat{V}_{-i,i} \|_2 \}_{i=1}^p$ (PPC or TPPC for PNJGL), 
$\{ \| (\hat{\bTheta}^1-\hat{\bTheta}^2)_{-i,i} \|_2 \}_{i=1}^p$ (PPC or TPPC for FGL/GL), $\{ \| \hat{V}^1_{-i,i} \|_2 \}_{i=1}^p$ and $\{ \| \hat{V}^2_{-i,i} \|_2 \}_{i=1}^p$ (PPC or TPPC for CNJGL), or $\{ \| \hat{\bTheta}^1_{-i,i} \|_2 \}_{i=1}^p$ and $\{ \| \hat{\bTheta}^2_{-i,i} \|_2 \}_{i=1}^p$ (PPC or TPPC for GGL/GL).  However, results are very insensitive to the value  of $t_s$, as is shown in Appendix~\ref{sec:Norm_Plots}.  (3) Frobenius error of estimation of $\bThetaone$ and $\bThetatwo$. }
\label{tbl:TableMetrics}
\end{table}

We now define several metrics used to measure algorithm performance. We wish to quantify each algorithm's (1) recovery of the support of the true inverse covariance matrices, (2) successful detection of co-hub and perturbed nodes, and (3) error in estimation of $\bThetaone = ({\boldsymbol \Sigma}^1)^{-1}$ and $\bThetatwo = ({\boldsymbol\Sigma}_2)^{-1}$.  Details are given in Table~\ref{tbl:TableMetrics}. These metrics are discussed further in Appendix \ref{sec:Norm_Plots}.

We compared the performance of PNJGL to its edge-based counterpart FGL, as well as to graphical lasso (GL). We  compared the performance of CNJGL to GGL and GL.
We expect CNJGL to be able to detect  co-hub nodes (and, to a lesser extent, perturbed nodes), and we expect PNJGL to be able to detect  perturbed nodes.  (The co-hub nodes will not be detected by PNJGL, since they are identical across the networks.)

\begin{figure}
\noindent
\makebox[1\linewidth][l]{\hspace{\dimexpr-\fboxsep-\fboxrule\relax}
\fbox {\parbox{0.97\linewidth}{ 

\fontsize{8}{12}\selectfont 
\noindent
\textcolor{color1}{-  - $\mathsmaller{\times}$ - -}  \hspace{0.001mm} PNJGL $\lambda_2 = 0.3n$
\hspace{2.2mm} \textcolor{color3}{-  - $\mathsmaller{\mathsmaller{\triangle}}$  - -}  \hspace{0.001mm} PNJGL $\lambda_2 = 1.0n$
\hspace{2.2mm} \textcolor{color5}{$\cdot$  - $\mathsmaller{\times}$ - $\cdot$}  \hspace{0.001mm} FGL $\lambda_2 = 0.01n$
\hspace{2.2mm} \textcolor{color7}{$\cdot$  - $\mathsmaller{\mathsmaller{\triangle}}$  - $\cdot$} \hspace{0.001mm} FGL $\lambda_2 = 0.05n$
\hspace{2mm} \textcolor{color9}{$\cdot$ -  $\mathsmaller{\mathsmaller{\bigtriangledown}}$  - $\cdot$} \hspace{0.001mm} GL\\

\fontsize{8}{12}\selectfont
\textcolor{color2}{-  - $\mathsmaller{+}$ - -} \hspace{0.001mm} PNJGL $\lambda_2 = 0.5n$ 
\hspace{2.2mm} \textcolor{color4}{-  - $\ast$ - -} \hspace{0.001mm} PNJGL $\lambda_2 = 2.0n$
\hspace{2.2mm} \textcolor{color6}{$\cdot$  - $\mathsmaller{+}$ - $\cdot$} \hspace{0.001mm} FGL $\lambda_2 = 0.03n$
\hspace{2.2mm} \textcolor{color8}{$\cdot$  - $\ast$ - $\cdot$} \hspace{0.001mm} FGL $\lambda_2 = 0.1n$ 
\fontsize{10}{12}\selectfont 
}}} \\ 

\centering
(a) $n = 25$ \\
\includegraphics[width= 0.32\linewidth,clip]{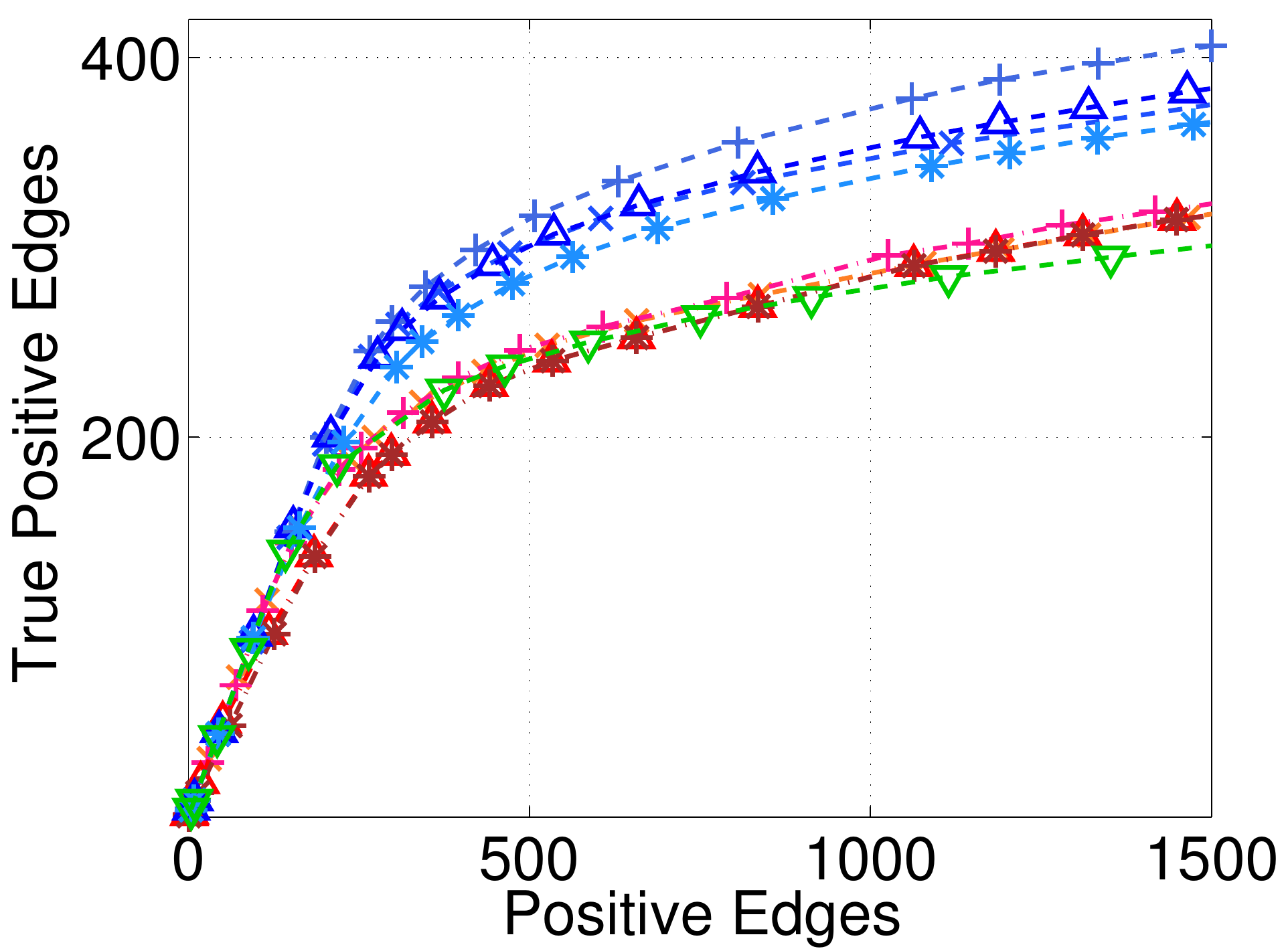}
\includegraphics[width= 0.32\linewidth,clip]{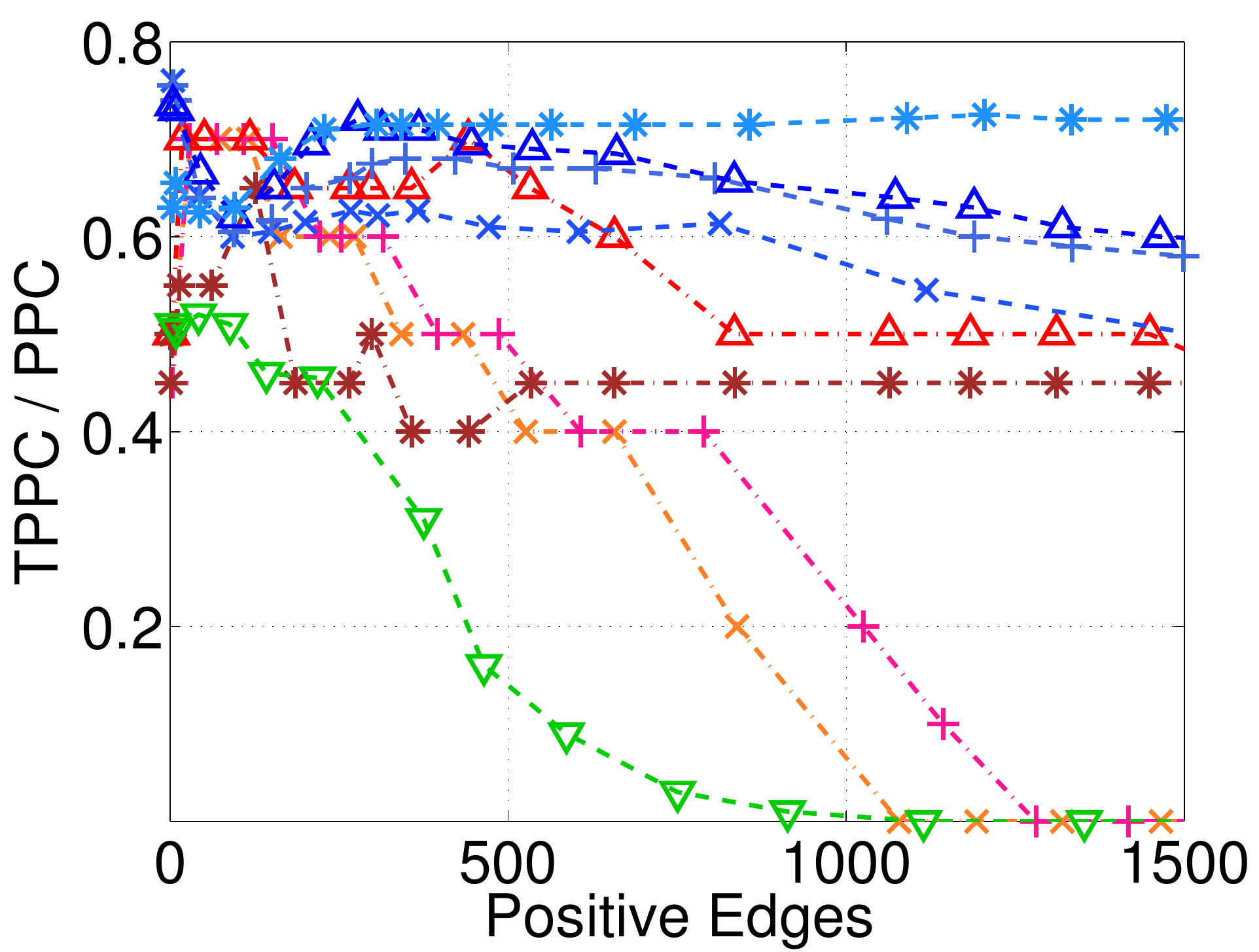}
\includegraphics[width= 0.32\linewidth,clip]{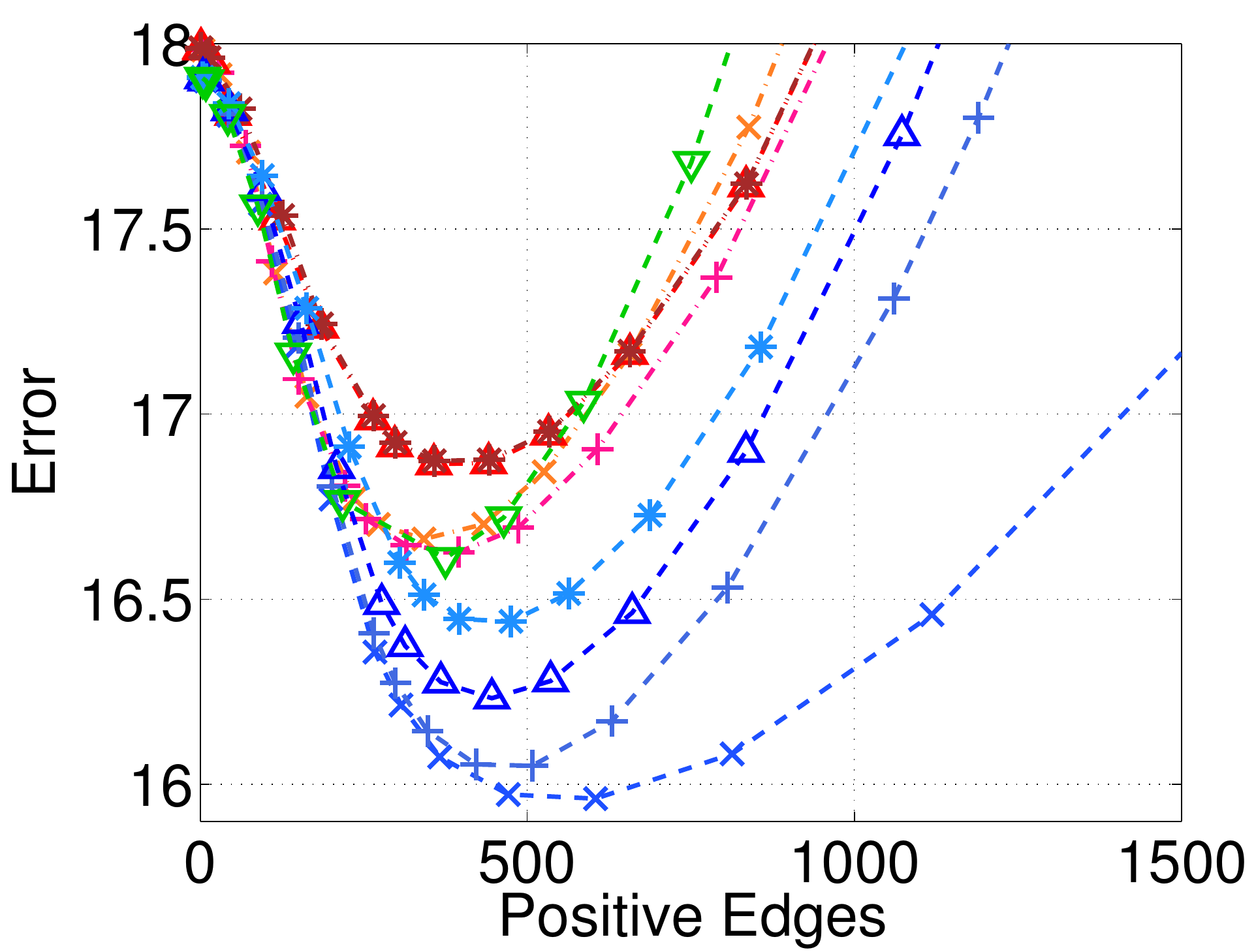} \\
(b) $n = 50$ \\
\includegraphics[width= 0.32\linewidth,clip]{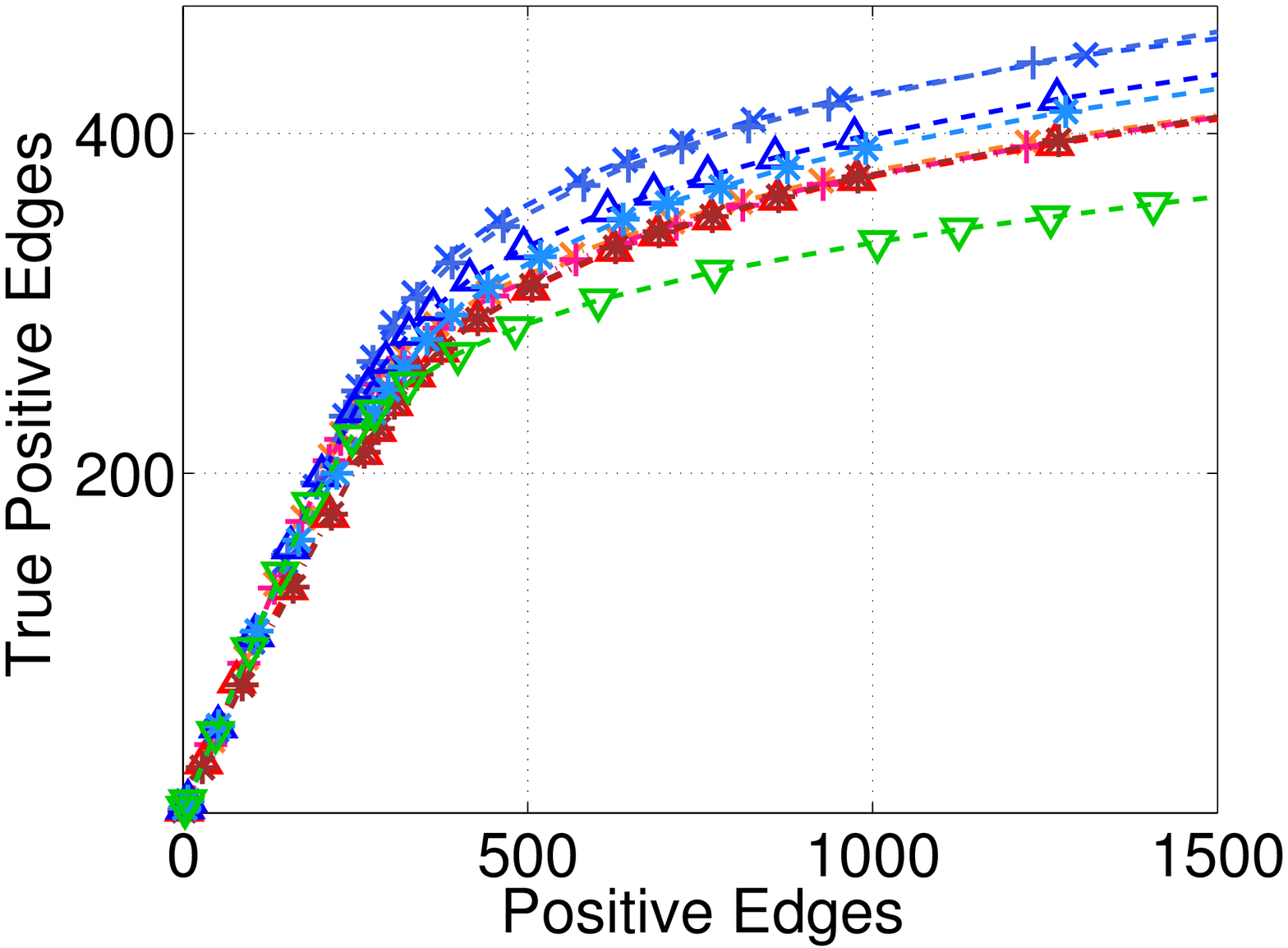}
\includegraphics[width= 0.32\linewidth,clip]{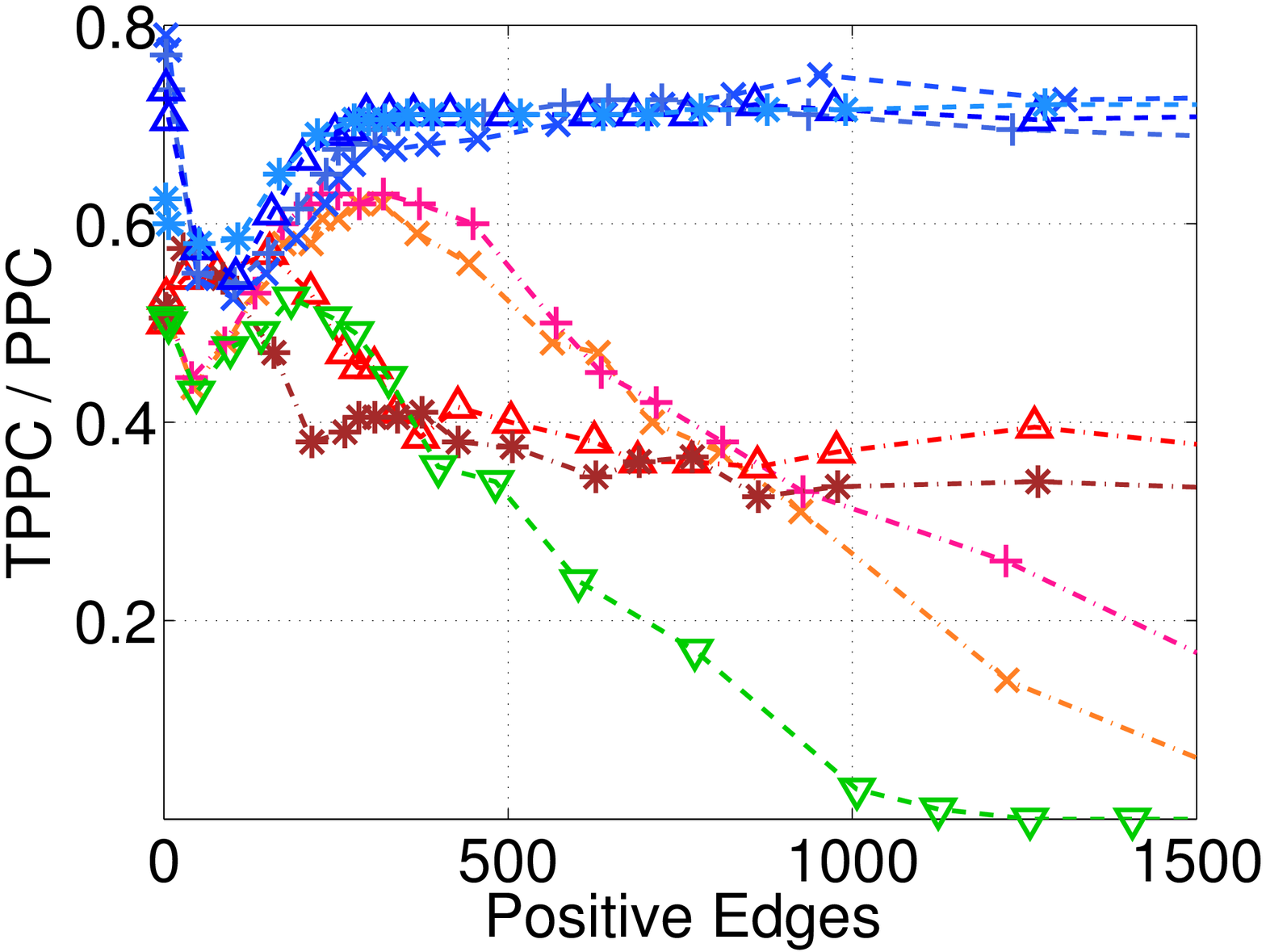}
\includegraphics[width= 0.32\linewidth,clip]{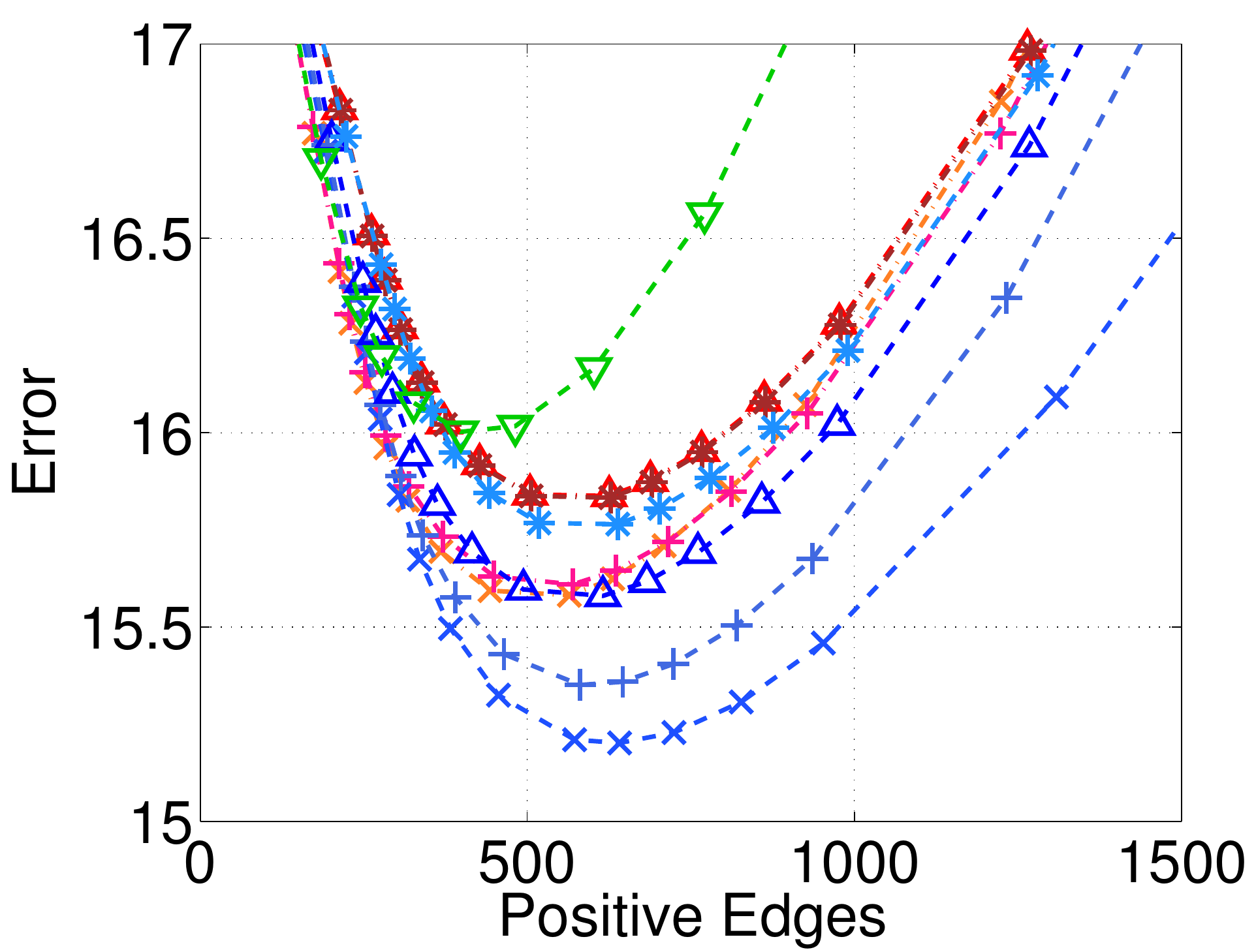} \\
(c) $n = 100$ \\
\includegraphics[width= 0.32\linewidth,clip]{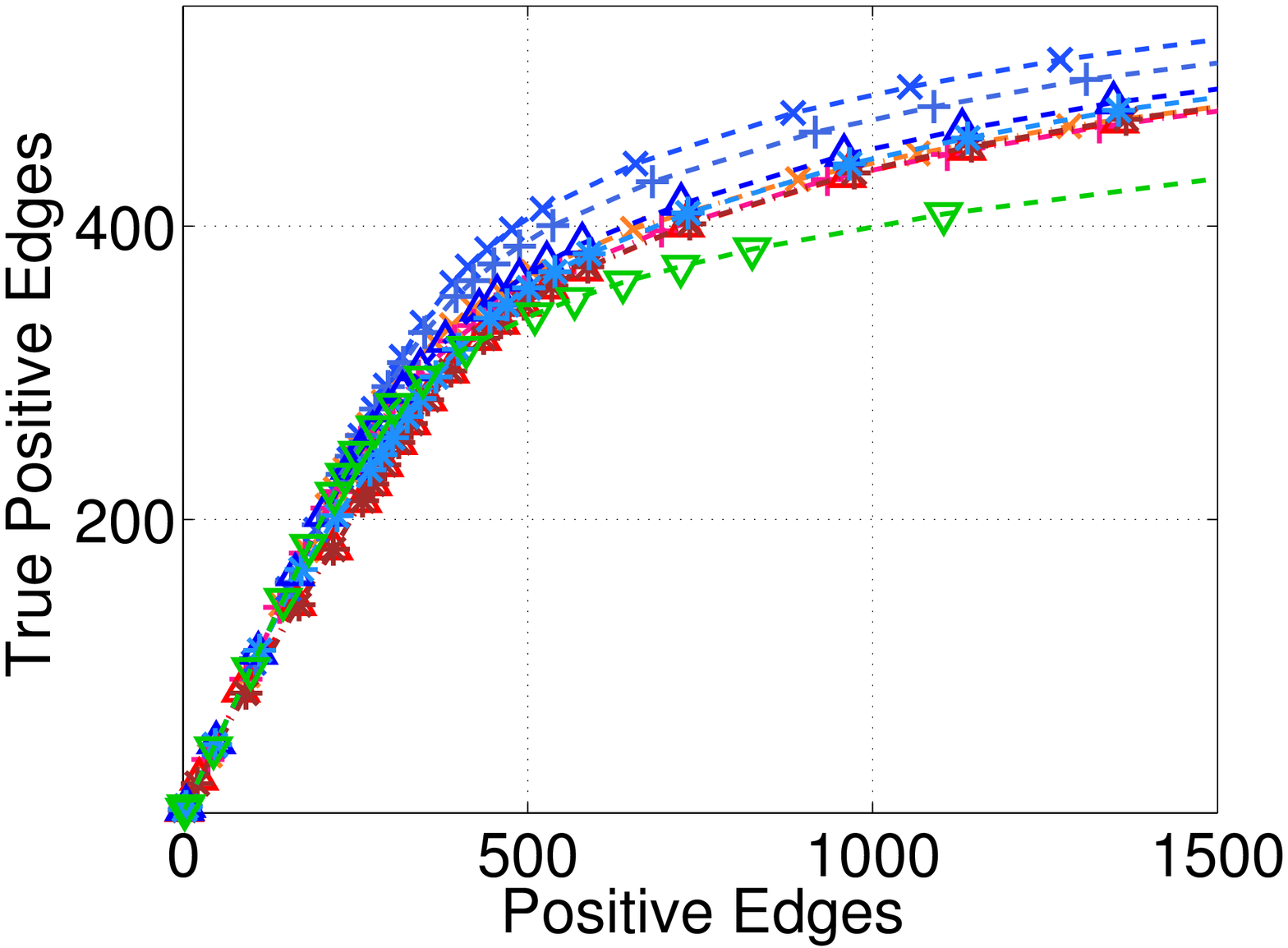}
\includegraphics[width= 0.32\linewidth,clip]{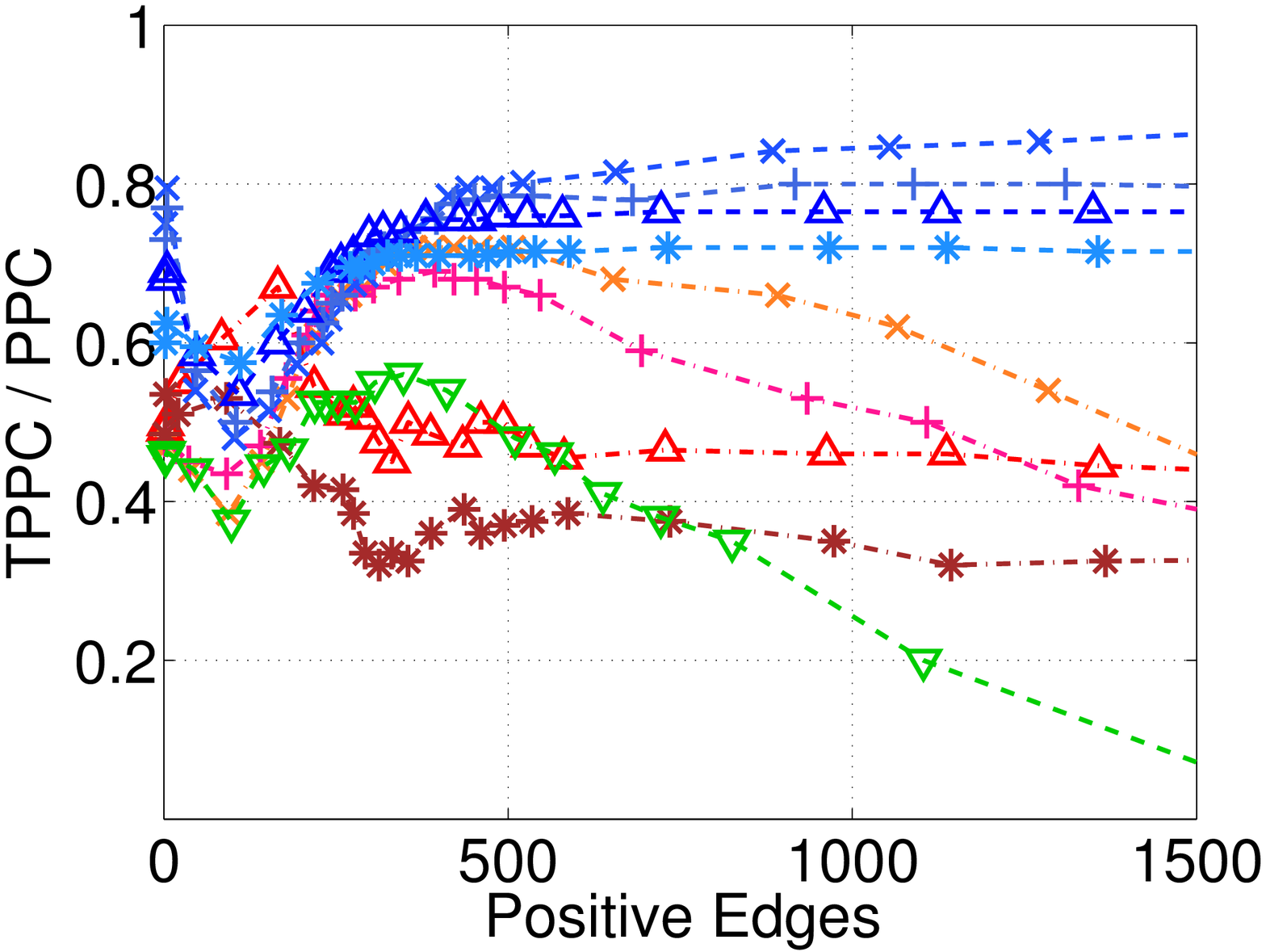}
\includegraphics[width= 0.32\linewidth,clip]{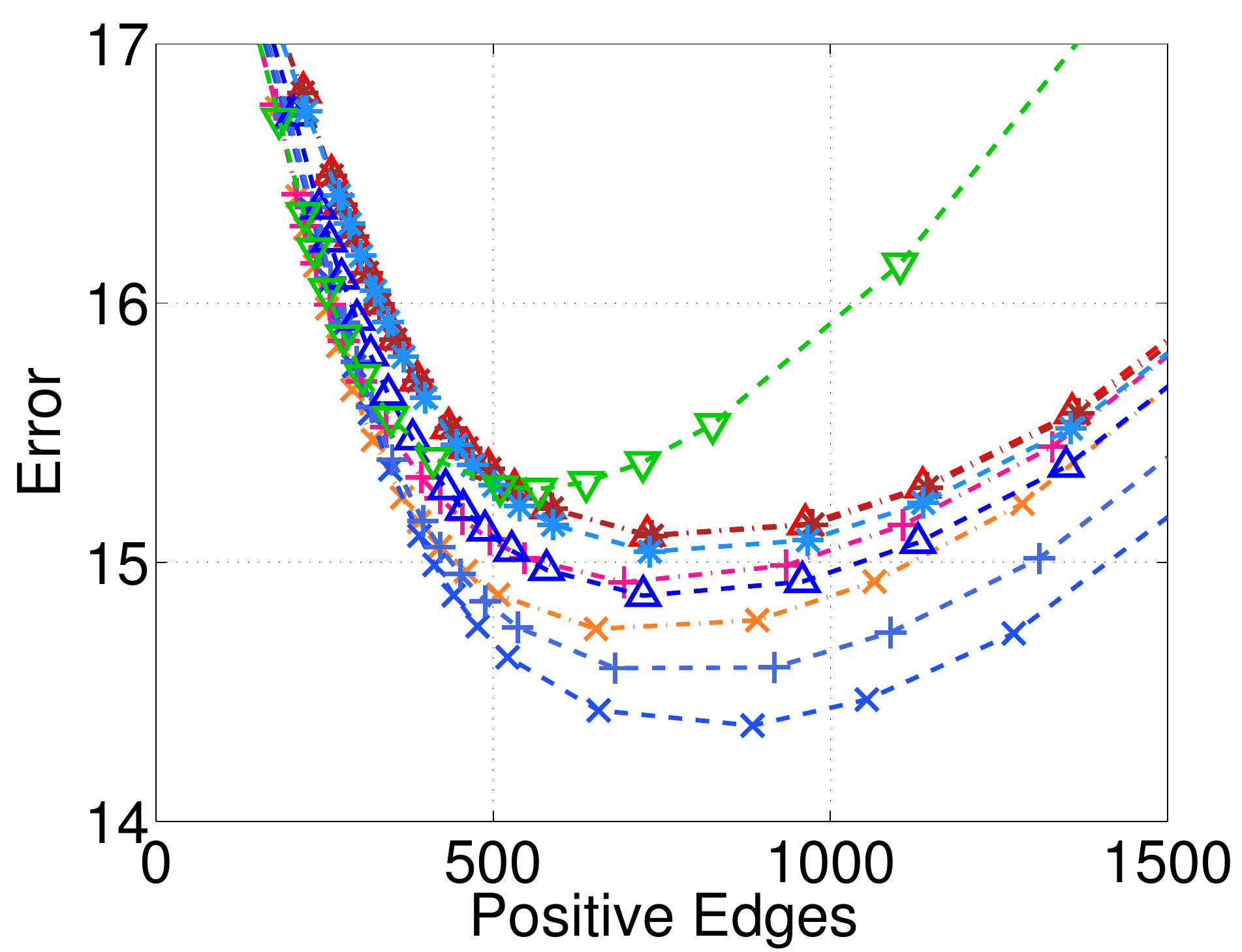} \\
(d) $n = 200$ \\
\includegraphics[width= 0.32\linewidth,clip]{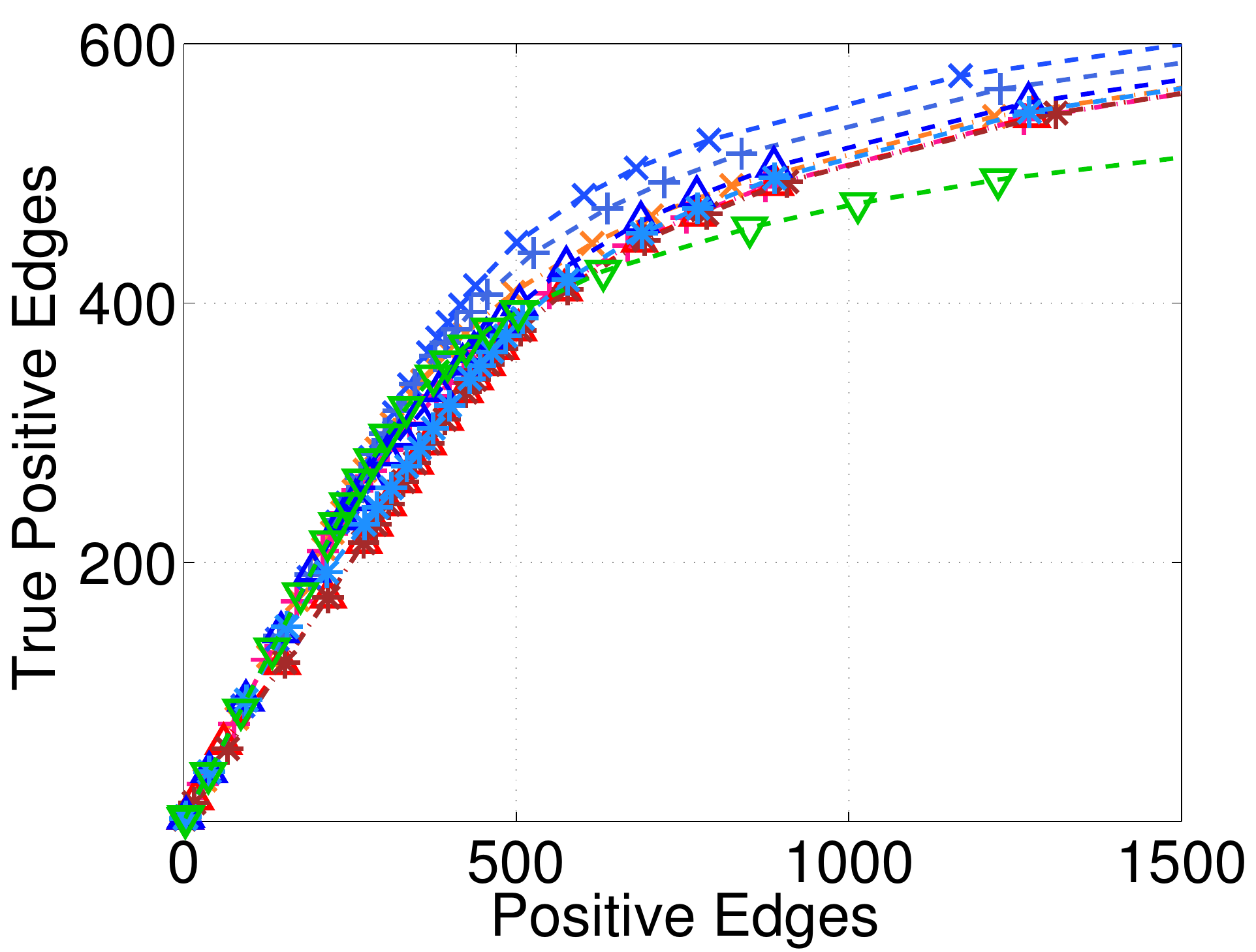}
\includegraphics[width= 0.32\linewidth,clip]{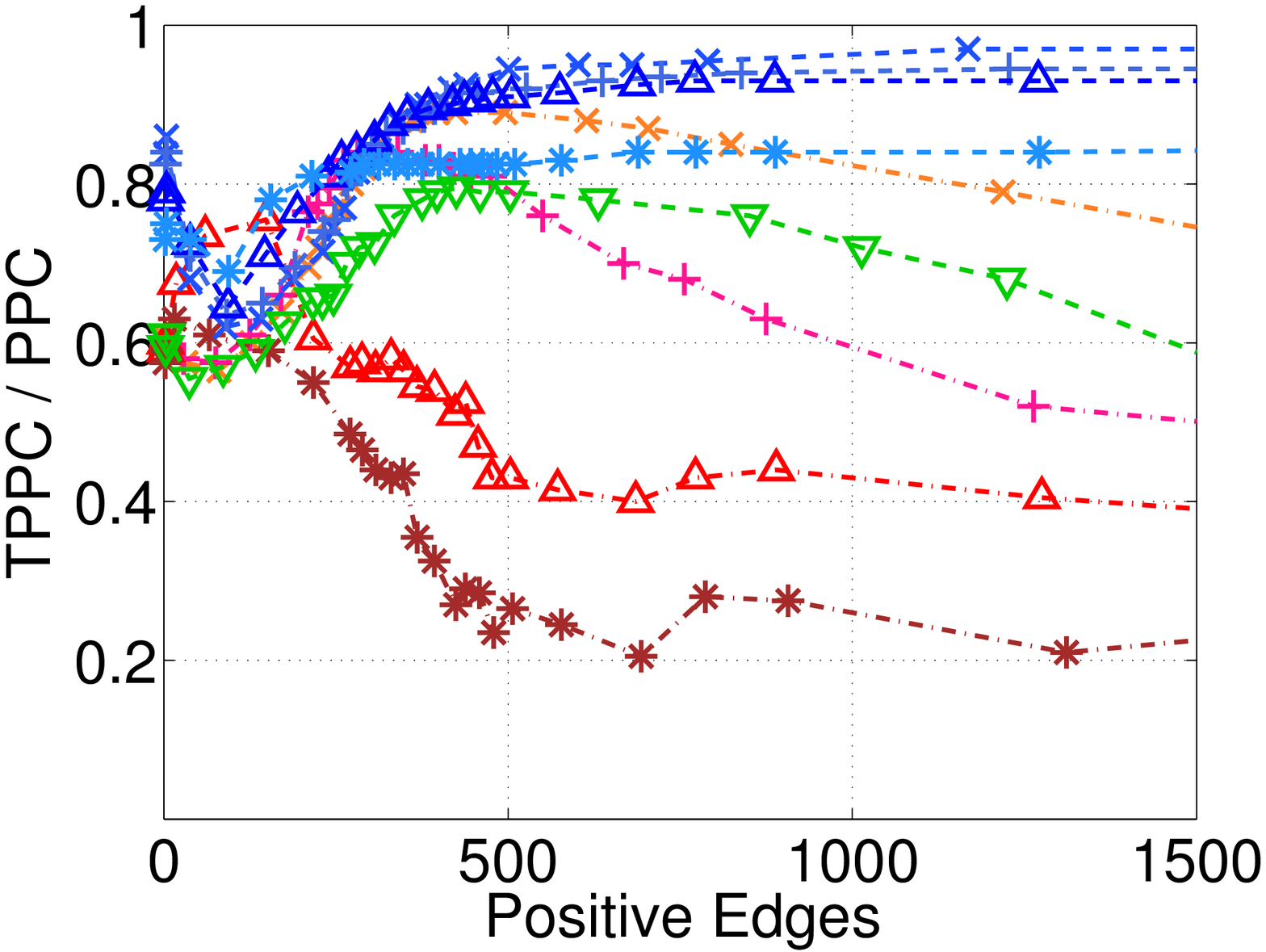}
\includegraphics[width= 0.32\linewidth,clip]{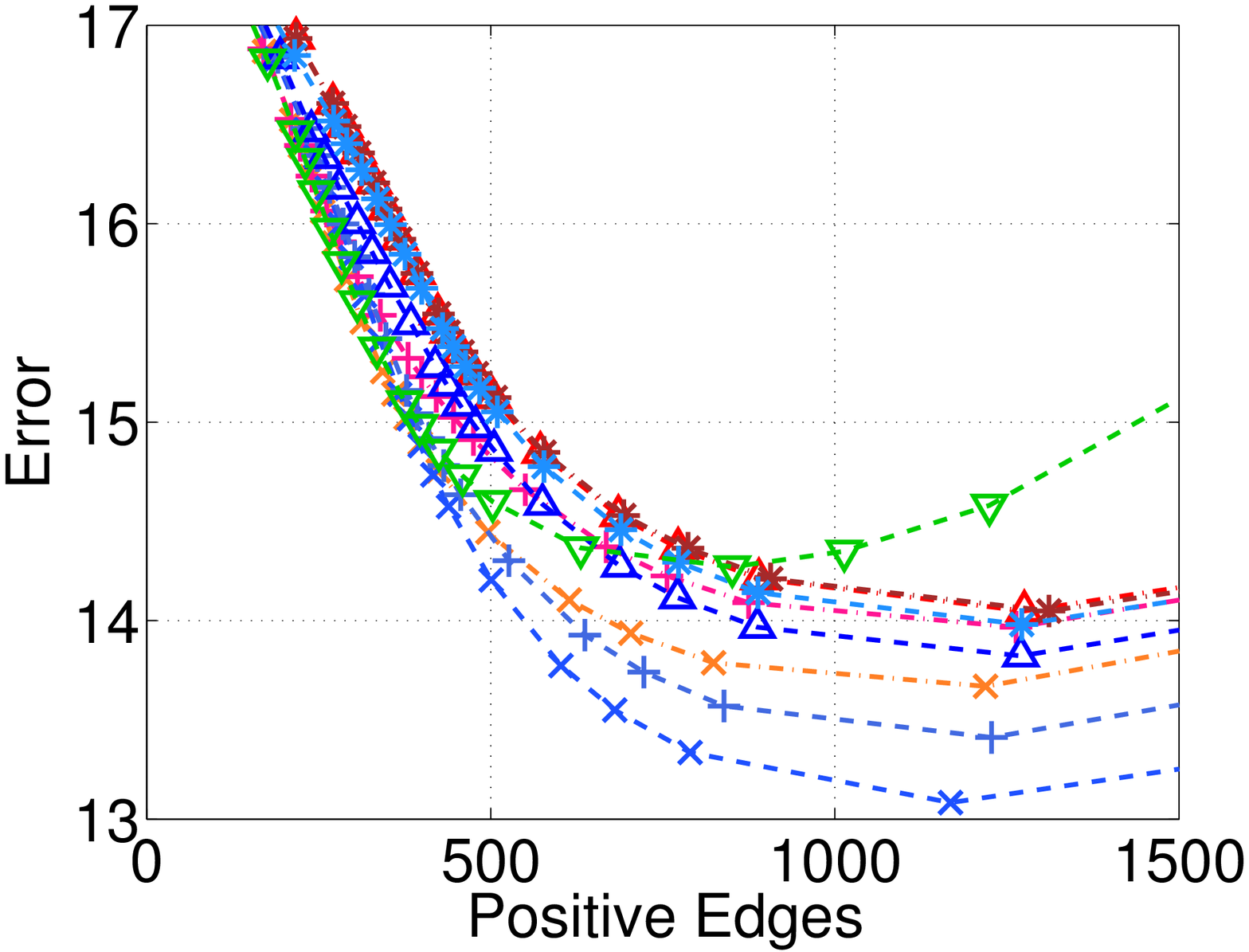}

\caption{\label{fig:PNJGL} Simulation results on Erdos-Renyi network  (Section~\ref{one}) for PNJGL with $q = 2$, FGL, and GL, for \emph{(a):} $n = 25$, \emph{(b):} $n = 50$, \emph{(c):} $n = 100$, \emph{(d):} $n = 200$, when $p = 100$. Each colored line corresponds to a fixed value of $\lambda_2$, as $\lambda_1$ is varied. Axes are described in detail in Table \ref{tbl:TableMetrics}.
Results are averaged over 100 random generations of the data.} 
\vspace{-2mm}
\end{figure}

\begin{figure}

\noindent
\makebox[1\linewidth][l]{\hspace{\dimexpr-\fboxsep-\fboxrule\relax}
\fbox {\parbox{0.97\linewidth}{ 

\fontsize{8}{12}\selectfont 
\noindent
\textcolor{color1}{-  - $\mathsmaller{\times}$ - -}  \hspace{0.001mm} CNJGL $\lambda_2 = 0.3n$
\hspace{2.2mm} \textcolor{color3}{-  - $\mathsmaller{\mathsmaller{\triangle}}$  - -}  \hspace{0.001mm} CNJGL $\lambda_2 = 1.0n$
\hspace{2.2mm} \textcolor{color5}{$\cdot$  - $\mathsmaller{\times}$ - $\cdot$}  \hspace{0.001mm} GGL $\lambda_2 = 0.01n$
\hspace{2.2mm} \textcolor{color7}{$\cdot$  - $\mathsmaller{\mathsmaller{\triangle}}$  - $\cdot$} \hspace{0.001mm} GGL $\lambda_2 = 0.05n$
\hspace{2mm} \textcolor{color9}{$\cdot$ -  $\mathsmaller{\mathsmaller{\bigtriangledown}}$  - $\cdot$} \hspace{0.001mm} GL\\

\fontsize{8}{12}\selectfont
\textcolor{color2}{-  - $\mathsmaller{+}$ - -} \hspace{0.001mm} CNJGL $\lambda_2 = 0.6n$ 
\hspace{2.2mm} \textcolor{color4}{-  - $\ast$ - -} \hspace{0.001mm} CNJGL $\lambda_2 = 1.5n$
\hspace{2.2mm} \textcolor{color6}{$\cdot$  - $\mathsmaller{+}$ - $\cdot$} \hspace{0.001mm} GGL $\lambda_2 = 0.03n$
\hspace{2.2mm} \textcolor{color8}{$\cdot$  - $\ast$ - $\cdot$} \hspace{0.001mm} GGL $\lambda_2 = 0.1n$
\fontsize{10}{12}\selectfont 
}}} \\

\centering
a) $n = 25$ 

\includegraphics[width= 0.32\linewidth,clip]{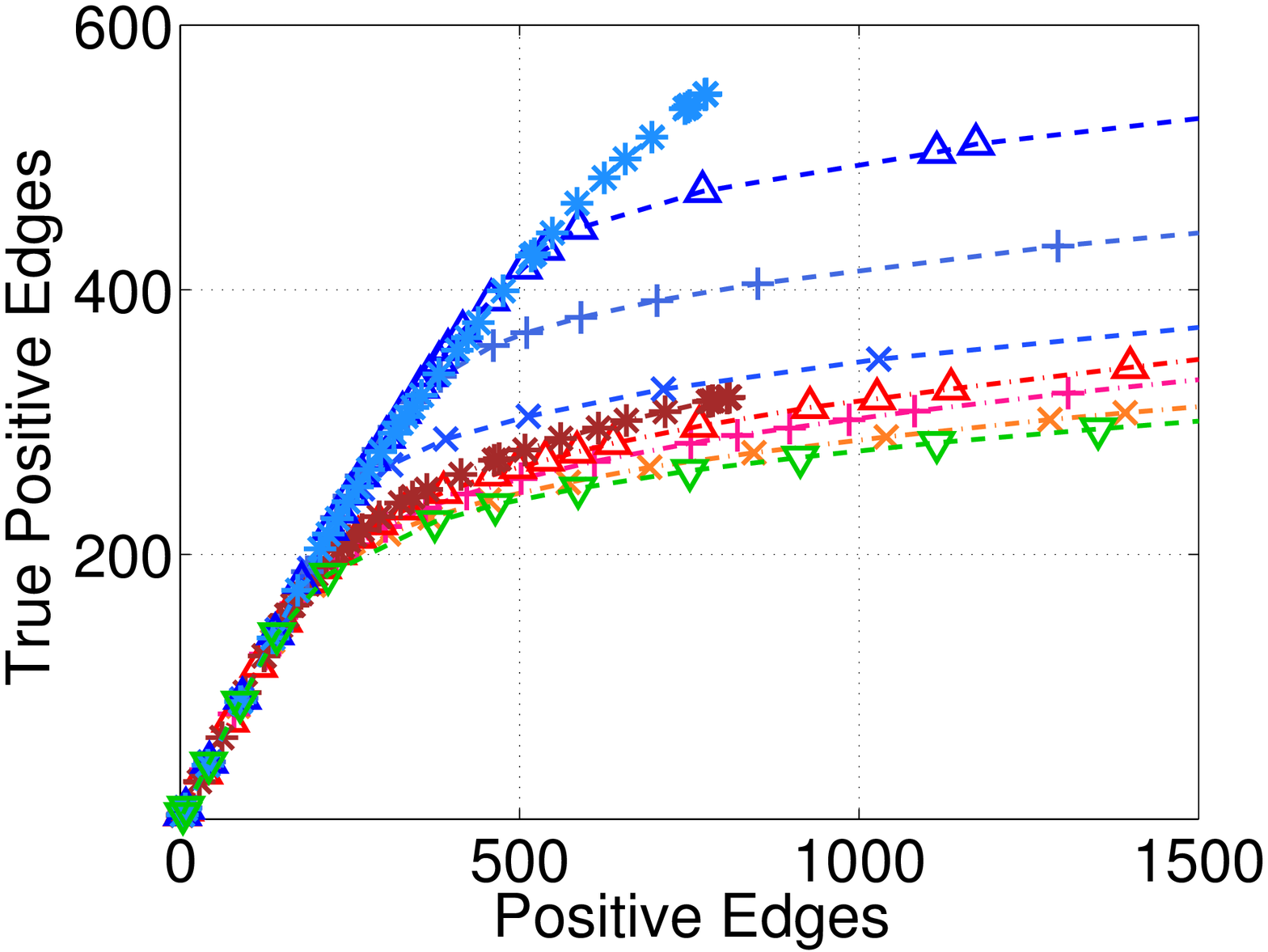}
\includegraphics[width= 0.32\linewidth,clip]{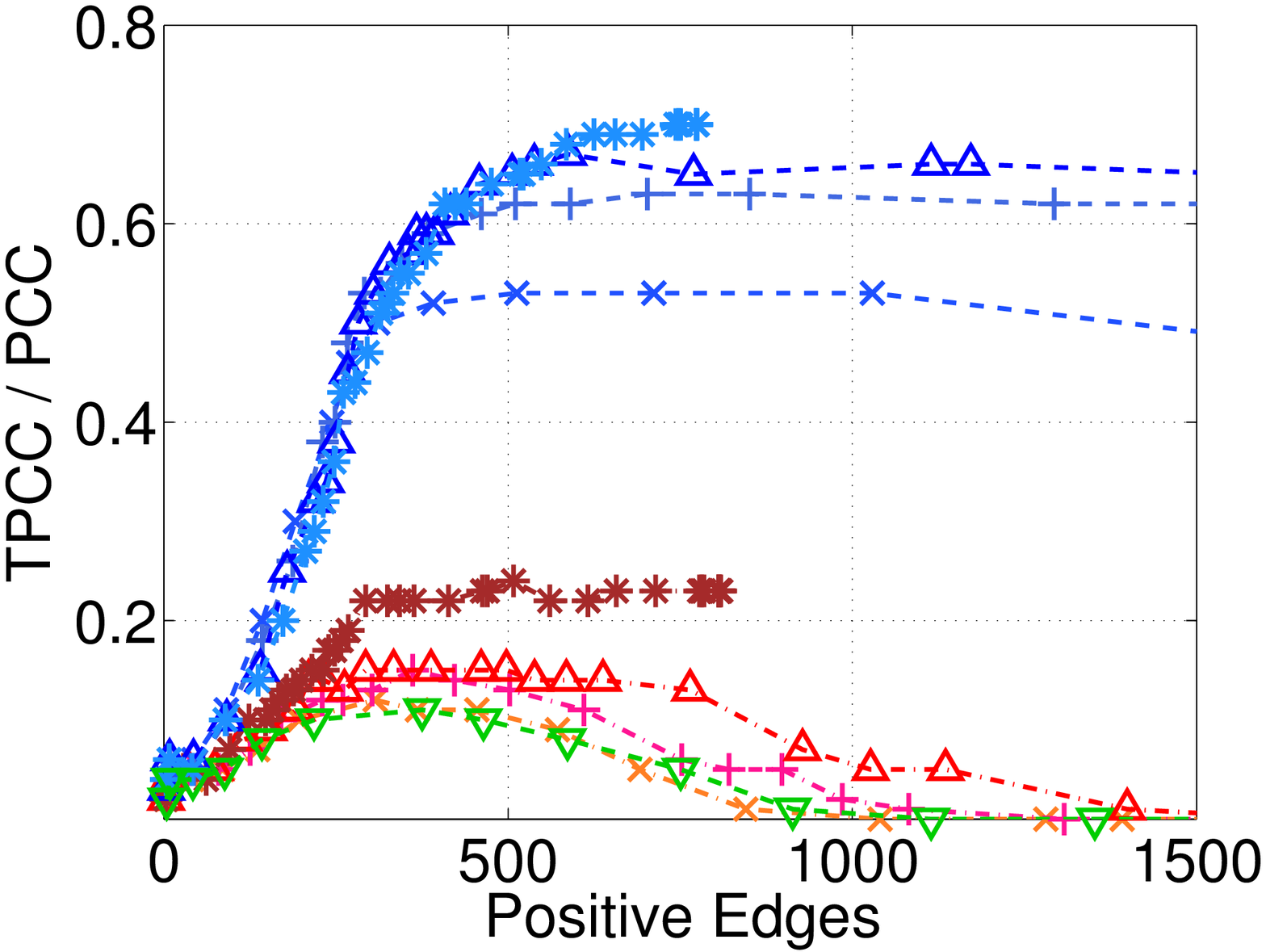}
\includegraphics[width= 0.32\linewidth,clip]{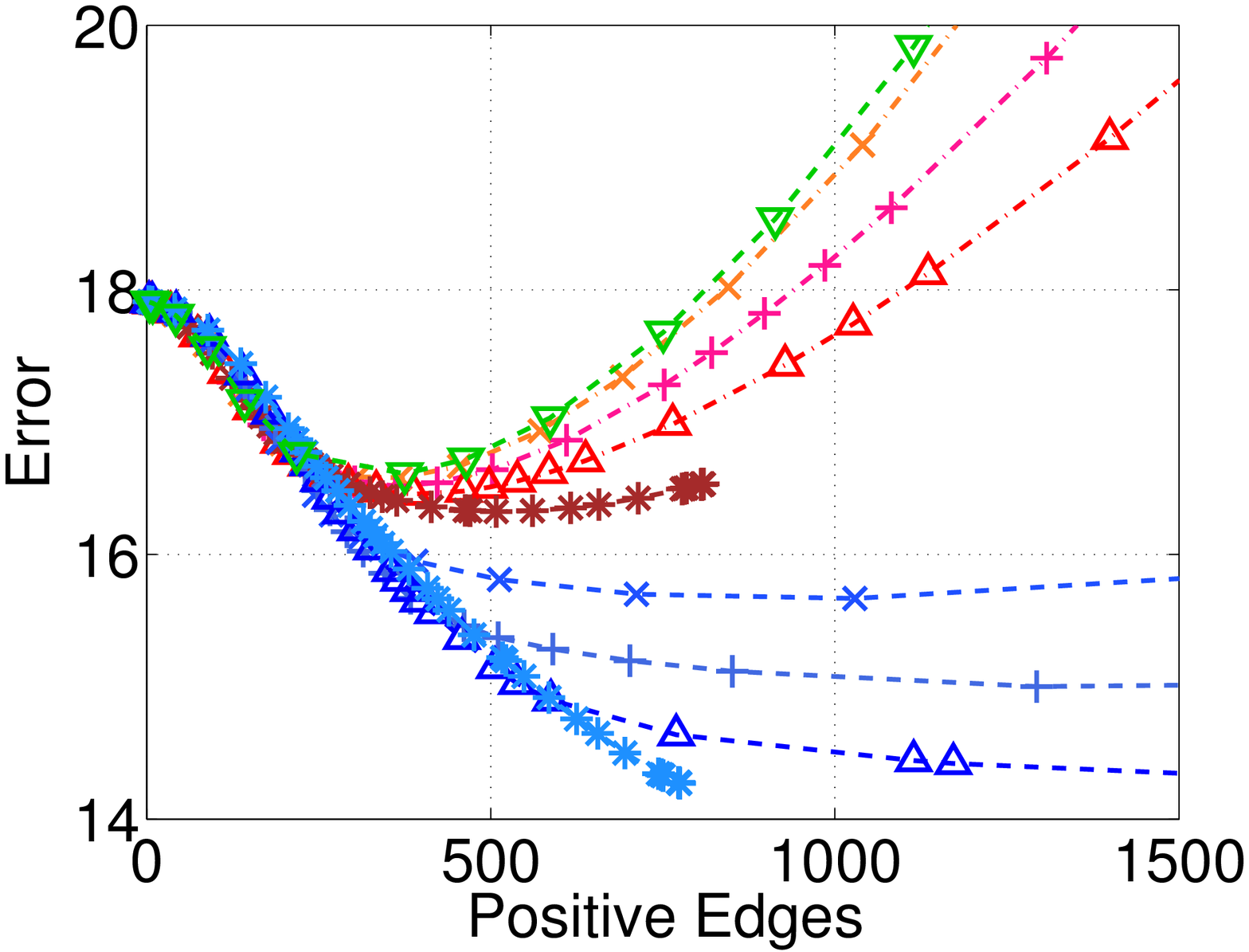} 

(b) $n = 50$

\includegraphics[width= 0.32\linewidth,clip]{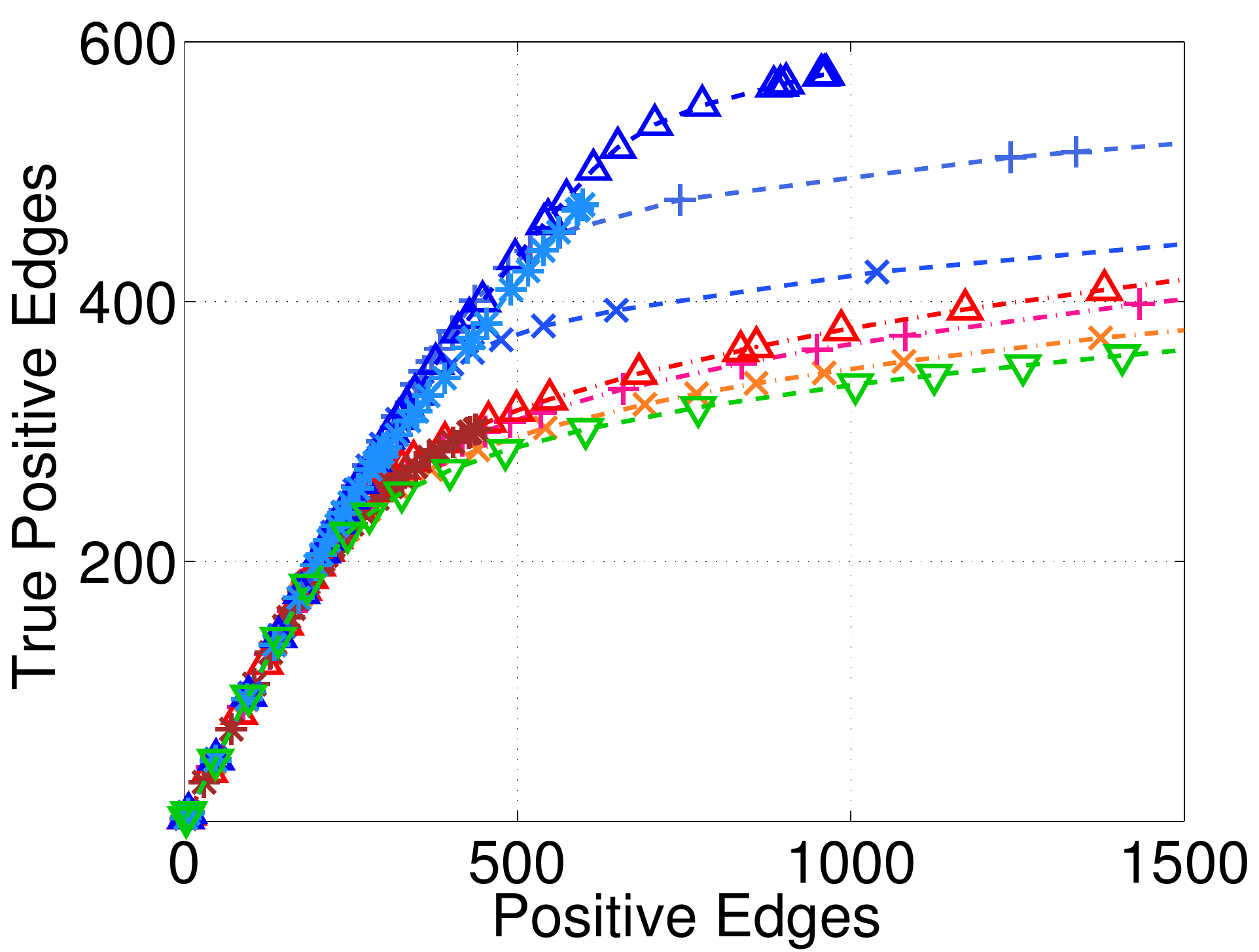}
\includegraphics[width= 0.32\linewidth,clip]{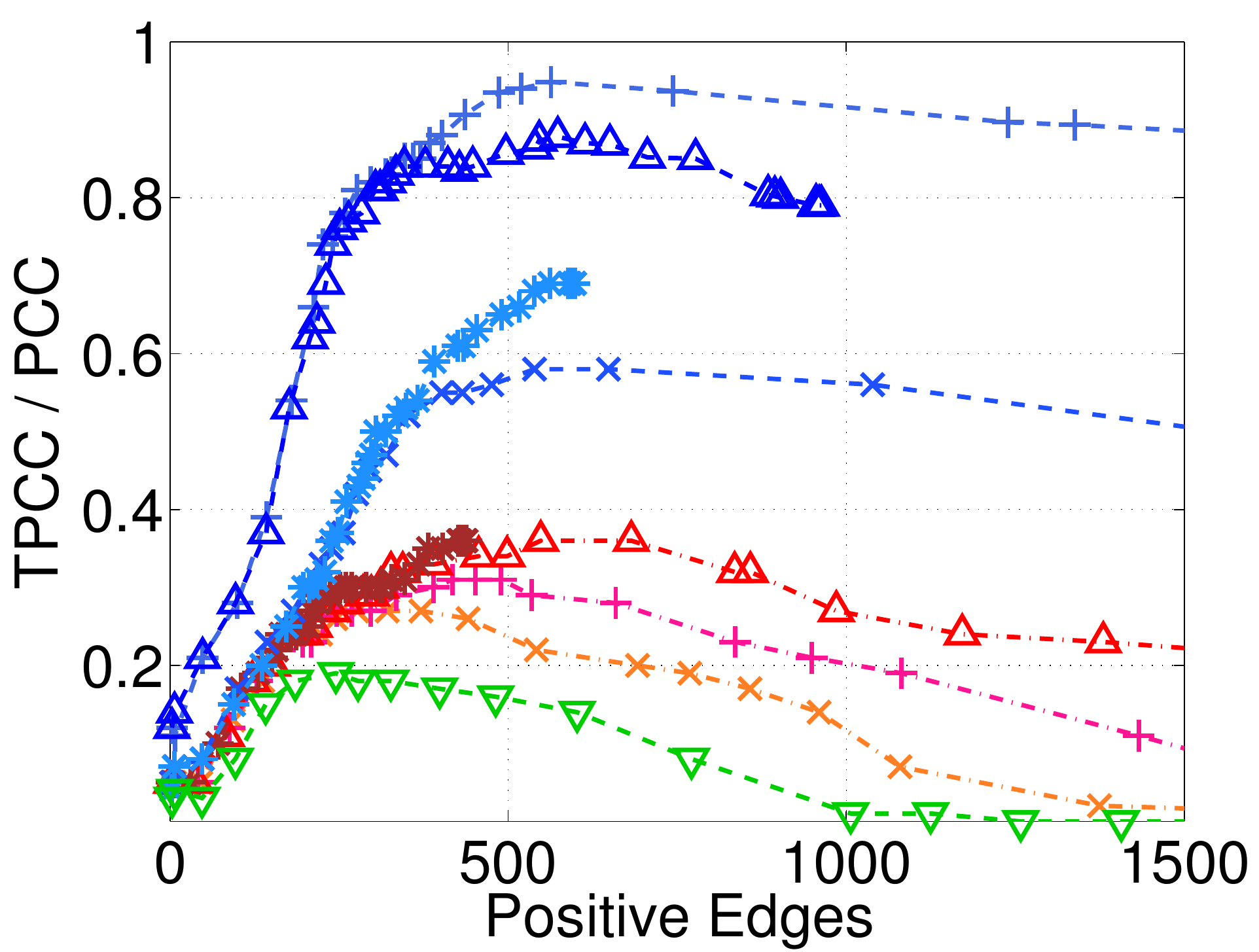}
\includegraphics[width= 0.32\linewidth,clip]{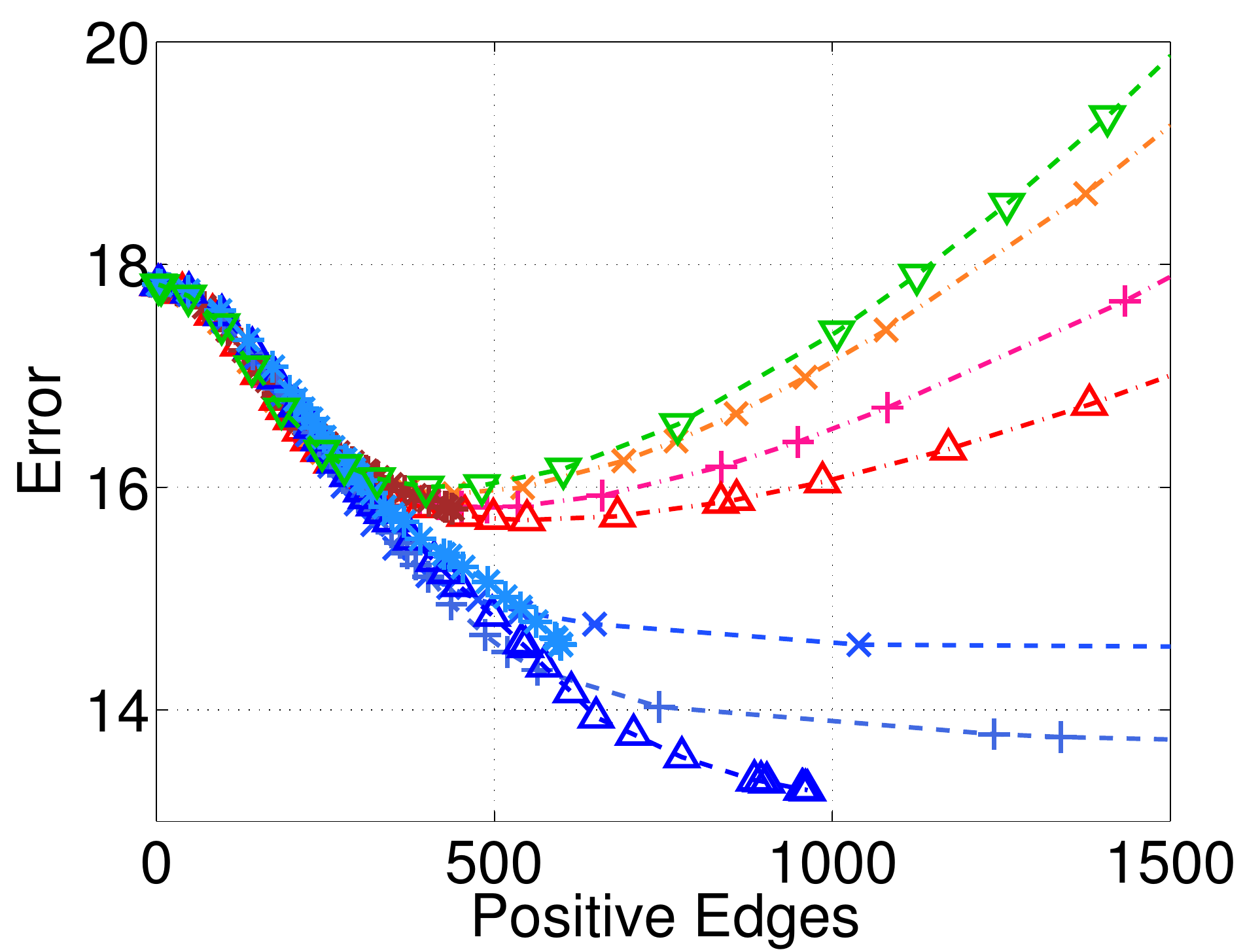}

(c) $n = 100$

\includegraphics[width= 0.32\linewidth,clip]{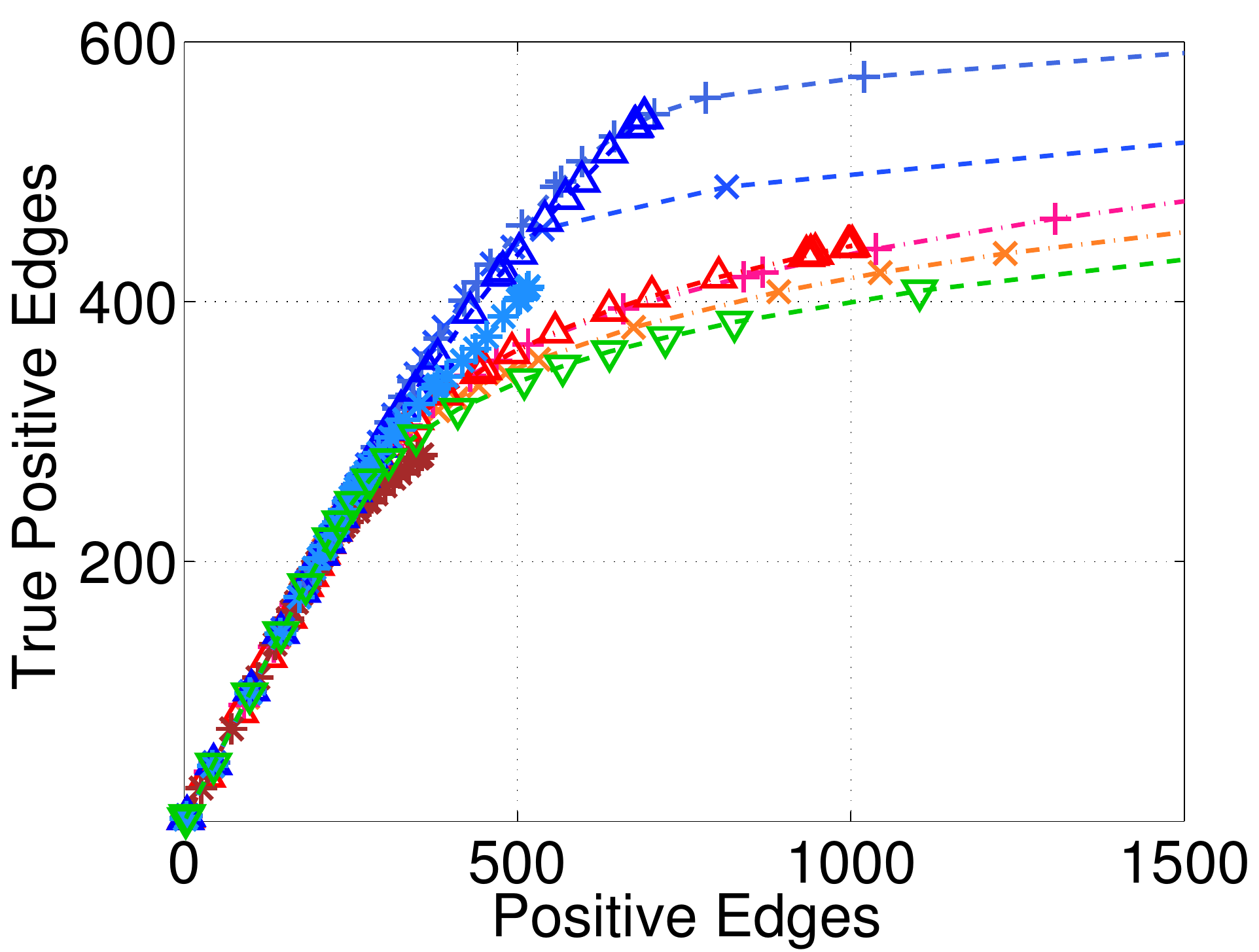}
\includegraphics[width= 0.32\linewidth,clip]{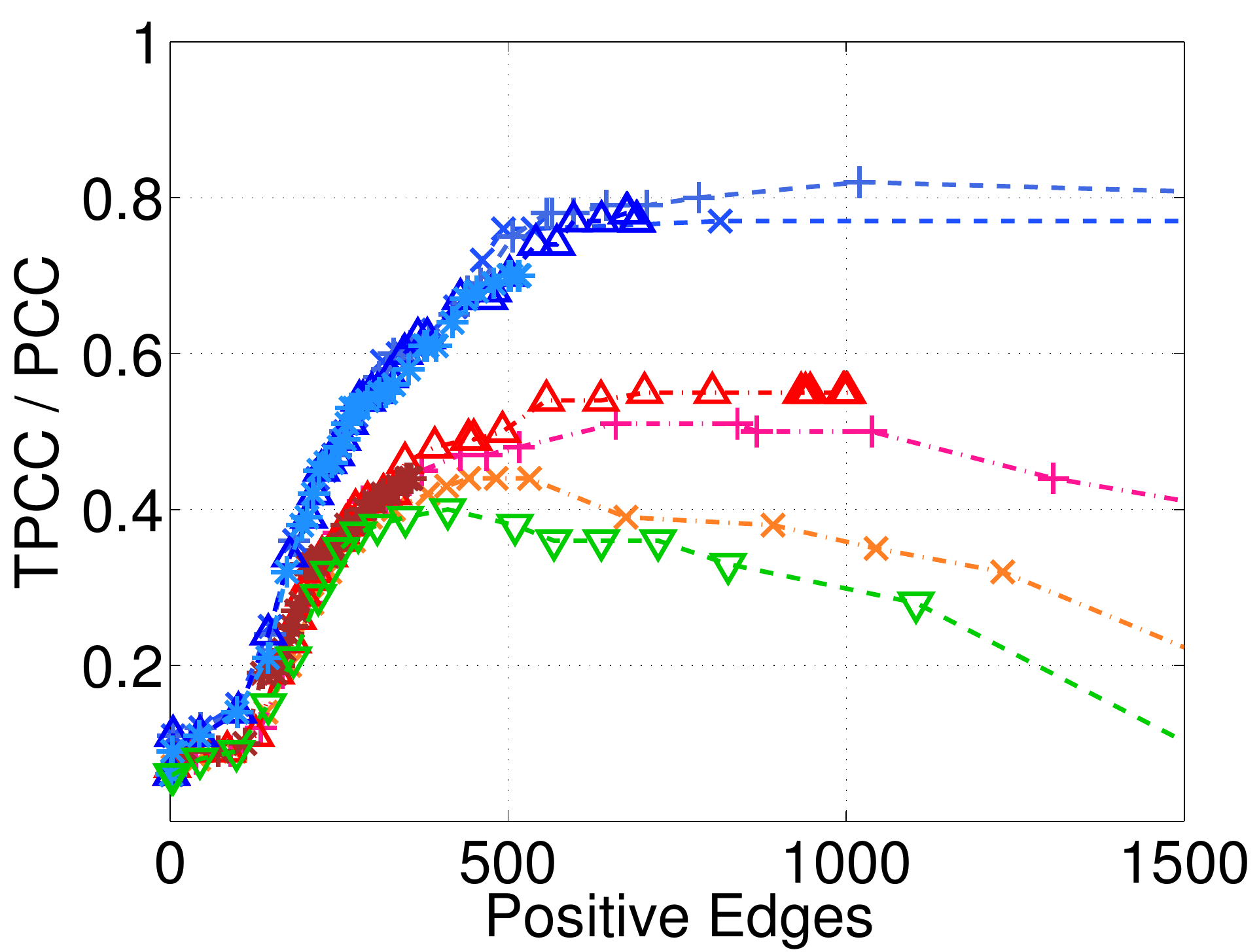}
\includegraphics[width= 0.32\linewidth,clip]{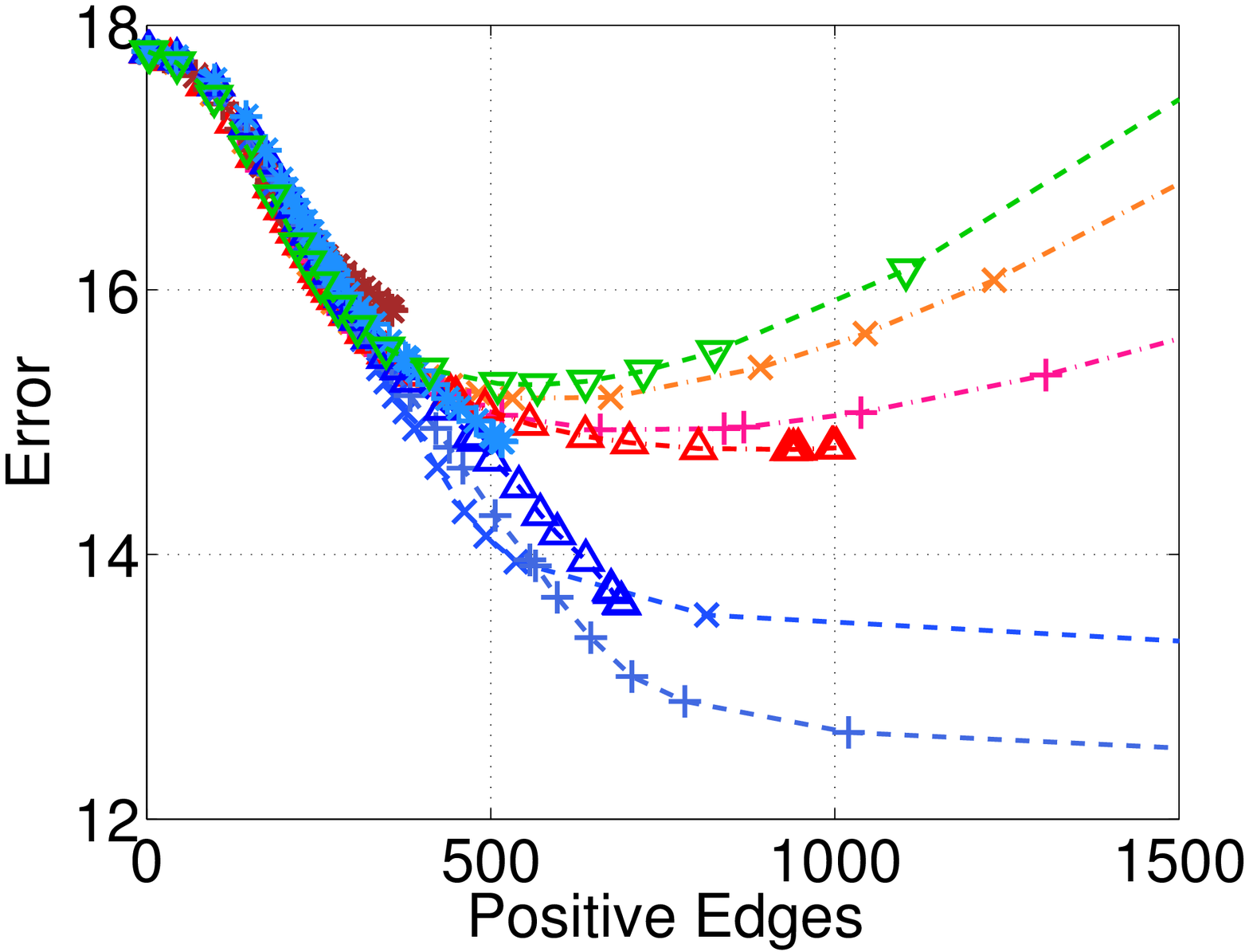}

(d) $n = 200$

\includegraphics[width= 0.32\linewidth,clip]{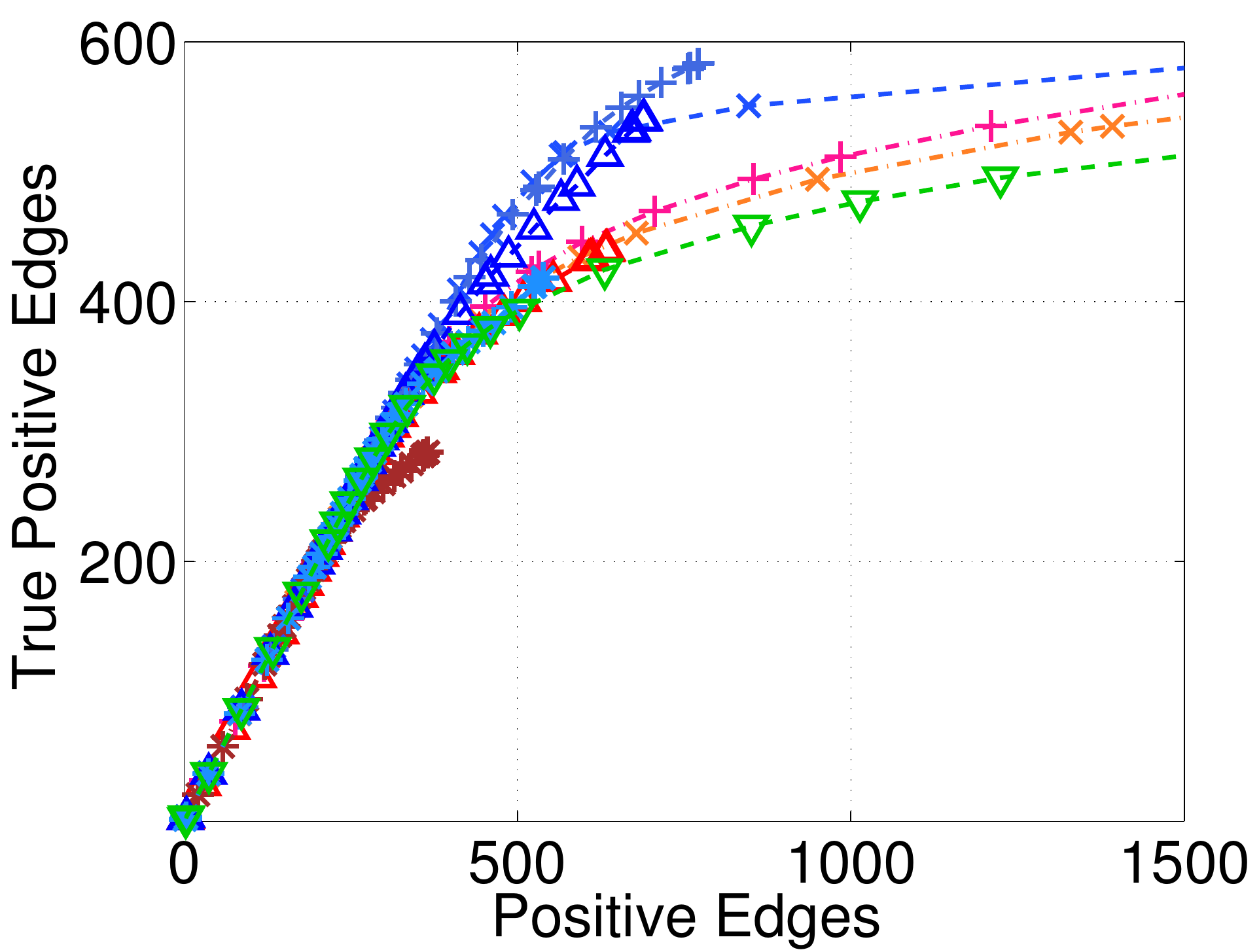}
\includegraphics[width= 0.32\linewidth,clip]{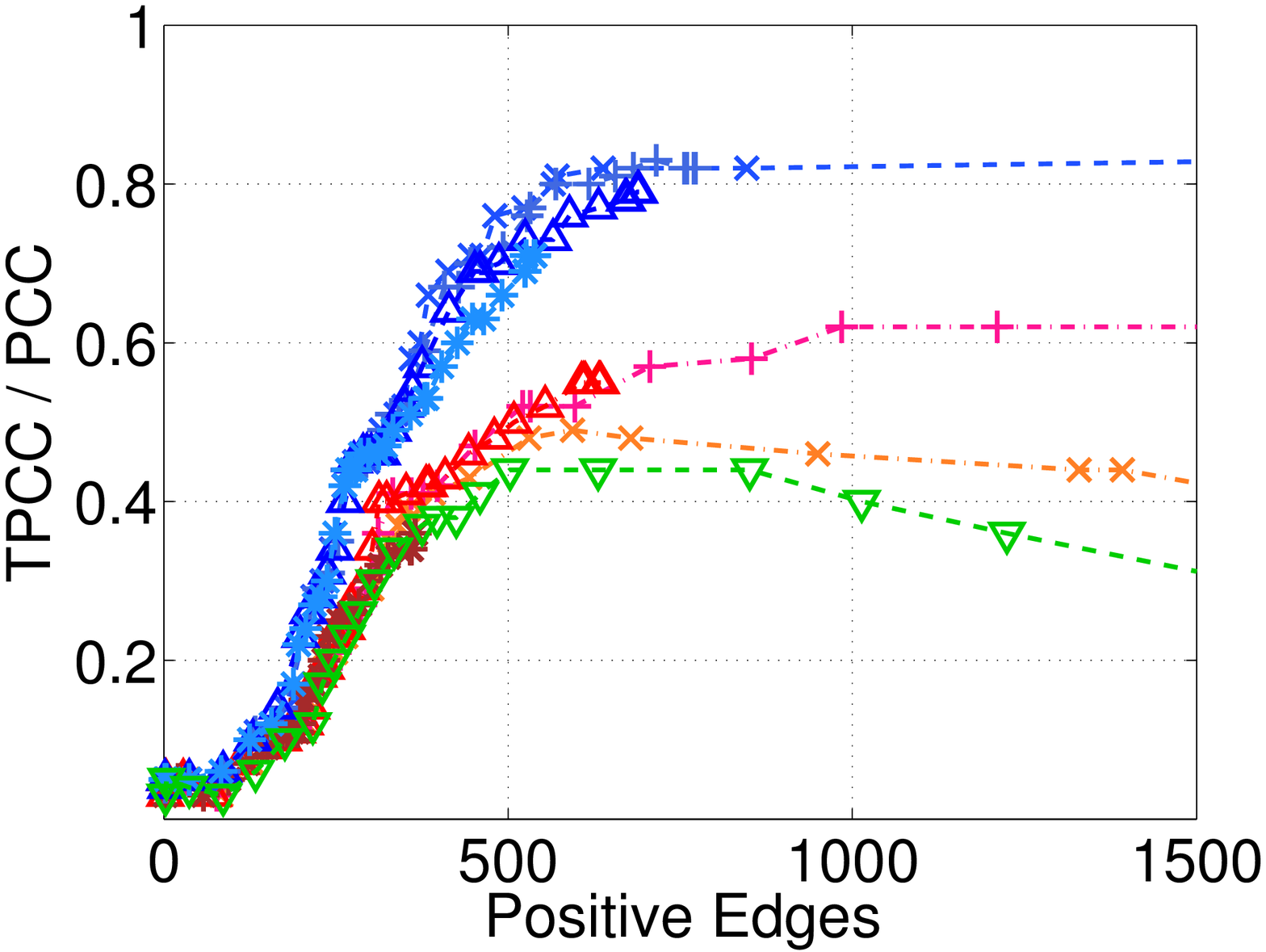}
\includegraphics[width= 0.32\linewidth,clip]{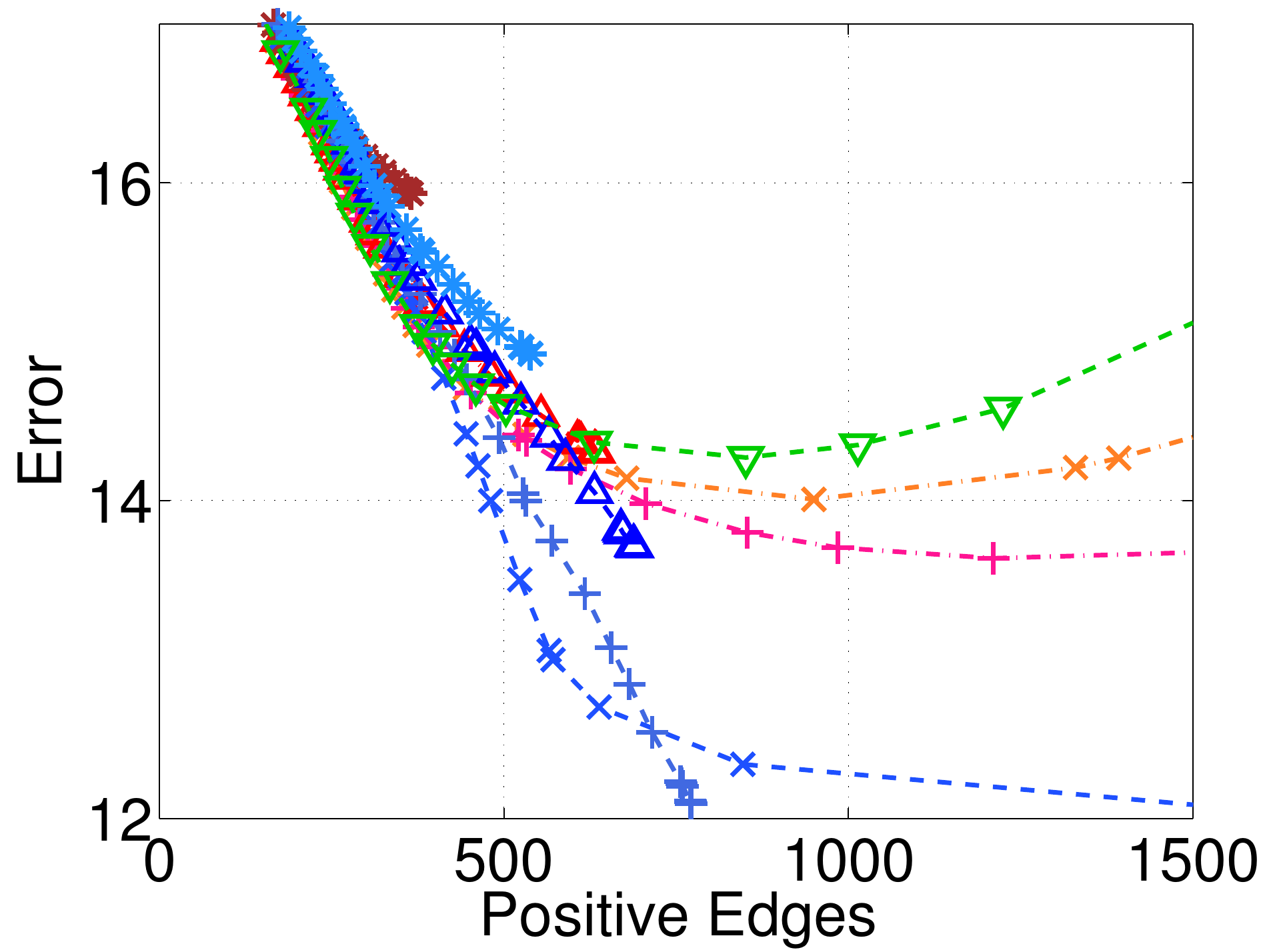}

\caption{\label{fig:CNJGL} Simulation results  on Erdos-Renyi network (Section~\ref{one}) for CNJGL with $q = 2$, GGL, and GL, for \emph{(a):} $n = 25$, \emph{(b):} $n = 50$, \emph{(c):} $n = 100$, \emph{(d):} $n = 200$, when $p = 100$. Each colored line corresponds to a fixed value of $\lambda_2$, as $\lambda_1$ is varied. Axes are described in detail in Table \ref{tbl:TableMetrics}. 
Results are averaged over 100 random generations of the data.}
\end{figure}

The simulation results for the set-up of Section~\ref{one} are displayed in Figures \ref{fig:PNJGL} and \ref{fig:CNJGL}. Each row corresponds to a sample size while each column corresponds to a performance metric.
In Figure~\ref{fig:PNJGL}, PNJGL, FGL, and GL are compared, and in Figure~\ref{fig:CNJGL}, CNJGL, GGL, and GL are compared.
Within each plot, each colored line corresponds to the results obtained using a fixed value of $\lambda_2$ (for either PNJGL, FGL, CNJGL, or GGL), as $\lambda_1$ is varied. Recall that GL corresponds to any of these four approaches with $\lambda_2=0$. 
Note that the number of positive edges (defined in Table~\ref{tbl:TableMetrics}) decreases approximately monotonically with the regularization parameter $\lambda_1$, and so on the $x$-axis we plot the number of positive edges, rather than $\lambda_1$,  for ease of 
interpretation. 

In Figure \ref{fig:PNJGL}, we observe that PNJGL outperforms FGL and GL for a suitable range of the regularization parameter $\lambda_2$, in the sense that for a fixed number of edges estimated, PNJGL identifies more true positives, correctly identifies a greater ratio of perturbed nodes, and yields a lower Frobenius error in the estimates of $\bThetaone$ and $\bThetatwo$.
In particular, PNJGL performs best relative to  FGL and GL when the number of samples is the smallest, that is, in the high-dimensional data setting. 
Unlike FGL, PNJGL fully exploits the fact that differences between $\bThetaone$ and $\bThetatwo$ are due to node perturbation.  
Not surprisingly, 
GL performs worst among the three algorithms, since it does not borrow strength across the conditions in estimating $\bThetaone$ and $\bThetatwo$. 

In Figure \ref{fig:CNJGL}, we note that CNJGL outperforms GGL and GL for a suitable range of the regularization parameter $\lambda_2$.
In particular, CNJGL outperforms GGL and GL by a larger margin when the number of samples is the smallest. Once again, GL performs the worst since it does not borrow strength across the two networks; CNJGL performs the best since it fully exploits the presence of hub nodes in the data.

We note one interesting feature of Figure~\ref{fig:CNJGL}: the colored lines corresponding to CNJGL with very large values of $\lambda_2$ do not extend beyond around 400 positive edges. This is because for CNJGL, a large value of $\lambda_2$ induces sparsity in the network estimates, even if $\lambda_1$ is small or zero. Consequently, it is not possible to obtain a dense estimate of $\bThetaone$ and $\bThetatwo$ if CNJGL is performed with a large value of $\lambda_2$. In contrast, in the case of PNJGL, sparsity is induced only by $\lambda_1$, and not at all by $\lambda_2$. We note that a similar situation occurs for the edge-based counterparts of CNJGL and PNJGL: when GGL is performed with a large value of $\lambda_2$ then the network estimates are necessarily sparse, regardless of the value of 
$\lambda_1$. But the same is not true for FGL.

\begin{figure}[!htbp]

\centering
(a) {PNJGL/FGL/GL:} \\[0.1in]

\noindent
\makebox[1\linewidth][l]{\hspace{\dimexpr-\fboxsep-\fboxrule\relax}
\fbox {\parbox{0.97\linewidth}{ 

\fontsize{8.3}{12}\selectfont
\textcolor{color1}{-  - $\mathsmaller{\times}$ - -}  \hspace{0.001mm} PNJGL $\lambda_2 = 0.3n$
\hspace{1.2mm} \textcolor{color2}{-  - $\mathsmaller{+}$ - -} \hspace{0.001mm} PNJGL $\lambda_2 = 0.5n$ 
\hspace{1.2mm} \textcolor{color3}{-  - $\mathsmaller{\mathsmaller{\triangle}}$  - -}  \hspace{0.001mm} PNJGL $\lambda_2 = 1.0n$
\hspace{1.2mm} \textcolor{color4}{-  - $\ast$ - -} \hspace{0.001mm} PNJGL $\lambda_2 = 2.0n$
\hspace{1mm} \textcolor{color9}{$\cdot$ -  $\mathsmaller{\bigtriangledown}$  - $\cdot$} \hspace{0.001mm} GL 

\fontsize{8.3}{12}\selectfont 
\noindent
\textcolor{color12}{$\cdot$  - $\mathsmaller{\times}$ - $\cdot$}  \hspace{0.001mm} FGL $\lambda_2 = 0.01n$
\hspace{1mm}\textcolor{color11}{$\cdot$  - $\mathsmaller{\times}$ - $\cdot$} \hspace{0.001mm} FGL $\lambda_2 = 0.02$ 
\hspace{1mm} \textcolor{color5}{$\cdot$  - $\mathsmaller{+}$ - $\cdot$} \hspace{0.001mm} FGL $\lambda_2 = 0.03n$
\hspace{1mm} \textcolor{color6}{$\cdot$  - $\mathsmaller{\mathsmaller{\triangle}}$  - $\cdot$} \hspace{0.001mm} FGL $\lambda_2 = 0.05n$
\hspace{1mm} \textcolor{color7}{$\cdot$  - $\ast$ - $\cdot$} \hspace{0.001mm} FGL $\lambda_2 = 0.2n$ 
}}}

\includegraphics[width= 0.32\linewidth,clip]{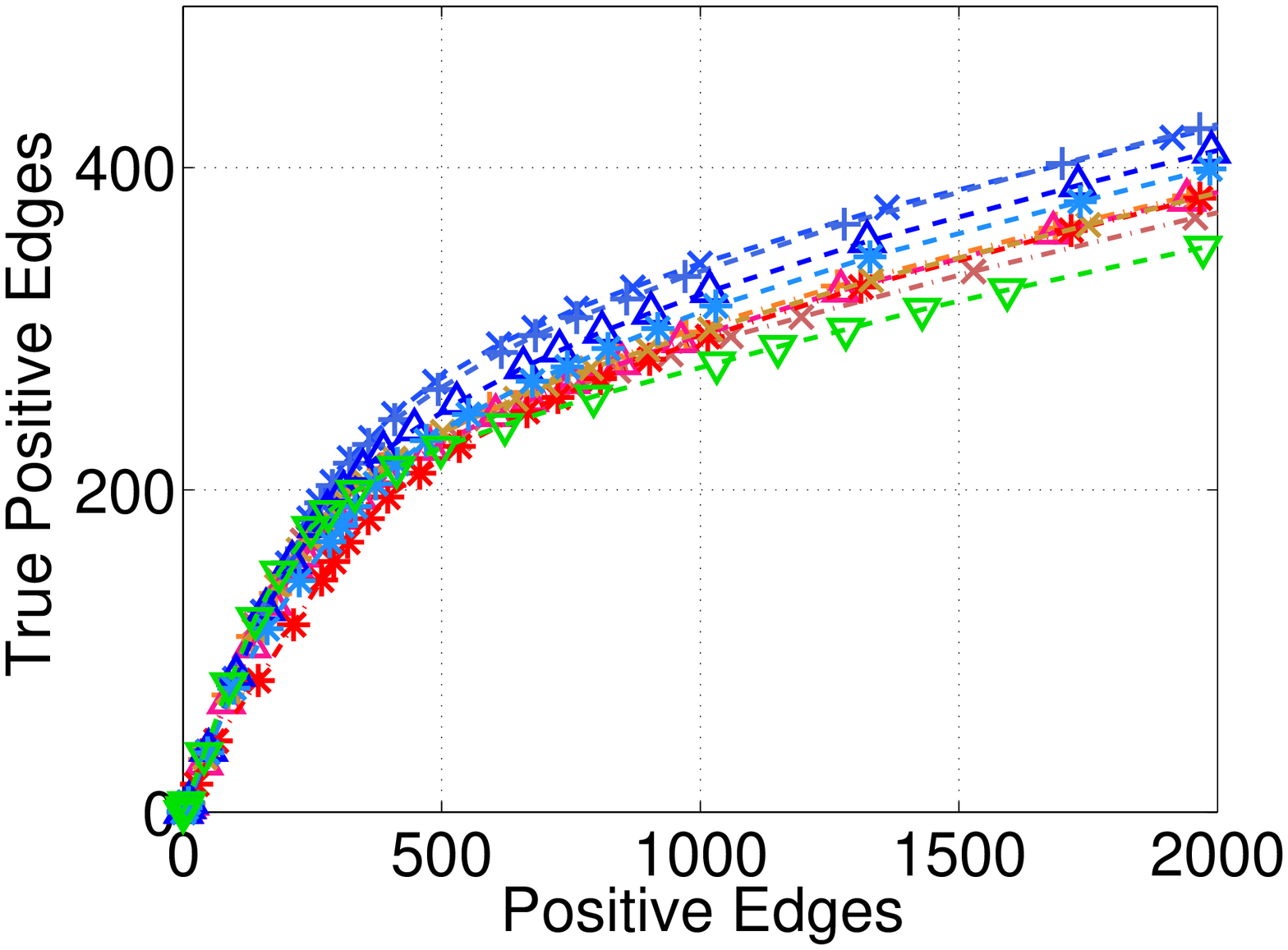}
\includegraphics[width= 0.32\linewidth,clip]{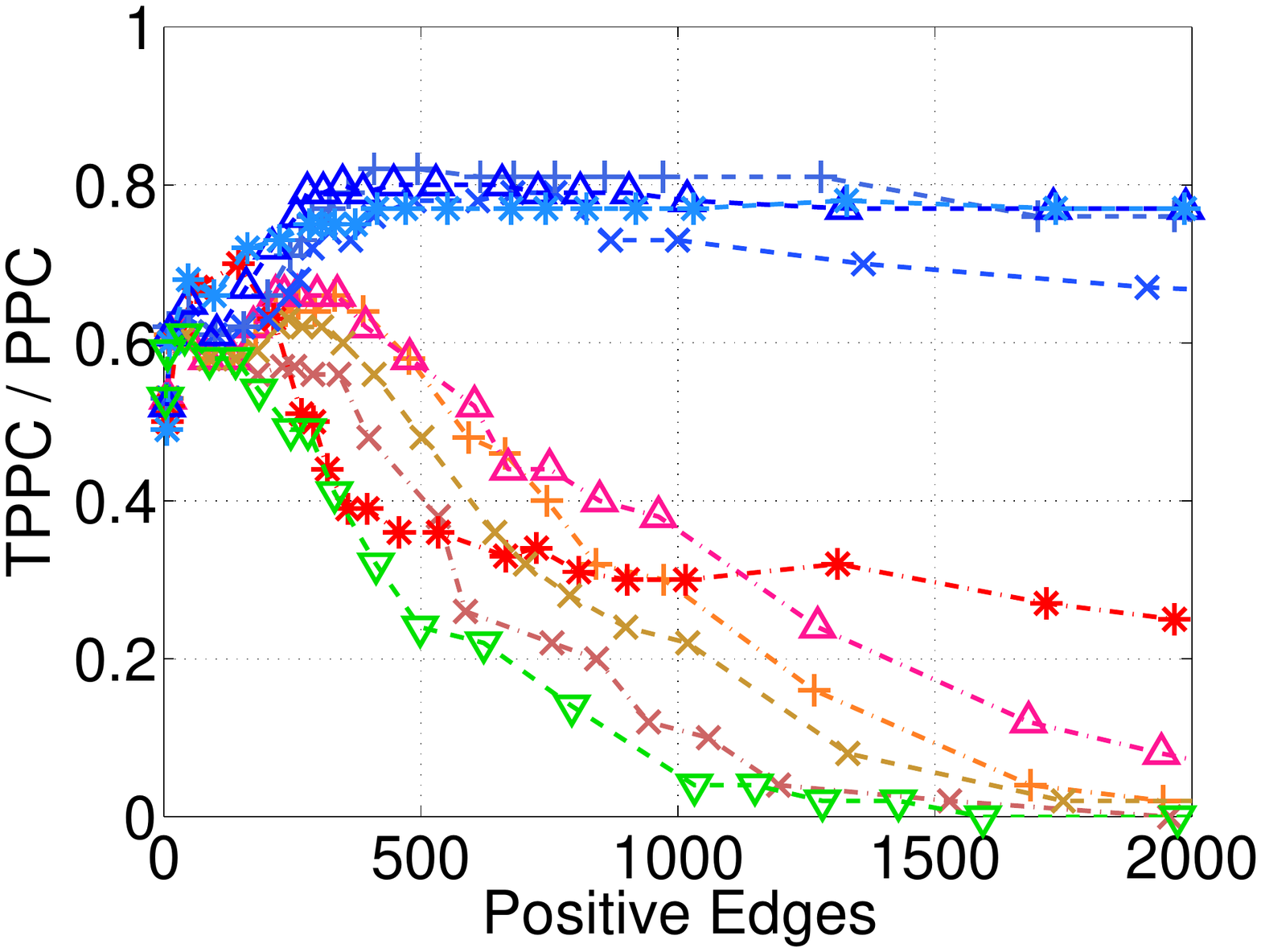}
\includegraphics[width= 0.32\linewidth,clip]{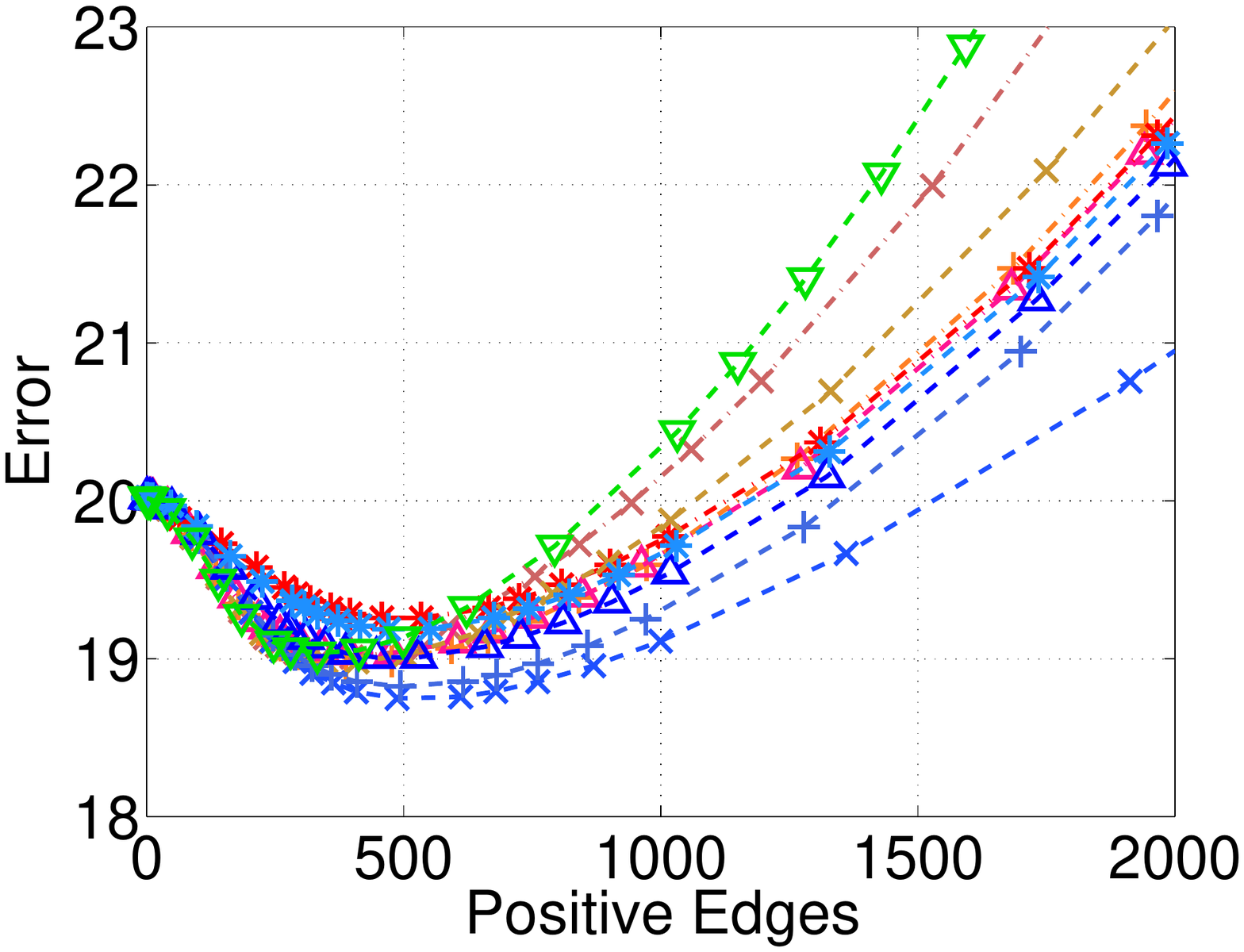}

(b) {CNJGL/GGL/GL:}\\[0.1in]

\noindent
\makebox[1\linewidth][l]{\hspace{\dimexpr-\fboxsep-\fboxrule\relax}
\fbox {\parbox{0.97\linewidth}{ 
\fontsize{8.5}{12}\selectfont
\textcolor{color1}{-  - $\mathsmaller{\times}$ - -}  \hspace{0.001mm} CNJGL $\lambda_2 = 0.3n$
\hspace{1mm}\textcolor{color2}{-  - $\mathsmaller{+}$ - -} \hspace{0.001mm} CNJGL $\lambda_2 = 0.6n$
\hspace{1mm} \textcolor{color3}{-  - $\mathsmaller{\triangle}$  - -}  \hspace{0.001mm} CNJGL $\lambda_2 = 1.0n$
\hspace{1mm} \textcolor{color4}{-  - $\ast$ - -} \hspace{0.001mm} CNJGL $\lambda_2 = 1.5n$

\fontsize{8.5}{12}\selectfont
\textcolor{color5}{$\cdot$ -  $\mathsmaller{\bigtriangledown}$  - $\cdot$}  \hspace{0.001mm} GGL $\lambda_2 = 0.01n$
\hspace{1mm} \textcolor{color6}{$\cdot$  - $\mathsmaller{\times}$  - $\cdot$} \hspace{0.001mm} GGL $\lambda_2 = 0.05n$
\hspace{1mm} \textcolor{color13}{$\cdot$  - $\ast$ - $\cdot$} \hspace{0.001mm} GGL $\lambda_2 = 0.005n$
\hspace{1mm} \textcolor{color9}{$\cdot$ -  $\mathsmaller{\bigtriangledown}$  - $\cdot$} \hspace{0.001mm} GL

}}} \\ 

\includegraphics[width= 0.32\linewidth,clip]{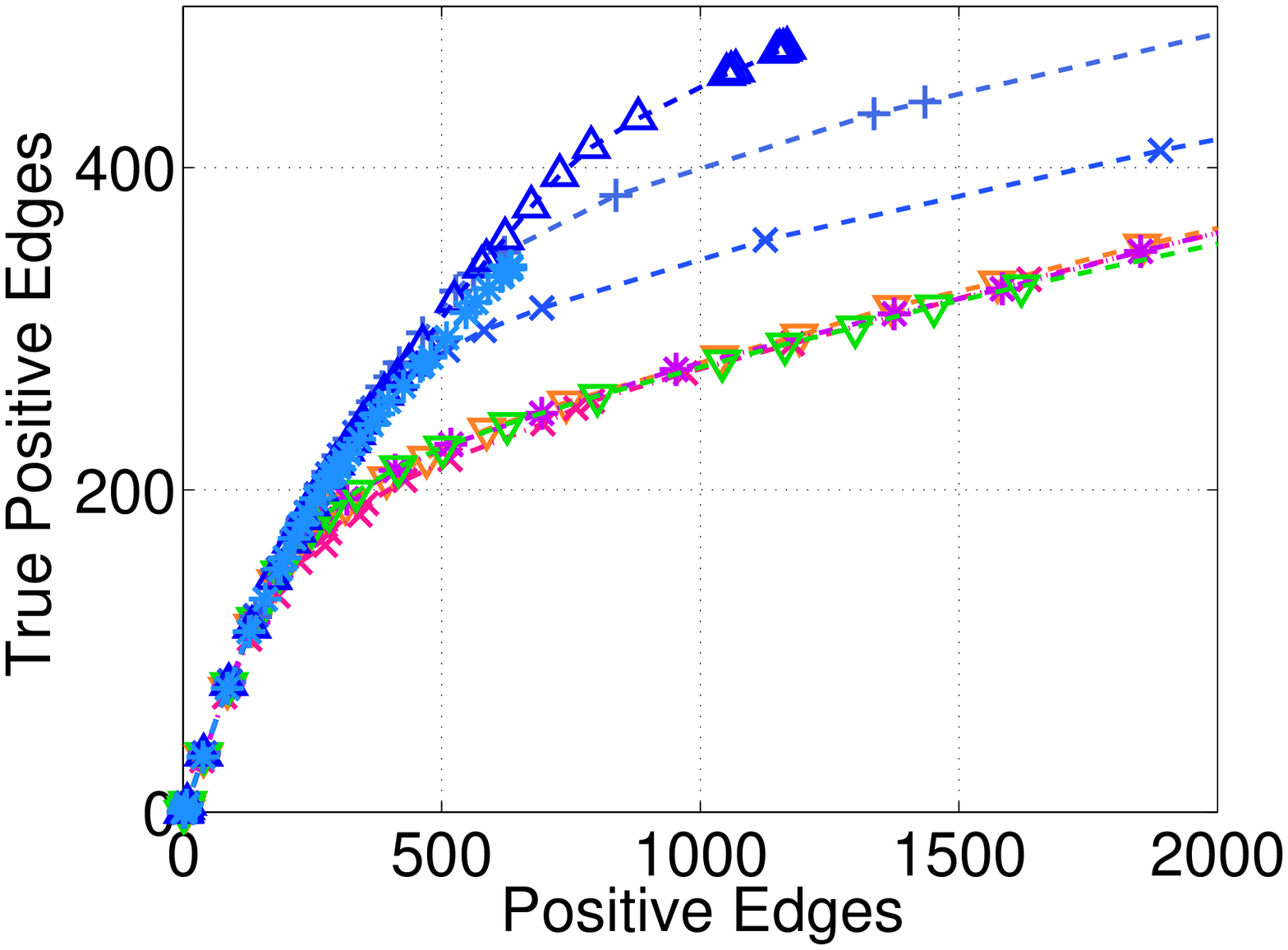}
\includegraphics[width= 0.32\linewidth,clip]{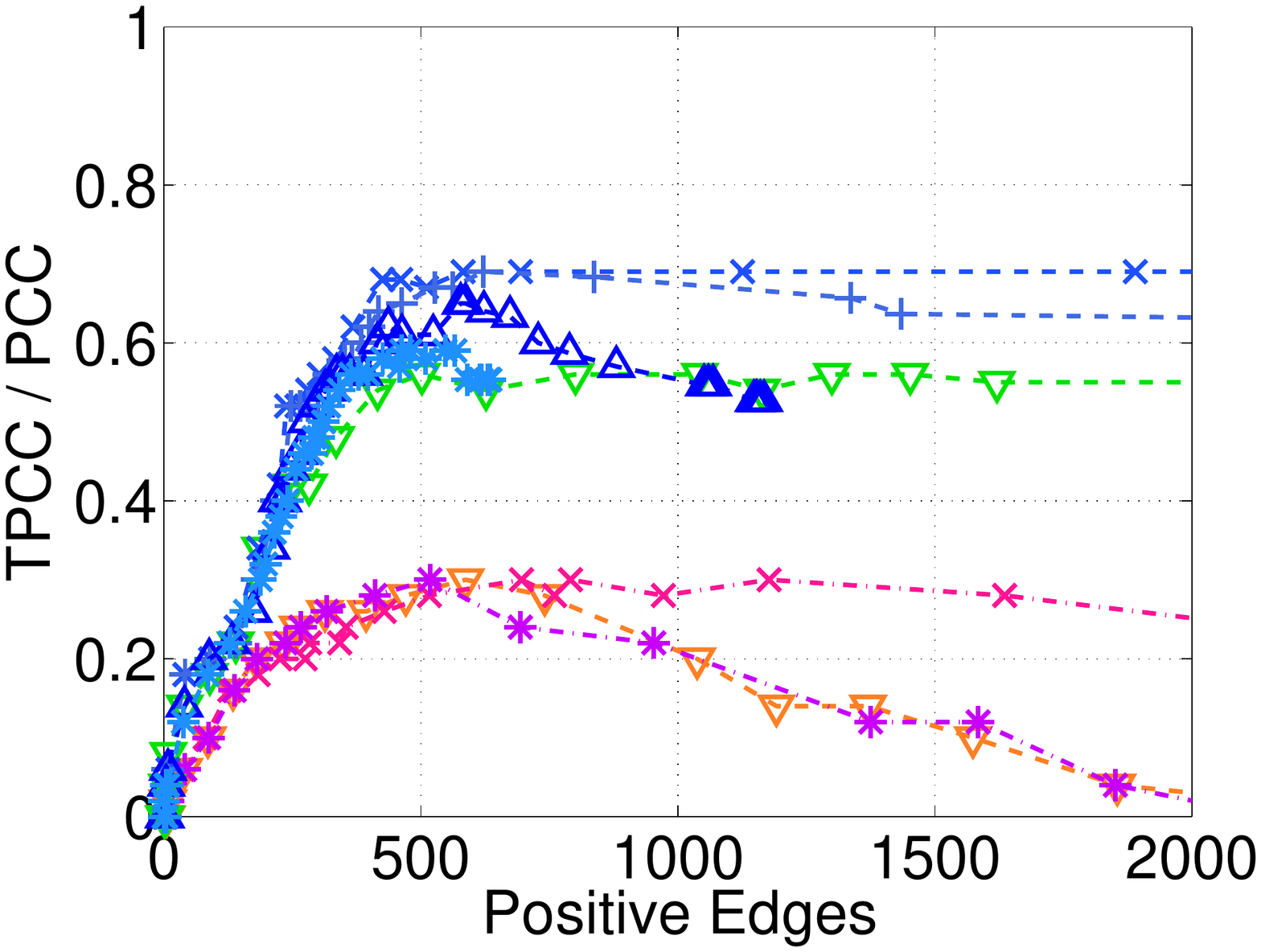}
\includegraphics[width= 0.32\linewidth,clip]{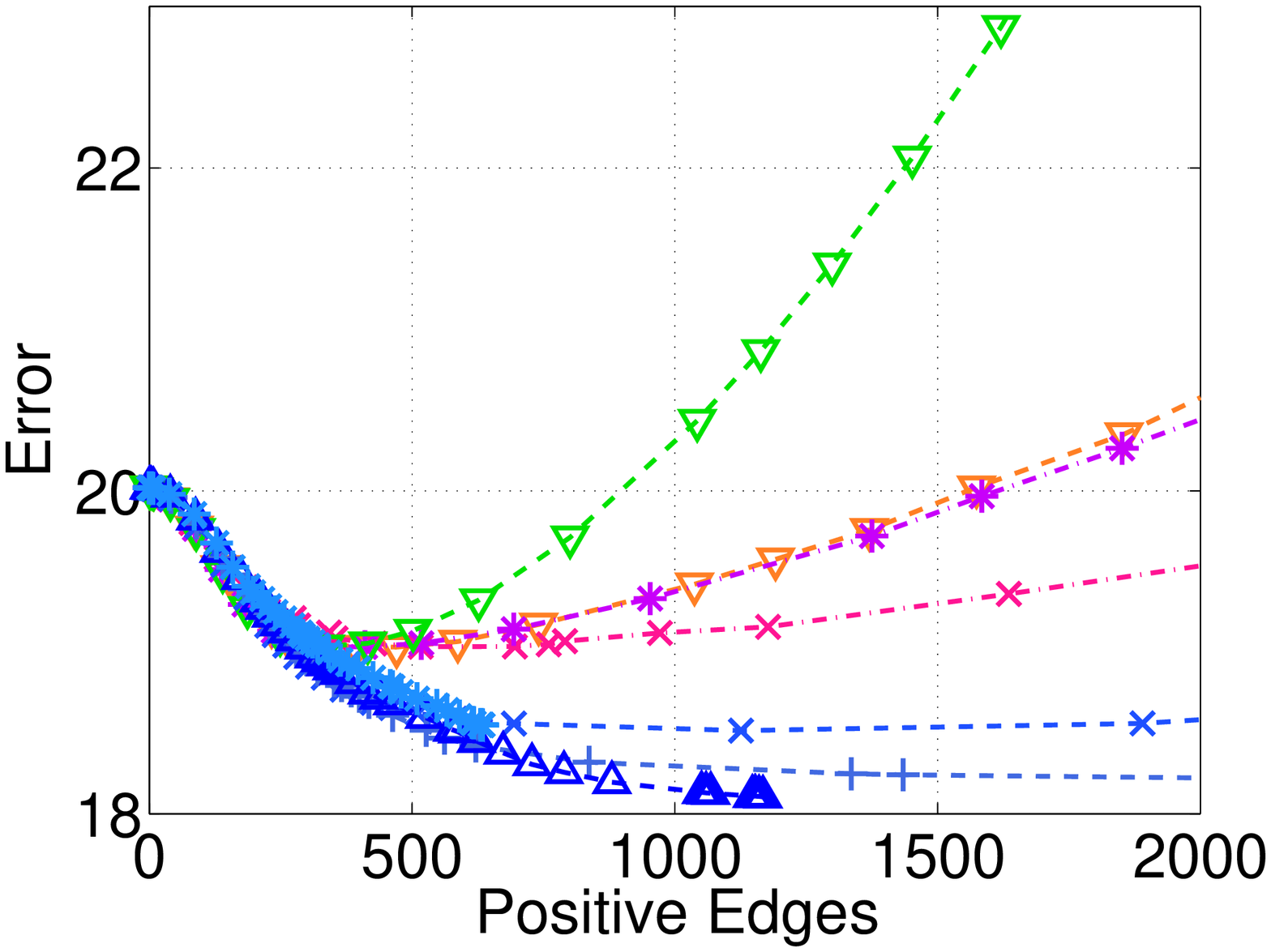}

\caption{\label{fig:scale-free} {Simulation results  on scale-free network (Section~\ref{two}) for \emph{(a):} PNJGL with $q=2$, FGL, and GL, and \emph{(b):} CNJGL with $q=2$, GGL, and GL, with $p =100$ and $n = 50$. Each colored line corresponds to a fixed value of $\lambda_2$, as $\lambda_1$ is varied. Axes are described in detail in Table \ref{tbl:TableMetrics}.
Results are averaged over 50 random generations of the data.}} 
\end{figure}

\begin{figure}[!htbp]

\centering
(a) {PNJGL/FGL/GL:}\\[0.1in]

\noindent
\makebox[1\linewidth][l]{\hspace{\dimexpr-\fboxsep-\fboxrule\relax}
\fbox {\parbox{0.97\linewidth}{ 
\fontsize{8}{12}\selectfont 
\noindent

\fontsize{8}{12}\selectfont
\textcolor{color1}{-  - $\mathsmaller{\times}$ - -}  \hspace{0.001mm} PNJGL $\lambda_2 = 0.3n$
\hspace{1.2mm} \textcolor{color2}{-  - $\mathsmaller{+}$ - -} \hspace{0.001mm} PNJGL $\lambda_2 = 0.5n$ 
\hspace{1.2mm} \textcolor{color3}{-  - $\mathsmaller{\triangle}$  - -}  \hspace{0.001mm} PNJGL $\lambda_2 = 1.0n$
\hspace{1.2mm} \textcolor{color4}{-  - $\ast$ - -} \hspace{0.001mm} PNJGL $\lambda_2 = 2.0n$
\hspace{1mm} \textcolor{color9}{$\cdot$ -  $\mathsmaller{\bigtriangledown}$  - $\cdot$} \hspace{0.001mm} GL 

\textcolor{color12}{$\cdot$  - $\mathsmaller{\times}$ - $\cdot$}  \hspace{0.001mm} FGL $\lambda_2 = 0.01n$
\hspace{1.2mm}\textcolor{color5}{$\cdot$  - $\mathsmaller{+}$ - $\cdot$} \hspace{0.001mm} FGL $\lambda_2 = 0.03$ 
\hspace{1.2mm} \textcolor{color6}{$\cdot$  - $\mathsmaller{\triangle}$ - $\cdot$} \hspace{0.001mm} FGL $\lambda_2 = 0.05n$
\hspace{1.2mm} \textcolor{color7}{$\cdot$  - $\ast$  - $\cdot$} \hspace{0.001mm} FGL $\lambda_2 = 0.2n$
\hspace{1.2mm} \textcolor{color8}{$\cdot$  - $\mathsmaller{\bigtriangledown}$ - $\cdot$} \hspace{0.001mm} FGL $\lambda_2 = 0.5n$ 
}}}

\includegraphics[width= 0.32\linewidth,clip]{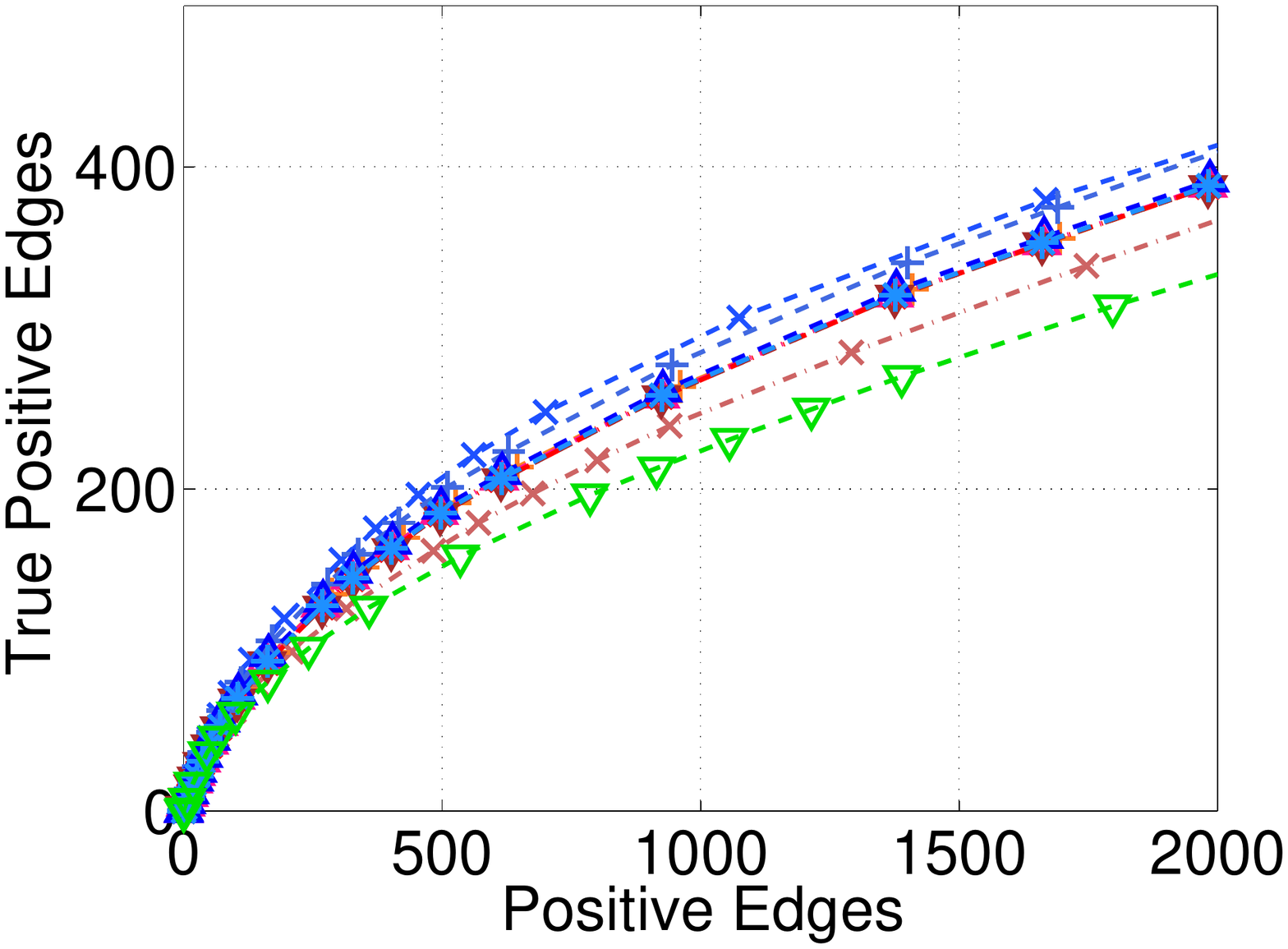}
\includegraphics[width= 0.32\linewidth,clip]{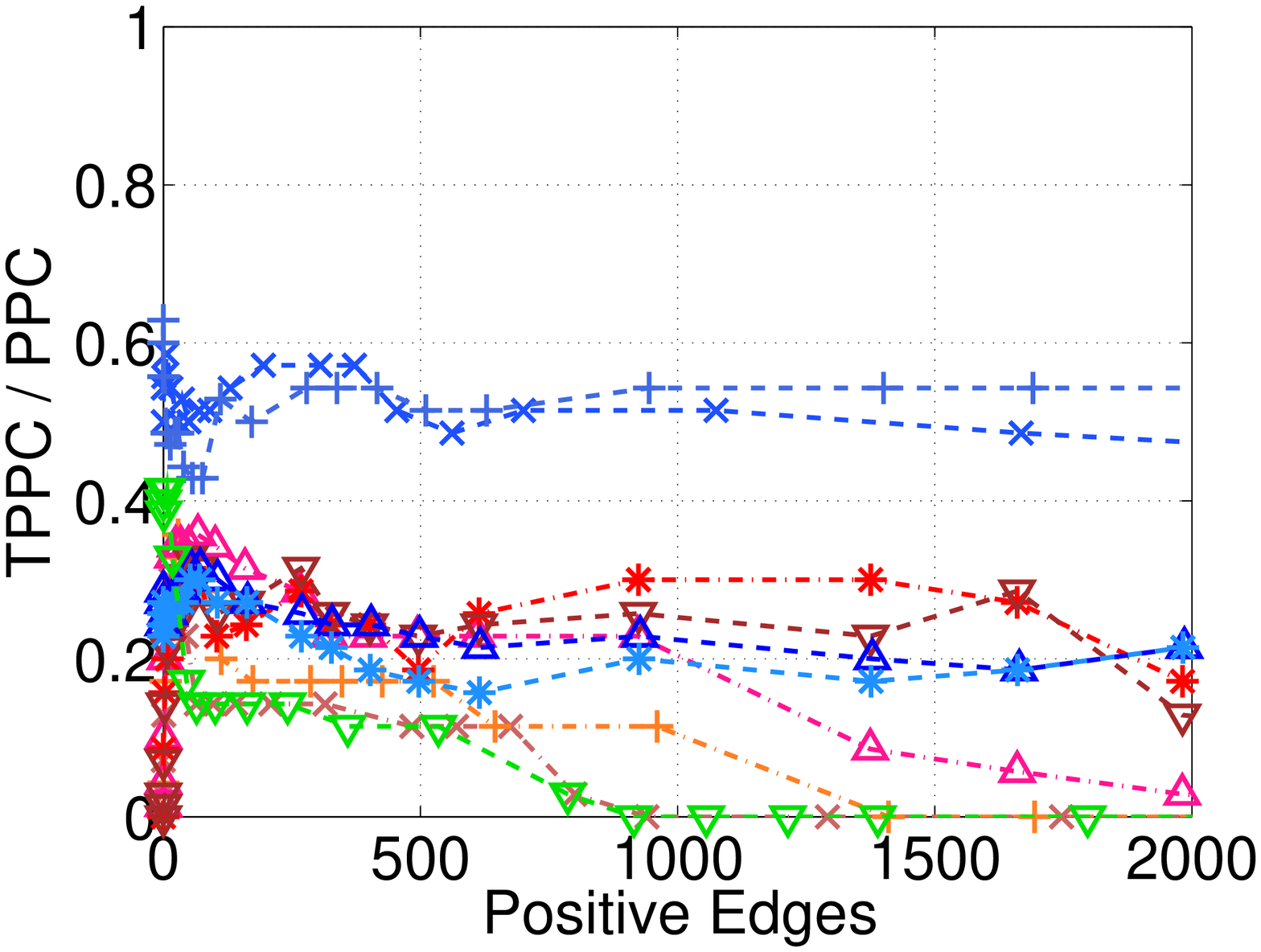}
\includegraphics[width= 0.32\linewidth,clip]{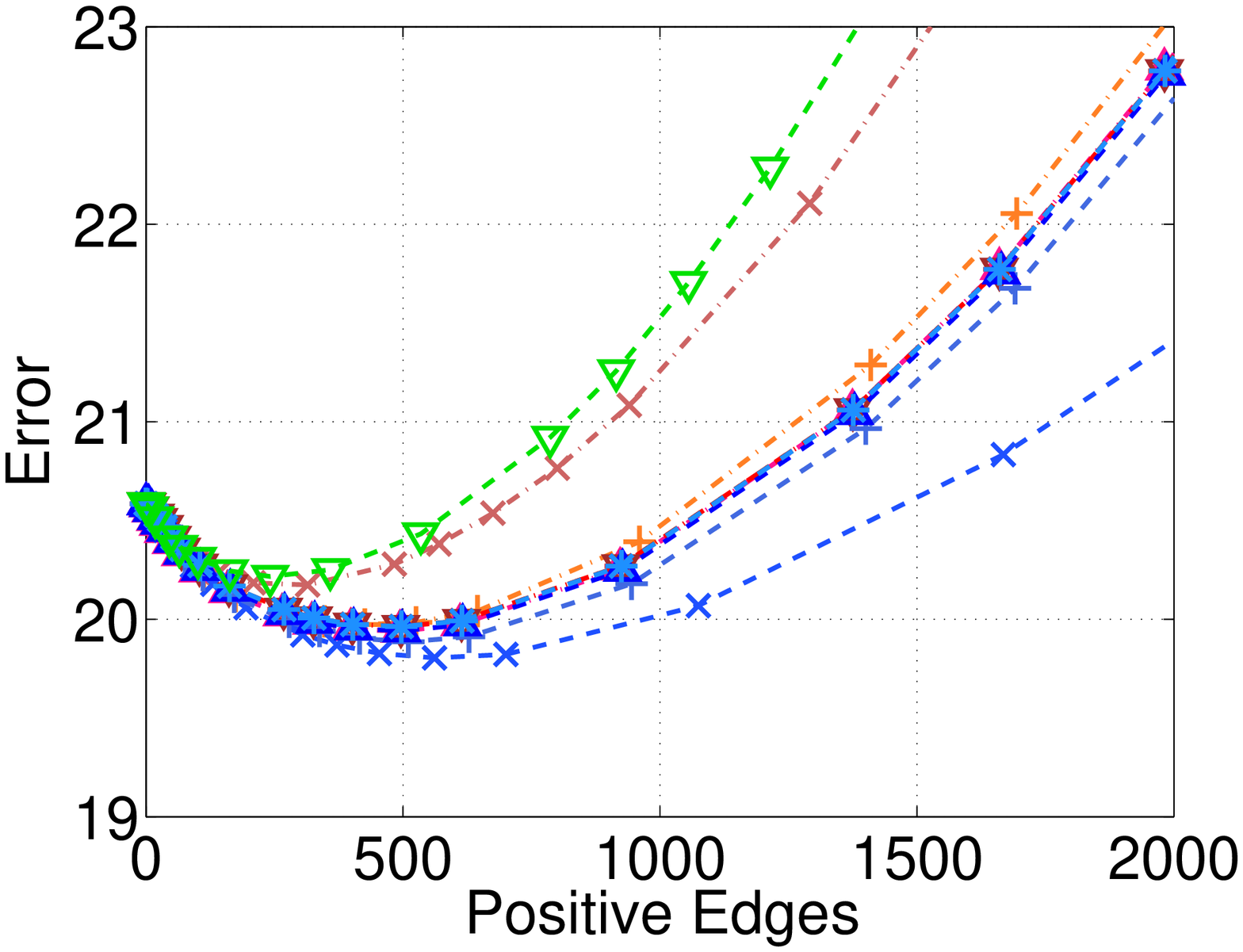}

(b) {CNJGL/GGL/GL:}\\[0.1in]
\noindent
\makebox[1\linewidth][l]{\hspace{\dimexpr-\fboxsep-\fboxrule\relax}
\fbox {\parbox{0.97\linewidth}{ 
\fontsize{8}{12}\selectfont
\textcolor{color1}{-  - $\mathsmaller{\times}$ - -}  \hspace{0.001mm} CNJGL $\lambda_2 = 0.3n$
\hspace{1.2 mm}\textcolor{color2}{-  - $\mathsmaller{+}$ - -} \hspace{0.001mm} CNJGL $\lambda_2 = 0.6n$
\hspace{1.2mm} \textcolor{color3}{-  - $\mathsmaller{\triangle}$  - -}  \hspace{0.001mm} CNJGL $\lambda_2 = 1.0n$
\hspace{1.2mm} \textcolor{color4}{-  - $\ast$ - -} \hspace{0.001mm} CNJGL $\lambda_2 = 1.5n$

\fontsize{8}{12}\selectfont
\textcolor{color5}{$\cdot$  - $\mathsmaller{\times}$ - $\cdot$}  \hspace{0.001mm} GGL $\lambda_2 = 0.01n$
\hspace{1.2mm} \textcolor{color6}{$\cdot$  - $\mathsmaller{+}$  - $\cdot$} \hspace{0.001mm} GGL $\lambda_2 = 0.03n$
\hspace{1.2mm} \textcolor{color7}{$\cdot$  - $\mathsmaller{\triangle}$ - $\cdot$} \hspace{0.001mm} GGL $\lambda_2 = 0.05n$
\hspace{1mm} \textcolor{color9}{$\cdot$ -  $\mathsmaller{\bigtriangledown}$  - $\cdot$} \hspace{0.001mm} GL

}}} \\ 

\includegraphics[width= 0.32\linewidth,clip]{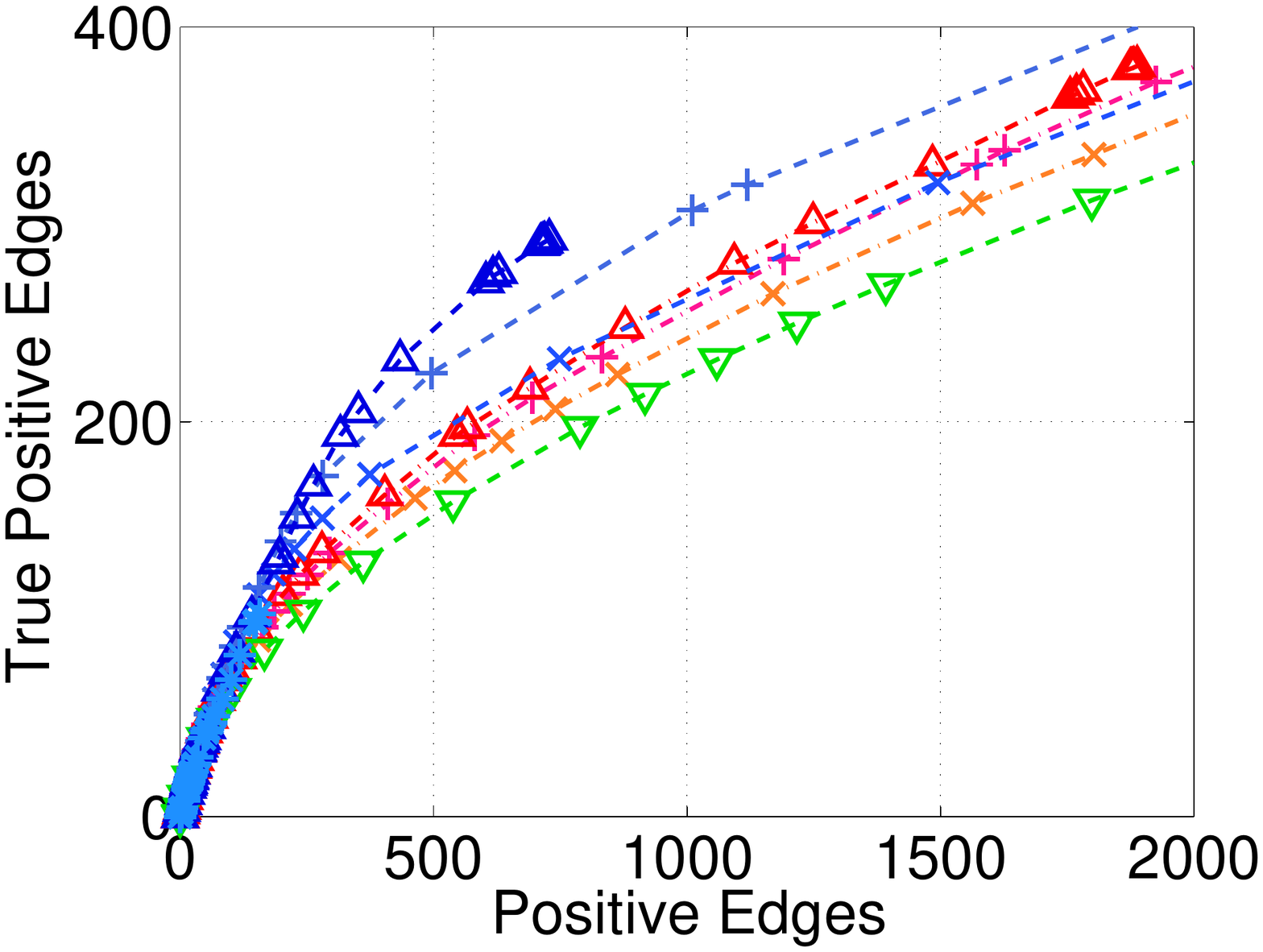}
\includegraphics[width= 0.32\linewidth,clip]{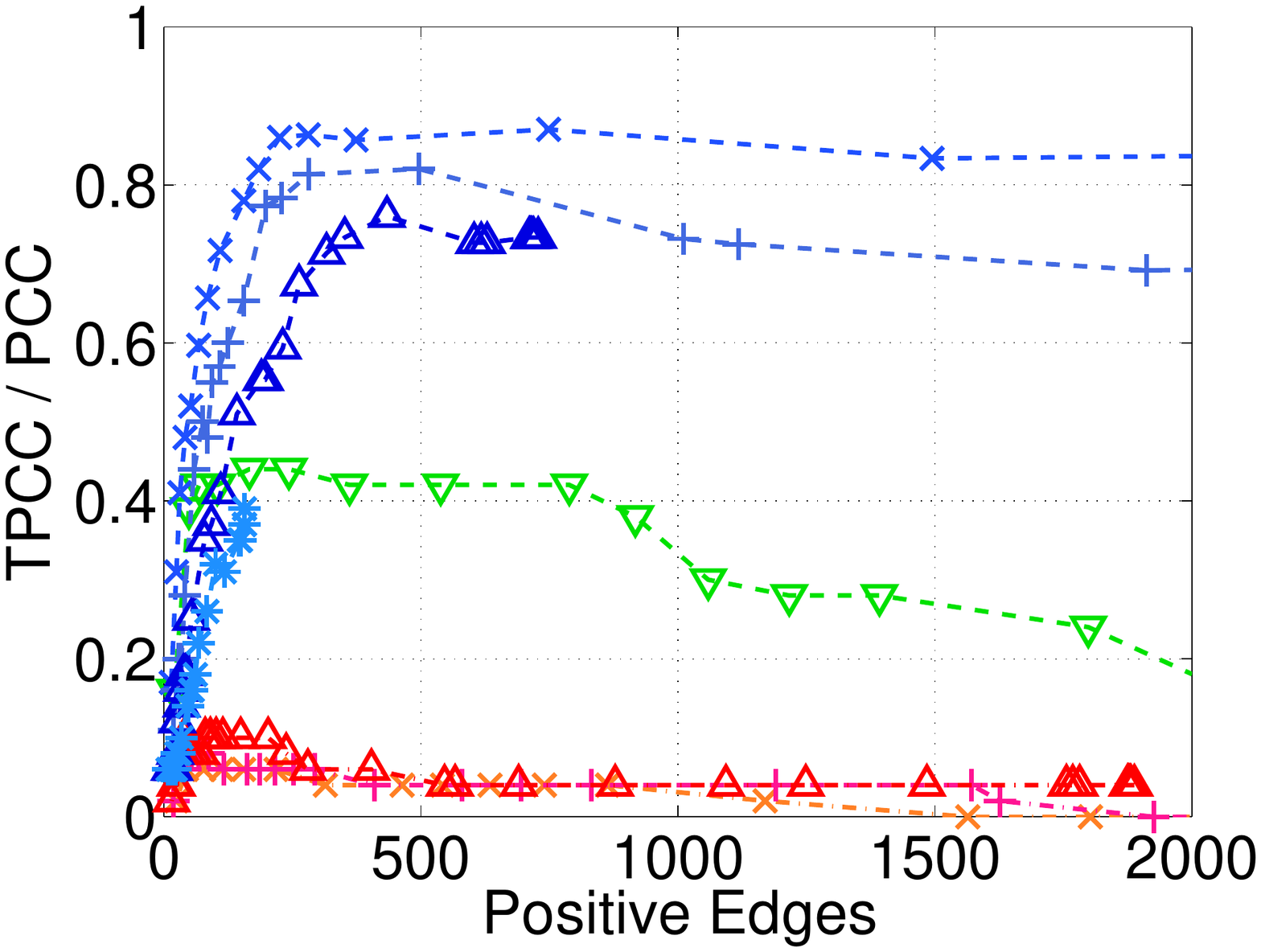}
\includegraphics[width= 0.32\linewidth,clip]{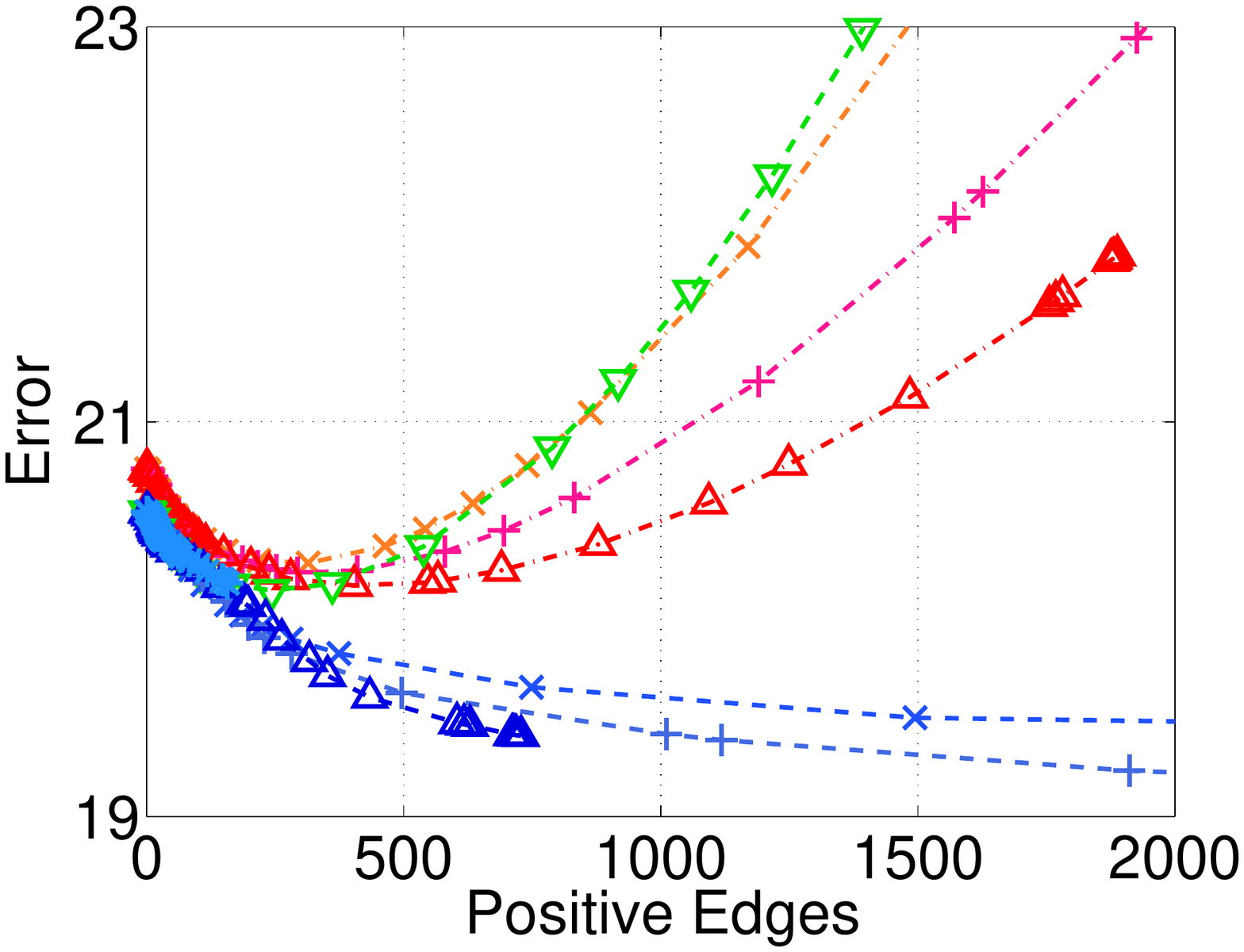}

\caption{\label{fig:community} {Simulation results on community network (Section~\ref{three}) for \emph{(a):} PNJGL with $q=2$, FGL, and GL, and \emph{(b):} CNJGL with $q=2$, GGL, and GL,   with $p =100$ and $n = 50$. Each colored line corresponds to a fixed value of $\lambda_2$, as $\lambda_1$ is varied. Axes are described in detail in Table \ref{tbl:TableMetrics}. 
Results are averaged over 50 random generations of the data.}}
\end{figure}

The simulation results for the set-ups of Sections~\ref{two} and \ref{three} are displayed in Figures \ref{fig:scale-free} and \ref{fig:community},
respectively, for the case $n=50$. The results show that once again, PNJGL and CNJGL substantially outperform the edge-based approaches on the three metrics defined earlier.

\section{Real Data Analysis} \label{sec:realdata}
In this section, we present the results of PNJGL and CNJGL applied to two real data sets: gene expression data set and university webpage data set.

\subsection{Gene Expression Data}

In this experiment, we aim to reconstruct the gene regulatory networks of two subtypes of glioblastoma multiforme (GBM), as well as to identify genes that can improve our understanding of the disease.  Cancer is caused by somatic (cancer-specific) mutations in the genes involved in various cellular processes including cell cycle, cell growth, and DNA repair; such mutations can lead to uncontrolled cell growth. 
We will show that  PNJGL and CNJGL can be used to identify genes that play central roles in the development and progression of cancer.  PNJGL tries to identify genes whose interactions with other genes vary significantly between the subtypes. Such genes are likely to have  {deleterious} somatic  mutations. CNJGL tries to identify genes that have interactions with many other genes in all subtypes. Such genes are likely to play an important role in controlling other genes' expression, and are typically called \emph{regulators}. 

We applied the proposed methods to a publicly available gene expression data set that measures mRNA expression levels of 11,861 genes in 220 tissue samples from patients with GBM \citep{cancer2012}. 
{The raw gene expression data were generated using the Affymetrix GeneChips technology. We downloaded the raw data in .CEL format from the The Caner Genome Atlas (TCGA) website. The raw data were normalized by using the Affymetrix  }\verb=MAS5= {algorithm, which has been shown to perform well in many studies \citep{array-norm}. The data were then log2 transformed and batch-effected corrected using the  software }\verb=ComBat= \citep{combat}.
Each patient has one of  four subtypes of GBM---Proneural, Neural, Classical, or Mesenchymal.  We selected two subtypes, Proneural (53 tissue samples) and Mesenchymal (56 tissue samples), that have the largest sample sizes. All analyses were restricted to the corresponding set of 109 tissue samples.

\begin{figure}[!htbp]
\begin{center}
\includegraphics[width= 0.25\linewidth,clip]{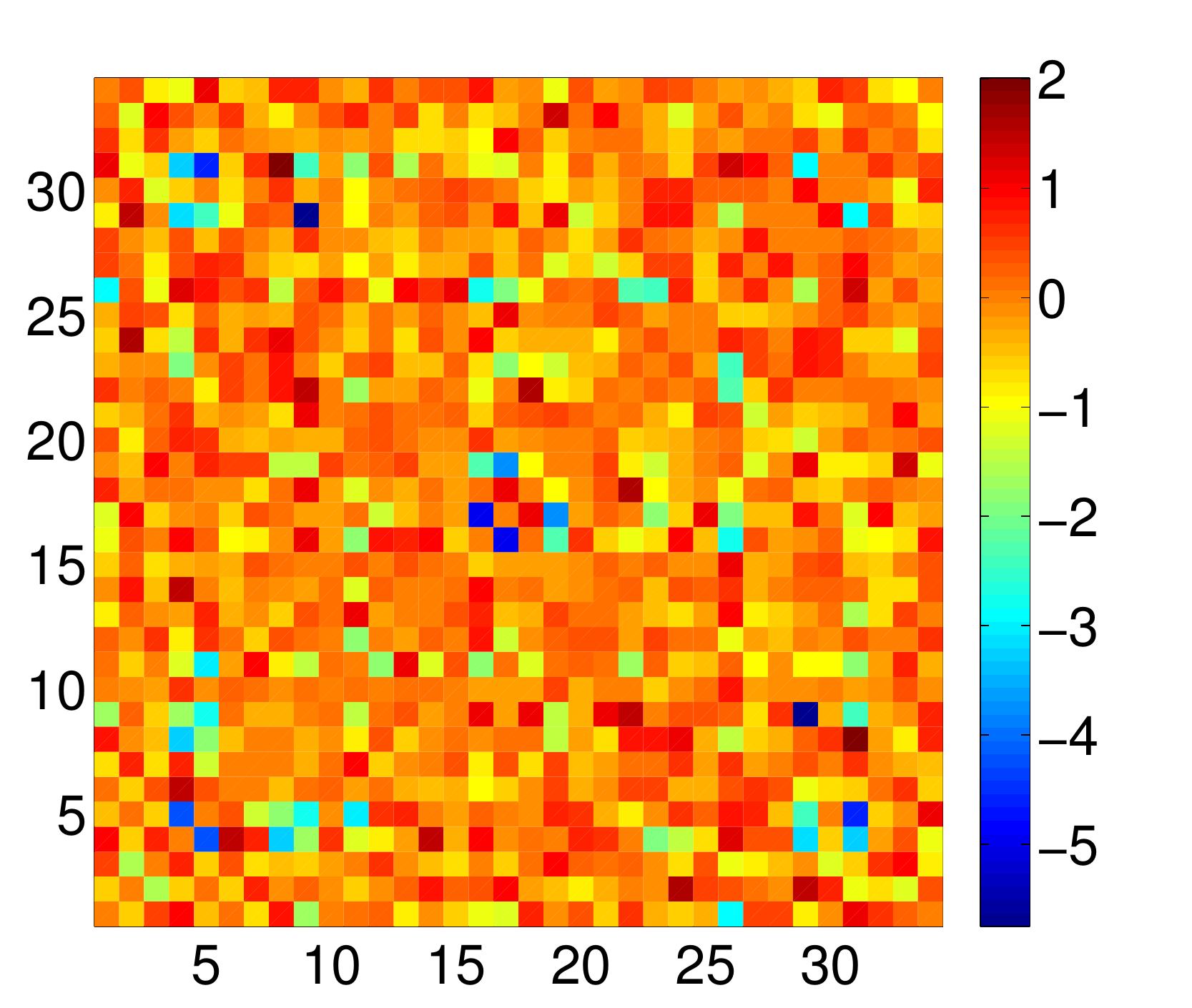}
\includegraphics[width= 0.25\linewidth,clip]{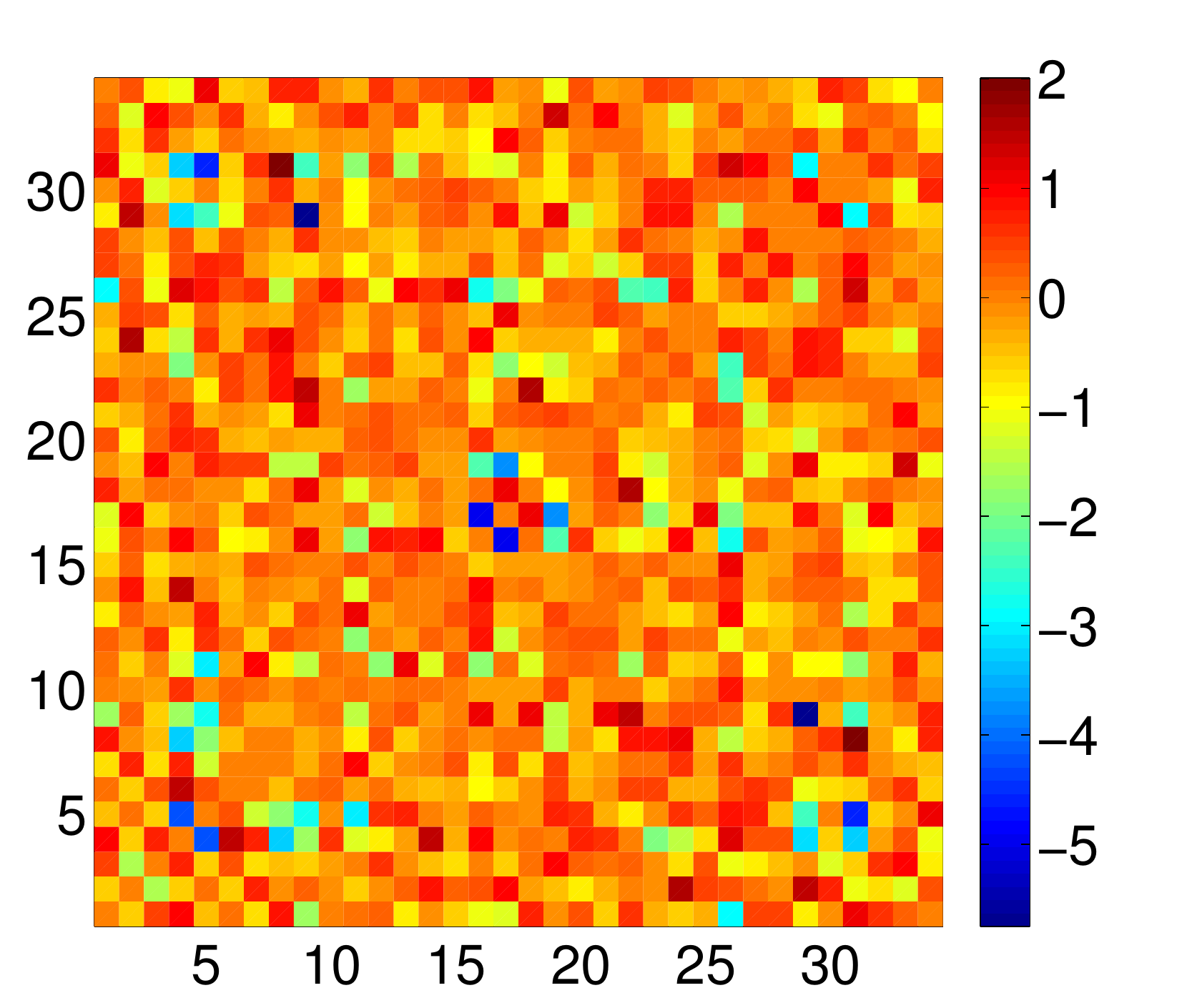}
\includegraphics[width= 0.25\linewidth,clip]{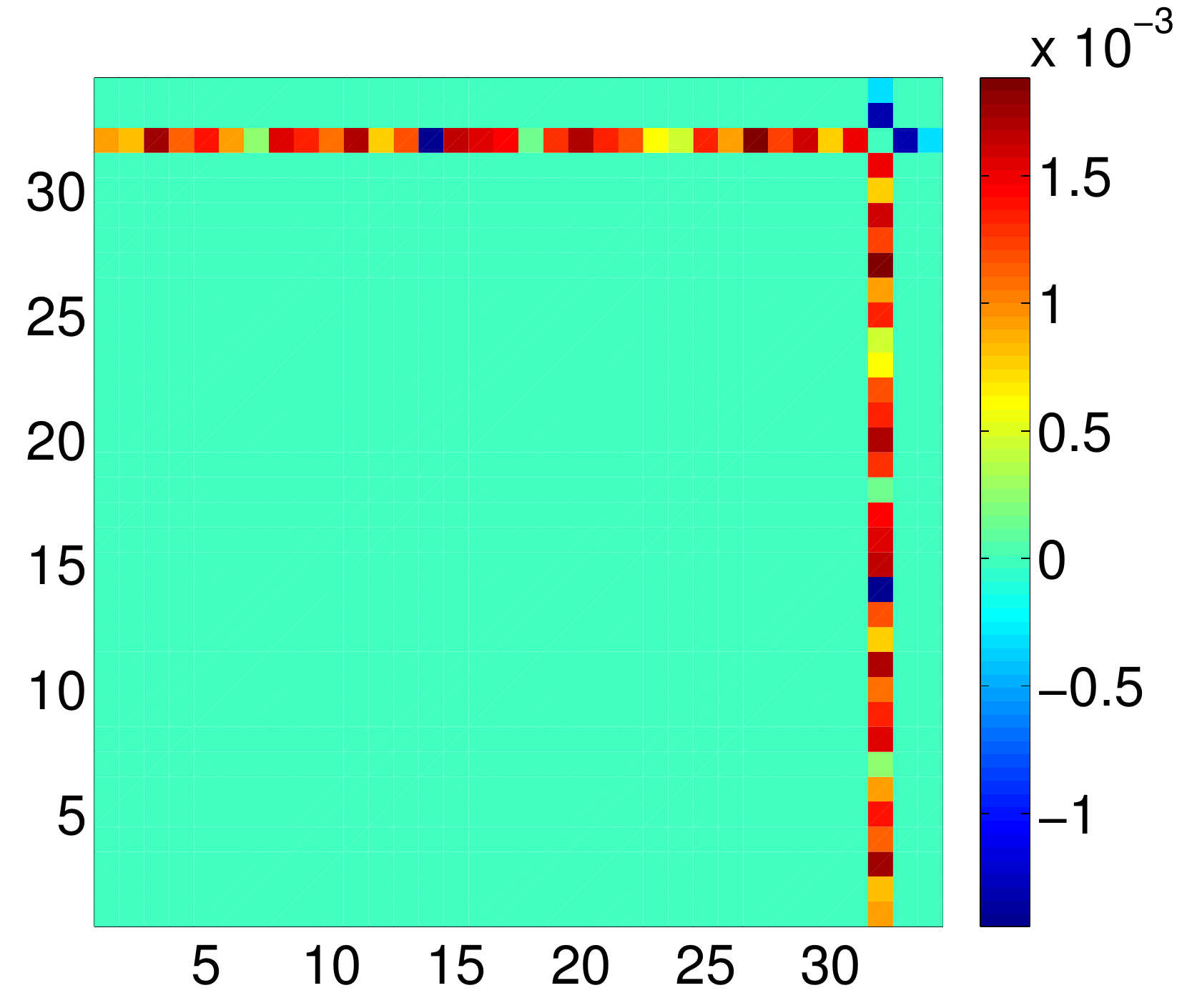} \\
(a) \hspace{33mm} (b) \hspace{33mm} (c)
\end{center}
\caption{\label{fig:PNJGL-real} GBM data analysis for PNJGL with $q = 2$. The sample covariance matrices $S^1$ and $S^2$ were generated from samples with two cancer subtypes, with sizes $n_1 = 53$ and $n_2 = 56$. Only the 34 genes contained in the {Reactome ``TCR Signaling"} pathway were included in this analysis. Of these genes, three are frequently mutated in GBM: PTEN, PIK3R1, and PIK3CA. These three genes correspond to the last three columns in the matrices displayed (columns 32 through 34).  PNJGL was performed with $\lambda_1 = 0$ and $\lambda_2 = 2$. We display \emph{(a):} the estimated matrix ${\hatvarone}$; \emph{(b):} the estimated matrix ${\hatvartwo}$; and \emph{(c):} the difference matrix  $\hatvarone - \hatvartwo$. The gene PTEN is identified as perturbed.}
\end{figure}

To evaluate PNJGL's ability to identify  genes with somatic mutations, we focused on the following 10 genes that have been suggested to be frequently mutated across the four GBM subtypes \citep{cancer2012}: TP53, PTEN, NF1, EGFR, IDH1, PIK3R1, RB1, ERBB2, PIK3CA, PDGFRA. We then considered inferring the regulatory network of a set of genes that is known to be involved in a single biological process, based on the Reactome database \citep{Reactome}. In particular, we focused our analysis on the ``TCR signaling'' gene set, which contains the largest number of mutated genes.  This gene set contains  34 genes, of which three (PTEN, PIK3R1, and PIK3CA) are in the list of 10 genes suggested to be mutated in GBM.  We  applied PNJGL with $q = 2$ to the resulting 53 $\times$ 34 and 56 $\times$ 34 gene expression data sets, after standardizing each gene to have variance one. As can be seen in Figure \ref{fig:PNJGL-real}, the pattern of network differences  indicates that one of the three highly-mutated genes is in fact perturbed across the two GBM subtypes. The perturbed gene is PTEN, a tumor suppressor gene, and it is known that mutations in this gene are associated with the development and progression of many cancers \citep[see, e.g.,][]{Chalhoub}.

\begin{figure}[!htbp]
\begin{center}
\includegraphics[width= 0.25\linewidth,clip]{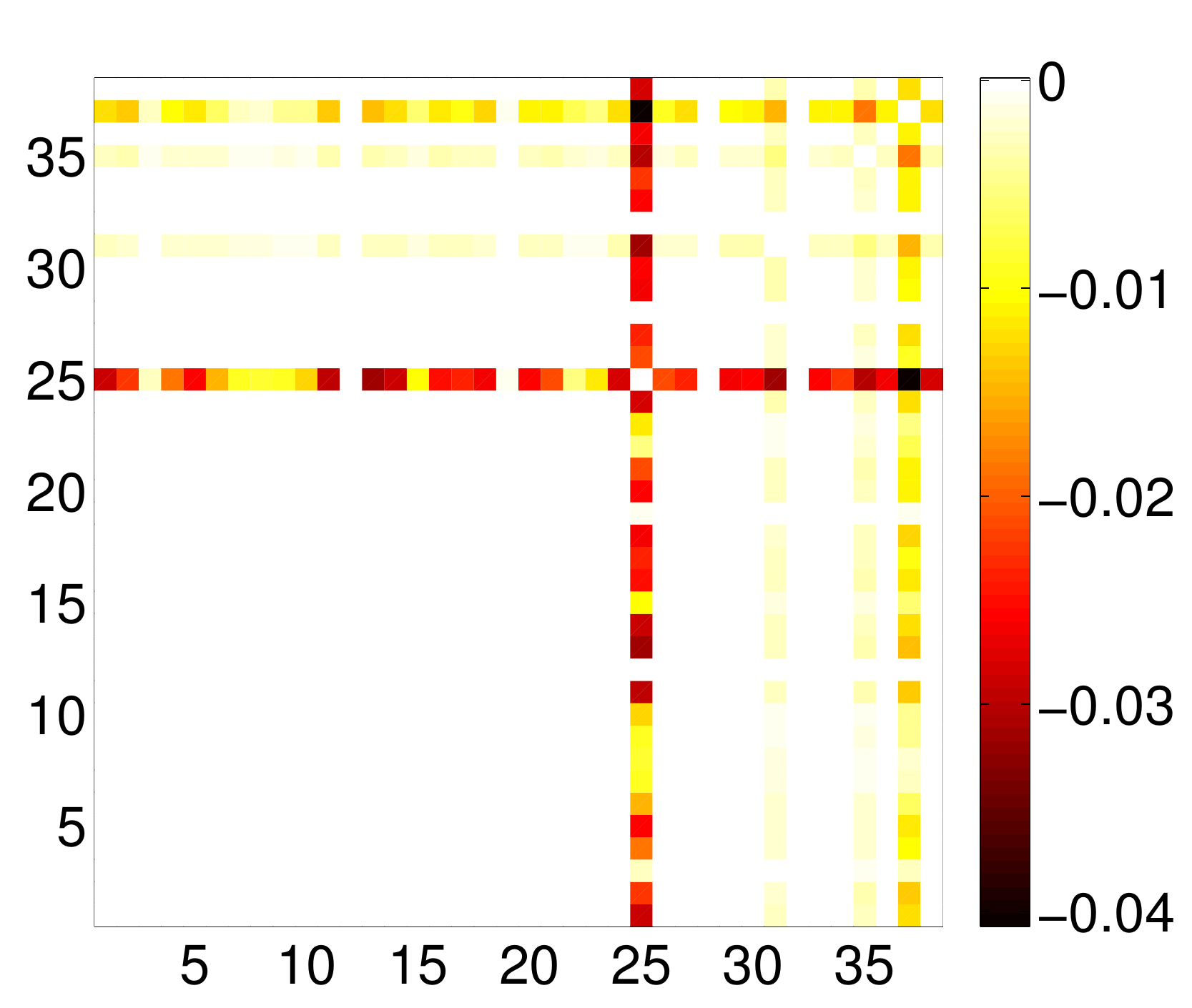}    \hspace{10mm}
\includegraphics[width= 0.25\linewidth,clip]{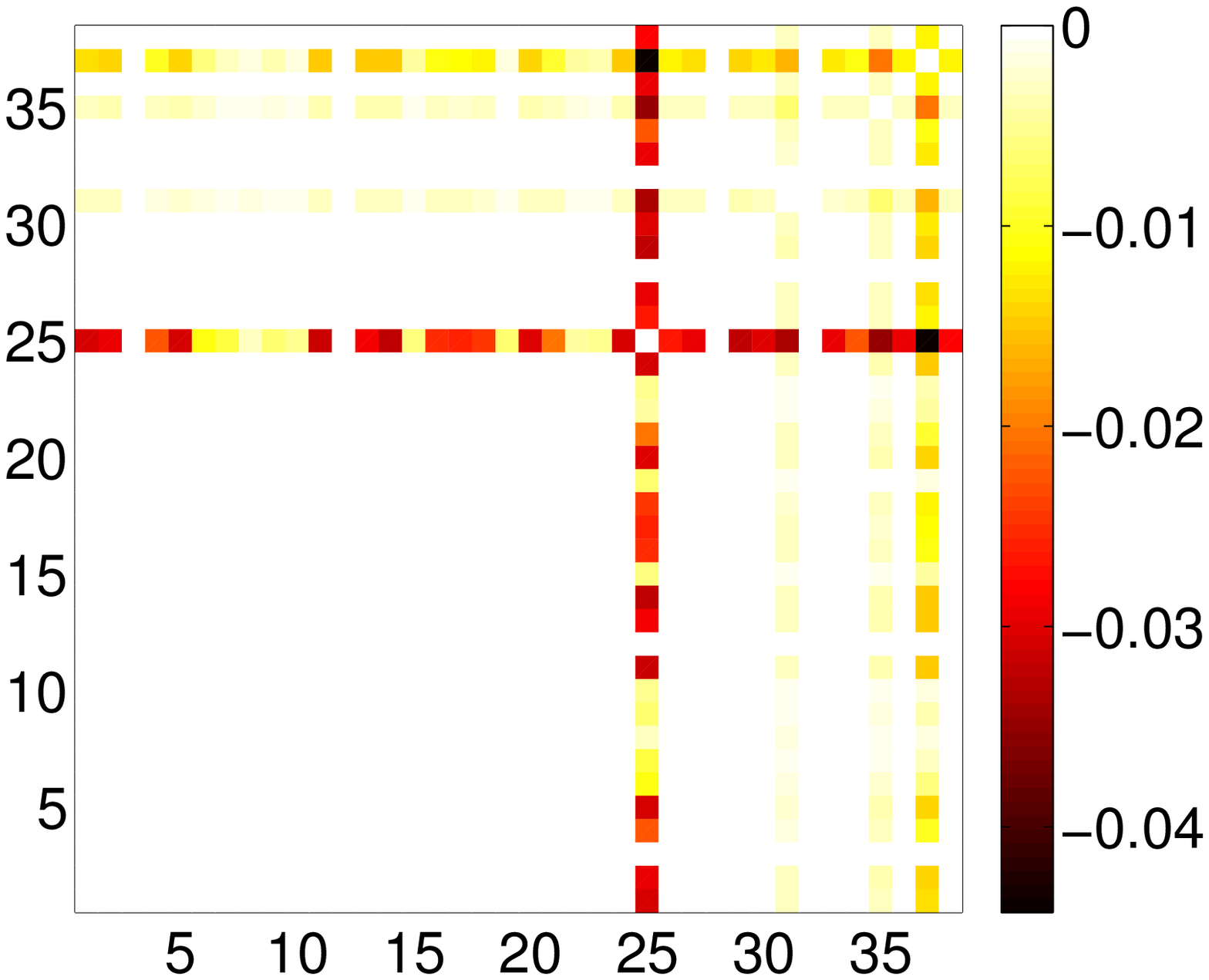}\\
(a) \hspace{45mm} (b)  
\end{center}

\caption{\label{fig:CNJGL-real} GBM data analysis for CNJGL with $q = 2$. The sample covariance matrices $S^1$ and $S^2$ were generated from samples with two cancer subtypes, with sizes $n_1 = 53$ and $n_2 = 56$. Only the 38 genes contained in  the Reactome ``G2/M checkpoints" pathway were included in this analysis. Of these genes, 15 have been previously identified as regulators. These 15 genes correspond to the last 15 columns in the matrices (columns 24 through 38). CNJGL was performed with $\lambda_1 = 13$ and $\lambda_2 = 410$. We display \emph{(a):} the estimated matrix ${\hatvarone}$; \emph{(b):} the estimated matrix ${\hatvartwo}$. Four of the regulator genes are identified  by CNJGL. These genes are CDC6, MCM6, CCNB1, and CCNB2.}
\end{figure}

To evaluate the performance of CNJGL in identifying genes known to be regulators, we used a manually curated list of genes that have been identified as regulators in a previous study \citep{regulators}; this list includes genes annotated as transcription factors, chromatin modifiers, or translation initiation genes.  We then selected a gene set from Reactome, called ``G2/M checkpoints,'' which is relevant to cancer and contains a large number of regulators. This gene set contains 38 genes of which 15 are regulators. We applied CNJGL to the resulting 53  $\times$ 38 and 56  $\times$ 38 gene expression data sets, to see if the 15 regulators tend to be identified as co-hub genes. Figure \ref{fig:CNJGL-real} indicates that all four co-hub genes (CDC6, MCM6, CCNB1 and CCNB2)  detected by CNJGL are known to be regulators.  

\vspace{-2mm} 
\vspace{-2mm} 

\subsection{University Webpage Data}
We applied PNJGL and CNJGL to the university webpages data set from the ``World Wide Knowledge Base" project at Carnegie Mellon University. 
This data set was pre-processed by \cite{webpage2011}. 
The data set describes the number of appearances of various terms, or words, on webpages from the computer science departments of Cornell, Texas, Washington and Wisconsin.  We consider the 544 student webpages, and the 374 faculty webpages.  We standardize the student webpage data so that each term has mean zero and standard deviation one, and we also standardize the faculty webpage data so that each term has mean zero and standard deviation one. Our goal is to identify terms that are perturbed or co-hub between the student and faculty webpage networks. We restrict our analysis to the 100 terms with the largest entropy.

\begin{figure}[!htbp]
\includegraphics[width= 0.5\linewidth,clip]{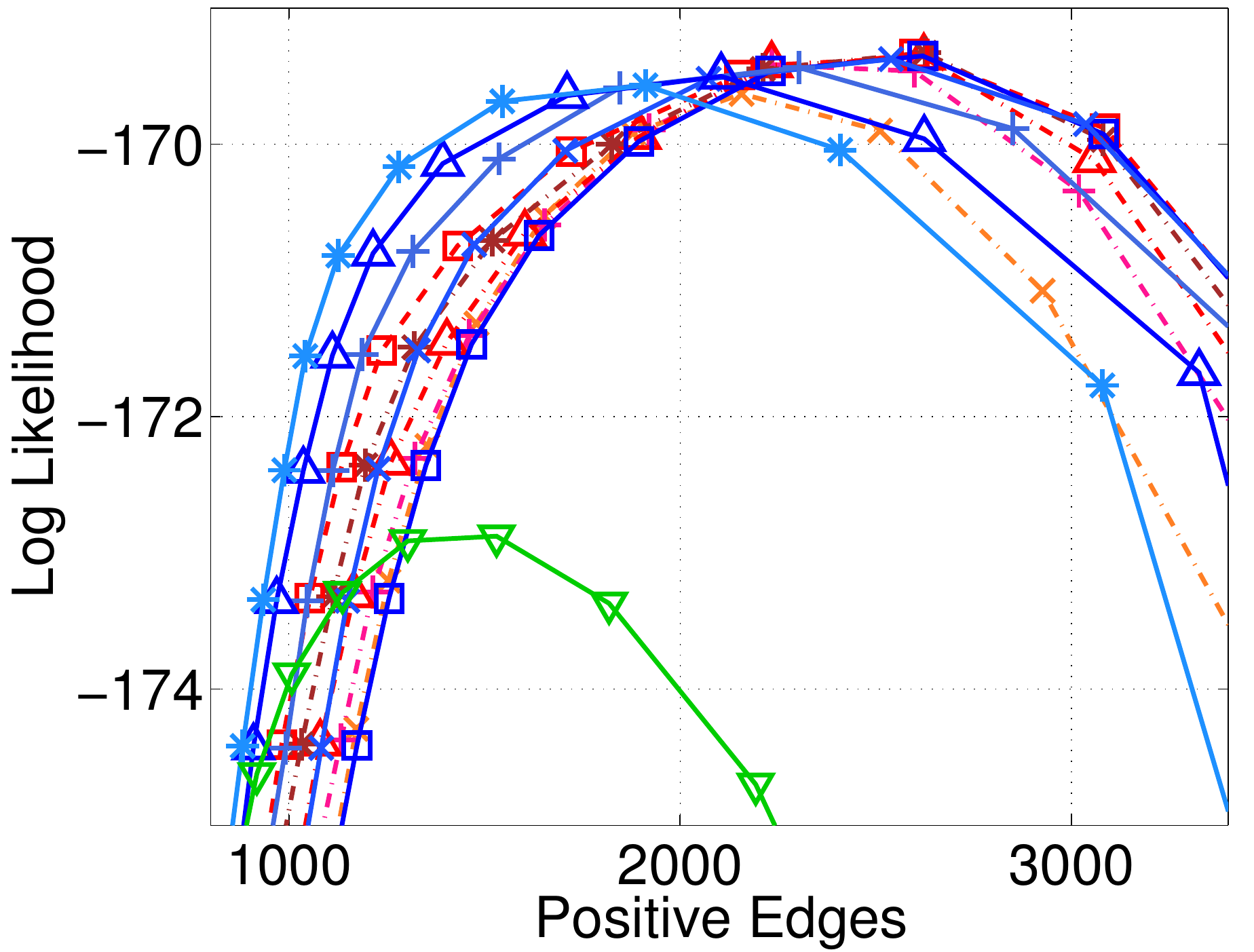}
\includegraphics[width= 0.5\linewidth,clip]{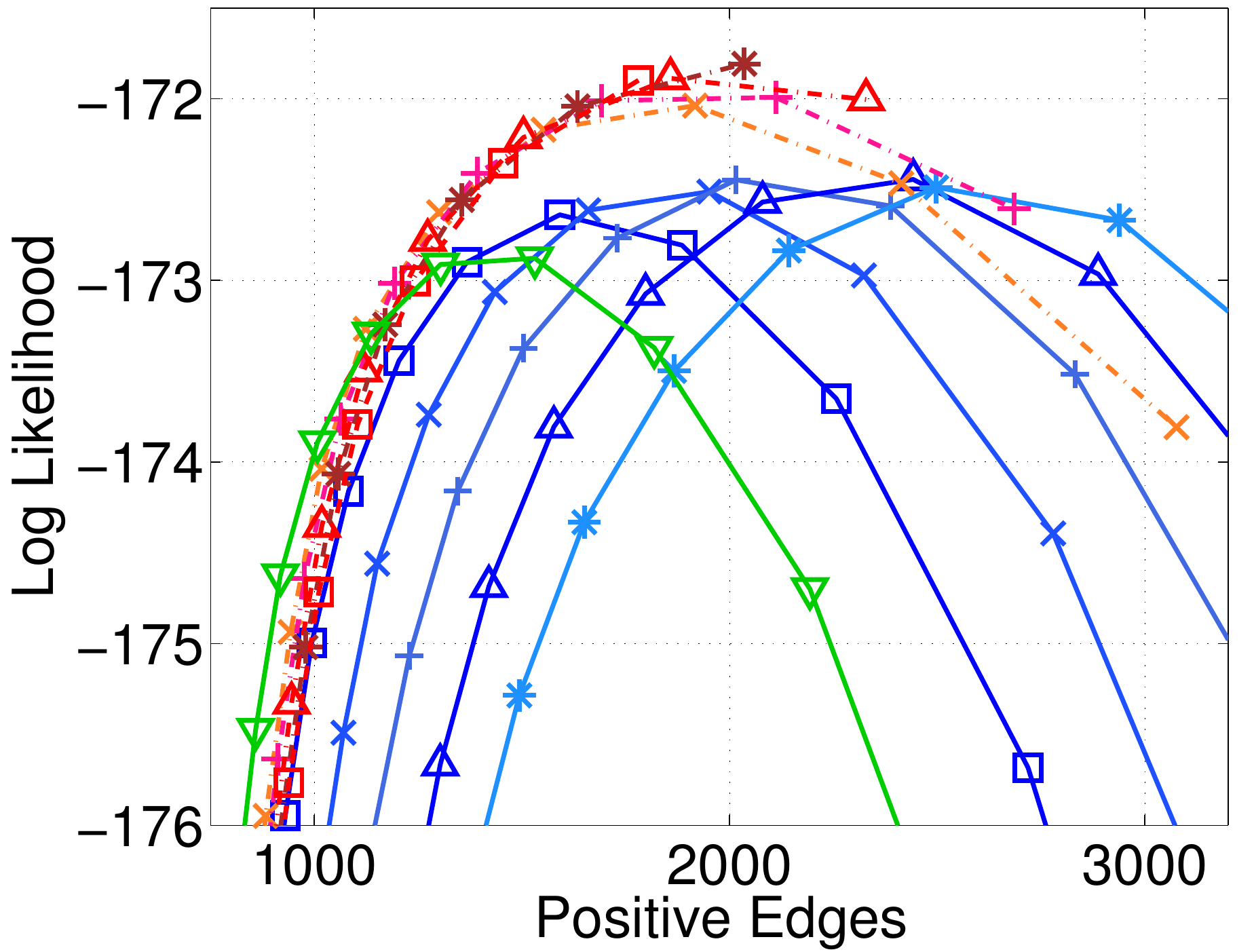} \\
\vfill
\hspace{12mm}
\makebox{\hspace{\dimexpr-\fboxsep-\fboxrule\relax}
\fbox {\parbox{0.392\linewidth}{ 
\fontsize{8}{12}\selectfont 
\noindent
\textcolor{color3} {-  $\mathsmaller{\mathsmaller{\square}}$ -}  \hspace{0.001mm} PNJGL $\lambda_2 = 221$
\hspace{4mm} \textcolor{color5}{$\cdot$  - $\mathsmaller{\times}$ - $\cdot$}  \hspace{0.001mm} FGL $\lambda_2 = 883$\\
\noindent 
\textcolor{color1}{- $\mathsmaller{\times}$ -}  \hspace{0.001mm} PNJGL $\lambda_2 = 258$
\hspace{4mm} \textcolor{color6}{$\cdot$  - $\mathsmaller{+}$ - $\cdot$} \hspace{0.001mm} FGL $\lambda_2 = 1178$ \\
\noindent \textcolor{color2}{- $\mathsmaller{+}$ -} \hspace{0.001mm} PNJGL $\lambda_2 = 295$ 
\hspace{4mm} \textcolor{color7}{$\cdot$  - $\mathsmaller{\mathsmaller{\triangle}}$  - $\cdot$} \hspace{0.001mm} FGL $\lambda_2 = 1325$  \\
\noindent  \textcolor{color3}{- $\mathsmaller{\mathsmaller{\triangle}}$  -}  \hspace{0.001mm} PNJGL $\lambda_2 = 331$
\hspace{4mm} \textcolor{color8}{$\cdot$  - $\ast$ - $\cdot$} \hspace{0.001mm} FGL $\lambda_2 = 1472$\\
\noindent  \textcolor{color4}{- $\ast$ -} \hspace{0.001mm} PNJGL $\lambda_2 = 405$
\hspace{4mm} \textcolor{color7}{$\cdot$  - $\mathsmaller{\mathsmaller{\square}}$ - $\cdot$}  \hspace{0.001mm} FGL $\lambda_2 = 1619$ \\
\hspace{2mm} \hspace{0.5mm} \textcolor{color9}{-  $\mathsmaller{\mathsmaller{\bigtriangledown}}$  -} \hspace{0.001mm} GL
\fontsize{10}{12}\selectfont 
}}} 
\hspace{13.5mm}
\makebox{\hspace{\dimexpr-\fboxsep-\fboxrule\relax}
\fbox {\parbox{0.392\linewidth}{ 
\fontsize{8}{12}\selectfont 
\noindent
\textcolor{color3}{-  $\mathsmaller{\mathsmaller{\mathsmaller{\square}}}$ -}  \hspace{0.001mm} CNJGL $\lambda_2 = 37$
\hspace{5.5mm} \textcolor{color5}{$\cdot$  - $\mathsmaller{\times}$ - $\cdot$}  \hspace{0.001mm} GGL $\lambda_2 = 18$ \\
\noindent \textcolor{color1}{- $\mathsmaller{\times}$ -}  \hspace{0.001mm} CNJGL $\lambda_2 = 74$
\hspace{5.5mm} \textcolor{color6}{$\cdot$  - $\mathsmaller{+}$ - $\cdot$} \hspace{0.001mm} GGL $\lambda_2 = 22$ \\
\noindent \textcolor{color2}{- $\mathsmaller{+}$ -} \hspace{0.001mm} CNJGL $\lambda_2 = 110$ 
\hspace{4mm} \textcolor{color7}{$\cdot$  - $\mathsmaller{\mathsmaller{\triangle}}$  - $\cdot$} \hspace{0.001mm} GGL $\lambda_2 = 26$\\
\noindent \textcolor{color3}{- $\mathsmaller{\mathsmaller{\triangle}}$  -}  \hspace{0.001mm} CNJGL $\lambda_2 = 147$
\hspace{4mm} \textcolor{color8}{$\cdot$  - $\ast$ - $\cdot$} \hspace{0.001mm} GGL $\lambda_2 = 29$\\
\noindent \textcolor{color4}{- $\ast$ -} \hspace{0.001mm} CNJGL $\lambda_2 = 184$
\hspace{4mm} \textcolor{color7}{$\cdot$  - $\mathsmaller{\mathsmaller{\square}}$ - $\cdot$}  \hspace{0.001mm} GGL $\lambda_2 = 33$ \\
\textcolor{color9}{-  $\mathsmaller{\mathsmaller{\bigtriangledown}}$  -} \hspace{0.001mm} GL
\fontsize{10}{12}\selectfont 
}}} \\
\vfill
\hspace{40mm} (a) \hspace{73mm} (b) \\ 
\caption{\label{fig:RD2_CV} {On the webpage data, five-fold cross-validation was performed for \emph{(a):} PNJGL, FGL, and GL;  and  \emph{(b):} CNJGL, GGL, and GL. Each colored line corresponds to a fixed value of $\lambda_2$, as $\lambda_1$ is varied. Positive edges are defined in Table \ref{tbl:TableMetrics}. The cross-validated log likelihood is displayed.}}
\end{figure}

We performed 5-fold cross-validation of the log-likelihood, computed as  $$\log \det \Theta^1 - \trace(S^1 \Theta^1) + \log \det \Theta^2 - \trace(S^2 \Theta^2),$$ for PNJGL, FGL, CNJGL, GGL, and GL, using a range of tuning parameters. The results for PNJGL, FGL and GL are found in Figure \ref{fig:RD2_CV}(a). PNJGL and FGL achieve comparable log-likelihood values. However, for a fixed number of non-zero edges, PNJGL outperforms FGL, suggesting that PNJGL can achieve a comparable model fit for a more interpretable model. Figure \ref{fig:RD2_CV}(b) displays the results for CNJGL, GGL and GL. It appears that PNJGL and FGL provide the best fit to the data.

\begin{figure}[!htbp]
\subfigure{ 
\hspace{10mm}
\includegraphics[width= 0.69\linewidth,clip]{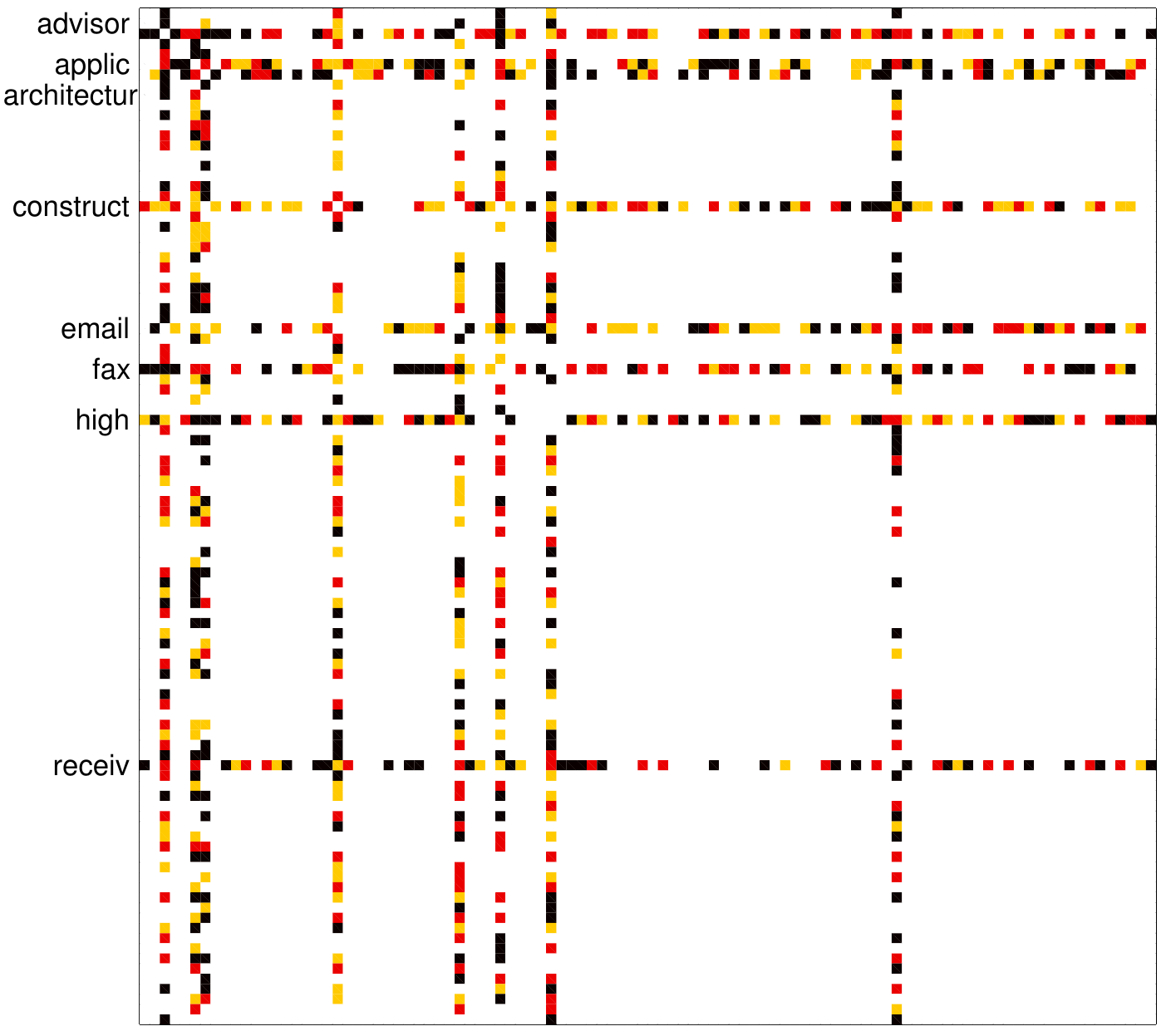} 
}
\subfigure{ 
\fontsize{8}{12}\selectfont
\makebox{\hspace{\dimexpr-\fboxsep-\fboxrule\relax}
\fbox {\parbox{0.16\linewidth}{ 
$\hat\Theta_{ij}^S \neq 0$, \hspace{0.2mm} $\hat\Theta_{ij}^F \neq 0$ 
\hspace{1mm}  {\color{black}\rule{2.2mm}{2.2mm}} \\
$\hat\Theta_{ij}^S \neq 0$, \hspace{0.2mm} $\hat\Theta_{ij}^F = 0$ 
\hspace{1mm} {\color{red}\rule{2.2mm}{2.2mm}}\\
$\hat\Theta_{ij}^S = 0$, \hspace{0.2mm} $\hat\Theta_{ij}^F \neq 0$ 
\hspace{1mm} {\color{color10}\rule{2.2mm}{2.2mm}}\\  
$\hat\Theta_{ij}^S = 0$, \hspace{0.2mm} $\hat\Theta_{ij}^F = 0$ 
\hspace{1mm} $\square$ 
}}}
\fontsize{10}{12}\selectfont 
}
\caption{\label{fig:RD2_PNJGL}  {Student and faculty webpage precision matrices, ${\hat\Theta^S}$ and ${\hat\Theta^F}$, for PNJGL performed with $\lambda_1 = 27$, $\lambda_2 = 381$. Eight perturbed nodes are labeled. The color of each square in the figure indicates whether the corresponding edge is present in both networks, absent in both networks, or present in only the student or only the faculty network. }}
\end{figure}

Given that PNJGL fits the data well, we highlight a particular solution, found in Figure \ref{fig:RD2_PNJGL}. PNJGL is performed with $\lambda_1 = 27$, $\lambda_2 = 381$; these values were chosen because they result in a high log-likelihood in Figure \ref{fig:RD2_CV}(a), and yield an interpretable pair of network estimates. Several perturbed nodes are identified: \emph{advisor, high, construct, email, applic, fax, and receiv}. The student and faculty webpage precision matrices, $\hat{\Theta}^S$ and $\hat{\Theta}^F$, are overlaid in Figure \ref{fig:RD2_PNJGL}.

For example, the perturbed node \emph{receiv}  is connected to the terms \emph{advis, inform,} and \emph{student} among the student webpages. In contrast, among faculty webpages, the phrase \emph{receiv} is connected to \emph{associate} and \emph{faculty}. \\

\section{Discussion}
\label{sec:discussion} We have proposed node-based learning of multiple Gaussian graphical models through the use of two convex formulations, {\npjgl} and {\cnjgl}. These techniques are well-motivated by many real-world applications, such as learning transcriptional regulatory networks in multiple contexts from gene expression data. 
Both of these formulations rely on the use of the {row-column overlap norm} penalty, which when applied to a matrix
encourages  a support that can be expressed as  the {union of a few rows and columns}.
We solve the convex optimization problems that correspond to \PNJGL and \CNJGL using the \ADMM algorithm,
which is more efficient and scalable than standard interior point methods and also first-order methods such as {projected subgradient}.
We also provide  necessary and sufficient conditions on the {regularization parameters} in \CNJGL and \PNJGL
so that the optimal solutions to these formulations are block diagonal, up to a permutation of the rows and columns.
When the sufficient conditions are met, any algorithm that is applicable to these two formulations can be sped up  by  breaking down the optimization problem into smaller subproblems.
Our proposed approaches  lead to  better performance  than two
alternative approaches: learning Gaussian graphical models under the assumption of edge perturbation or shared edges, or simply learning each
model separately.

We next discuss possible directions for future work.
\begin{itemize}

\item We have focused on promoting a row-column structure in either the difference of the networks or in the networks themselves.
      However, the RCON penalty can be generalized {to other forms of structured sparsity}. For instance, we might believe that particular {sets of genes in the same pathway} tend to be simultaneously activated {or perturbed}
      across multiple distinct conditions; a modification of the RCON penalty can be used in this setting.
\item Convergence of the ADMM algorithm in the presence of more than two sets of variable updates has only been addressed partially in the literature.
However, the PNJGL and CNJGL formulations can be rewritten along the lines of an approach given in
 \cite{MaXueZou}, so that only two sets of primal variables are involved, so that convergence
is guaranteed. {We leave for future study an investigation of whether this alternative approach leads to better performance in practice.}

\item Transcriptional regulatory networks involve tens of thousands of genes. Hence it is imperative that our algorithms scale up to large problem sizes.
    {In future work, speed-ups of our \ADMM algorithm as well as adaptations of other fast algorithms such as the accelerated proximal gradient method or second-order methods can be considered.}

\item In Section~\ref{sec:comp-improvements}, we presented a set of conditions that allow us to break up the CNJGL and PNJGL optimization problems into many independent subproblems. However, there is a gap between the necessary and sufficient conditions  that we presented. Making this gap tighter could potentially lead to greater computational improvements.
      \item Tuning parameter selection in high-dimensional unsupervised settings remains an open problem. An existing approach such as stability selection \citep{StabilitySelection} could be applied in
      order to select the tuning parameters $\lambda_1$ and $\lambda_2$ for CNJGL and PNJGL.
      \item The CNJGL and PNJGL formulations are aimed at jointly learning several high-dimensional Gaussian graphical models. These approaches could be modified in order to learn {other types of probabilistic graphical models} \citep[see, e.g.,][]{RavikumarIsing,YangRavikumar}.

\item {It is well-known that adaptive weights can improve the performance of penalized estimation approaches in other contexts (e.g., the adaptive lasso of
\citealp{AdaptiveLasso06} improves over the lasso of \citealp{Ti96}). In a similar manner, the use of adaptive weights may provide improvement over the PNJGL and CNJGL proposals in this paper.
Other options include \emph{reweighted $\ell_1$ norm} approaches that adjust the weights iteratively: one example is the algorithm proposed in \cite{LoboFazelBoyd-portfolio} and further studied in \cite{CandesWakinBoyd08}. This algorithm uses a weight for each variable that is proportional to the inverse  of its value in the previous iteration, yielding improvements over the use of an $\ell_1$ norm. This method can be seen as locally minimizing the sum of the logarithms of the entries, solved by iterative linearization. In general, any of these approaches can be explored for the problems in this paper.
}
\end{itemize}
\verb=Matlab= code implementing  CNJGL and PNJGL  is available at
\url{http://faculty.washington.edu/mfazel/}, \url{http://www.biostat.washington.edu/~dwitten/software.html}, and \\ \url{http://suinlee.cs.washington.edu/software}.

\clearpage
\acks{The authors acknowledge funding from the following sources: NIH DP5OD009145 and NSF CAREER DMS-1252624  to DW,  NSF CAREER  ECCS-0847077 to MF, and Univ. Washington Royalty Research Fund  to DW, MF, and SL.}

\appendix

\section{Dual Characterization of RCON}
\label{sec:appendix_overlap}

\begin{lemma}\label{lemma:dual_rep_RCON}
The dual representation of $\Omega$ is given by
\begin{eqnarray} \label{eq:dual_rep_RCON_2}
\begin{array}{rcll}
{\Omega}(\bThetaone,\ldots,\bThetaK) \meq \MAX_{\Lambda^1,\ldots,\LambdaK \in \R^{p\times p} } & \SUM_{i=1}^K \langle \Lambdai, \bThetai \rangle \\
                                         && \mathrm{subject \; to } & \left\|\left[\begin{array}{c} \Lambdaone + (\Lambdaone)^T \\
                                         \vdots \\ \LambdaK + (\LambdaK)^T \end{array} \right]_j \right\|_* \leq 1 \mbox{ for } j = 1,2,\ldots,p,
\end{array}
\end{eqnarray}
where $\|\cdot\|$ denotes any norm, and $\|\cdot\|_*$ its corresponding dual norm.

\end{lemma}

\begin{proof}
Recall that $\Omega$ is given by
\begin{eqnarray}\label{eq:recall_RCON}
\begin{array}{rc}
\Omega(\bTheta^1,\ldots,\bTheta^K)= \MIN_{V^1,\ldots,V^K \in \R^{p \times p}} & \left\|\left[\begin{array}{c} \Vone \\  \vdots \\ \VK \end{array} \right] \right\| \\
\mbox{subject to} & \bThetai = \Vi + (\Vi)^T, \; i = 1,2,\ldots,K.
\end{array}
\end{eqnarray}
Let $\tZ = \left[\begin{array}{c} \Zone \\ \vdots \\ \ZK \end{array} \right]$ where $\Zk \in \R^{p \times p}$.
Then (\ref{eq:recall_RCON}) is equivalent to
\begin{eqnarray} \label{eq:dual_RCON}
\begin{array}{lll}
\Omega(\bTheta^1,\ldots,\bTheta^K)=

& \MIN_{V^i: \; \bThetai = V^i + (V^i)^T, \; i=1,\ldots,K}
& \MAX_{\tZ: \|\tZ\|_* \leq 1 \quad} \SUM_{i=1}^K \langle \Zi, \Vi \rangle,

\end{array}
\end{eqnarray}
where  $\|.\|_*$ is the dual norm to $\|.\|$.
Since in (\ref{eq:dual_RCON}) the cost function is bilinear in the two sets of variables and the constraints are compact convex sets, by the minimax theorem, we can swap max and min to get

\begin{eqnarray}
 \label{eq:dual_RCON_2}
\begin{array}{lll}
\Omega(\bTheta^1,\ldots,\bTheta^K)=
& \MAX_{\tZ: \|\tZ\|_* \leq 1}
 & \MIN_{V^i: \; \bThetai = V^i + (V^i)^T, \; i=1,\ldots,K}   \SUM_{i=1}^K \langle \Zi,\Vi \rangle
 \end{array}.
\end{eqnarray}
Now, note that the dual to the inner minimization problem with respect to $V^1,\ldots,V^K$ in (\ref{eq:dual_RCON_2}) is given by
\begin{eqnarray} \label{eq:dual_inner}
\begin{array}{rc}
\Maximize_{\Lambda^1,\ldots,\Lambda^K} & \SUM_{i=1}^K \langle \Lambda^i, \bThetai \rangle \\
\mbox{subject to} & \Zi = \Lambda^i + (\Lambda^i)^T, \; i = 1,2,\ldots,K.
\end{array}
\end{eqnarray}
Plugging (\ref{eq:dual_inner}) into (\ref{eq:dual_RCON_2}), the lemma follows.
\end{proof}
 {By definition,}
the subdifferential of $\Omega$ is given by
the set of all $K$-tuples $(\Lambdaone,\ldots,\LambdaK)$ that are optimal solutions to problem (\ref{eq:dual_rep_RCON_2}).
Note that if $(\Lambdaone,\ldots,\LambdaK)$ is an optimal solution to (\ref{eq:dual_rep_RCON_2}),
then any $(\Lambda^1 + Y^1,\ldots,\Lambda^K+Y^K)$ with skew-symmetric matrices  $Y^1,\ldots,Y^K$ is also an optimal solution. 

\section{Proof of Theorem~\ref{thm:PNJGL_nec}}
\label{app:pnjgl_nec}

The optimality conditions for the PNJGL optimization problem (\ref{eqn:sjgl}) with $K=2$ are given by
\begin{eqnarray} 
-n_1(\varone)^{-1} + n_1\tS^1 + \lambda_1 \Gamma^1 + \lambda_2 \Lambda&=&0, \label{eq:opt_PNJGL} \\
-n_2(\vartwo)^{-1} + n_2\tS^2 + \lambda_1 \Gamma^2 - \lambda_2 \Lambda&=&0, \label{eq:opt_PNJGL2}
\end{eqnarray}
where $\Gamma^1$ and $\Gamma^2$ are subgradients of $\|\varone\|_1$ and $\|\vartwo\|_1$, and $(\Lambda, - \Lambda)$ is a subgradient of $\Omega_q(\varone - \vartwo)$. (Note that $\Omega_q(\varone - \vartwo)$ is a composition of $\Omega_q$ with the linear function $\varone-\vartwo$, and apply the chain rule.)
Also note that the right-hand side of the above equations is a zero matrix of size $p\times p$.

Now suppose that $\varone$ and $\vartwo$ that solve (\ref{eqn:sjgl}) are supported on $\tT$. Then since $(\varone)^{-1}, (\vartwo)^{-1}$ are supported on $\tT$, we have that
\begin{eqnarray} \label{eq:opt_PNJGL_single}
n_1\tS^1_{\tT^c} + \lambda_1 \Gamma^1_{\tT^c} + \lambda_2 \Lambda_{\tT^c} &=& 0, \nonumber \\
n_2\tS^2_{\tT^c} + \lambda_1 \Gamma^2_{\tT^c} - \lambda_2 \Lambda_{\tT^c} &=& 0.
\end{eqnarray}
Summing the two equations in (\ref{eq:opt_PNJGL_single}) yields
\begin{eqnarray} \label{eq:opt_PNJGL_single_2}
(n_1\tS^1_{\tT^c} + n_2 \tS^2_{\tT^c}) + \lambda_1 (\Gamma^1_{\tT^c} + \Gamma^2_{\tT^c}) = 0.
\end{eqnarray}
It thus follows from (\ref{eq:opt_PNJGL_single_2}) that
\begin{equation} \label{eq:cond_PNJGL_1}
\|n_1\tS^1_{\tT^c} + n_2 \tS^2_{\tT^c}\|_{\infty} \leq \lambda_1 \|\Gamma^1_{\tT^c} + \Gamma^2_{\tT^c}\|_{\infty} \leq 2\lambda_1, 
\end{equation}						
where here $\| \cdot \|_{\infty}$ indicates the maximal absolute element of a matrix, and where the second inequality in (\ref{eq:cond_PNJGL_1}) follows from the fact that the subgradient of 
the $\ell_1$ norm is bounded in absolute value by one.

We now assume, without loss of generality, that the $\Lambda$ that solves (\ref{eq:opt_PNJGL}) and (\ref{eq:opt_PNJGL2}) is symmetric.
(In fact, one can easily show that there exist symmetric subgradients $\Gamma^1$, $\Gamma^2$, and $\Lambda$ that satisfy (\ref{eq:opt_PNJGL}) and (\ref{eq:opt_PNJGL2}).)
Moreover, recall from Lemma~\ref{lemma:dual_rep_RCON} that $\|(\Lambda + \Lambda^T)_j\|_s \leq 1$. Therefore,  $\|\Lambda_j\|_s \leq \frac{1}{2}$. Using Holder's inequality and noting that $\|y\|_1=\langle y,\sgn(y) \rangle$ for a vector $y$, we obtain
\begin{eqnarray} \label{eq:opt_PNJGL_single_4}
\begin{array}{rcl}
\|\Lambda_{\tT^c}\|_1
&=& \langle \Lambda_{\tT^c}, \sgn(\Lambda_{\tT^c})\rangle\\
&\leq& \|\sgn(\Lambda_{\tT^c})\|_q \|\Lambda_{\tT^c}\|_s\\
&\leq& {|\tT^c|}^{\frac{1}{q}}\|\Lambda_{\tT^c}\|_s \\
&\leq& {|\tT^c|}^{\frac{1}{q}}\|\Lambda \|_s \\
&\leq& \frac{1}{2} {|\tT^c|}^{\frac{1}{q}} p^{\frac{1}{s}},
\end{array}
\end{eqnarray}
where the last inequality follows from the fact that
$\|\Lambda \|_s^s=\sum_{j=1}^p \|\Lambda_j\|_s^s \leq p(\frac{1}{2})^s$, and where in (\ref{eq:opt_PNJGL_single_4}), $\| { A} \|_q$ and $\|{A} \|_s$ 
indicate the $\ell_q$ and $\ell_s$ norms of $\mathrm{vec}({A})$ respectively.

From (\ref{eq:opt_PNJGL_single}), we have for each $k \in \{1,2\}$ that
\begin{eqnarray*} \label{eq:opt_PNJGL_single_5}
\begin{array}{rcl}
n_k\|\tS^k_{\tT^c}\|_1
&\leq& \| \lambda_1  \Gamma^k_{\tT^c}  + \lambda_2  \Lambda_{\tT^c} \|_1 \\
 &\leq& \lambda_1 \| \Gamma^k_{\tT^c} \|_1 + \lambda_2 \| \Lambda_{\tT^c} \|_1 \\
		       &\leq& \lambda_1 |\tT^c| + \lambda_2 \frac{{|\tT^c|}^{\frac{1}{q}}{p}^{\frac{1}{s}}}{2},
\end{array}
\end{eqnarray*}
where the last inequality follows from the fact that the elements of $\Gamma^k$ are bounded in absolute value by one, and (\ref{eq:opt_PNJGL_single_4}).
The theorem now follows by noting from (\ref{eq:opt_PNJGL_single}) that for each $k \in \{1,2\}$,
$$ n_k\|\tS^k_{\tT^c}\|_{\infty} 
		 \leq \lambda_1 \|\Gamma^k_{\tT^c}\|_{\infty} + \lambda_2 \|\Lambda_{\tT^c}\|_{\infty} 
		 \leq \lambda_1 + \frac{\lambda_2}{2}.$$

\section{Proof of Theorem \ref{thm:CNJGL_nec}}
\begin{proof}
The optimality conditions for the CNJGL problem (\ref{eqn:cnjgl_form}) are given by
\begin{eqnarray} \label{eq:opt_CNJGL}
-n_k(\vark)^{-1} + n_k\tS^k + \lambda_1 \Gamma^k + \lambda_2 \Lambda^k=0, \quad k=1,\ldots,K,
\end{eqnarray}
where $\Gamma^k$ is a subgradient of $\|\vark\|_1$. Also, the $K$-tuple $(\Lambda^1,\ldots,\Lambda^K)$ is a subgradient of $\Omega_q(\varone - \diag(\varone),\ldots,\varK - \diag(\varK))$, 
and the right-hand side is a $p\times p$ matrix of zeros.  
We can assume, without loss of generality, that the subgradients $\Gamma^k$ and $\Lambda^k$ that satisfy (\ref{eq:opt_CNJGL}) are symmetric, since Lemma~\ref{lemma:dual_rep_RCON} indicates that  if $(\Lambda^1,\ldots,\Lambda^K)$ is a subgradient of $\Omega_q(\varone-\diag(\varone),\ldots,\vark-\diag(\vark))$, then $((\Lambda^1+(\Lambda^1)^T)/2,\ldots,(\Lambda^K+(\Lambda^K)^T)/2)$ is a subgradient as well.  

Now suppose that $\var^1,\ldots,\var^K$ that solve (\ref{eqn:cnjgl_form}) are supported on $T$. Since $(\vark)^{-1}$ is supported on $\tT$ for all $k$, we have
\begin{eqnarray} \label{eq:opt_single_off_4}
n_k\tS^k_{\tT^c} + \lambda_1 \Gamma^k_{\tT^c} + \lambda_2 \Lambda^k_{\tT^c} = 0.
\end{eqnarray}
We use the triangle inequality for the $\ell_1$ norm (applied elementwise to the matrix) to get 
\begin{eqnarray}
\label{eq:opt_single_off_5}
n_k \|\tS^k_{\tT^c}\|_{1} \leq \lambda_1\|\Gamma^k_{\tT^c}\|_1 + \lambda_2 \|\Lambda^k_{\tT^c}\|_1.
\end{eqnarray}
We have $\|\Gamma^k\|_{\infty} \leq 1$ since $\Gamma^k$ is a subgradient of the $\ell_1$ norm, which gives $\|\Gamma^k_{\tT^c}\|_1 \leq |\tT^c|$.

Also $\Lambda^k$ is a part of a subgradient to $\Omega_q$, so by Lemma~\ref{lemma:dual_rep_RCON}, $\|(\Lambda^k + (\Lambda^k)^T)_j\|_s \leq 1 \; \mbox{ for } j \in \{1,2,\ldots,p\}$.
Since $\Lambda^k$ is symmetric, we have that $\|\Lambda^k_j\|_s \leq \frac{1}{2}$.
Using the same reasoning as in (\ref{eq:opt_PNJGL_single_4}) of Appendix \ref{app:pnjgl_nec}, we obtain 
\begin{equation} \label{eq:opt_single_off_8}
\|\Lambda^k_{\tT^c}\|_1 
					     \leq \frac{1}{2} {|\tT^c|}^{\frac{1}{q}}{p}^{\frac{1}{s}}.
\end{equation}

Combining (\ref{eq:opt_single_off_5}) and (\ref{eq:opt_single_off_8}) yields
\begin{eqnarray*}
\begin{array}{rcl}
n_k \|\tS^k_{\tT^c}\|_1 \leq \lambda_1 |\tT^c| + \frac{\lambda_2}{2} {|\tT^c|}^{\frac{1}{q}}{p}^{\frac{1}{s}}.
\end{array}
\end{eqnarray*}
The theorem follows by noting from (\ref{eq:opt_single_off_4}) that
\begin{eqnarray*}
\begin{array}{rcl}
n_k\|\tS^k_{\tT^c}\|_{\infty} \leq \lambda_1 \|\Gamma^k_{\tT^c}\|_{\infty} + \lambda_2 \|\Lambda^k_{\tT^c}\|_{\infty} \leq \lambda_1 + \frac{\lambda_2}{2}.
\end{array}
\end{eqnarray*}

\end{proof}

\section{Proof of Theorem \ref{thm:generic_opt_suff_cond} }
\label{sec:proof-generic-suff-cond}

Assume that the sufficient condition holds. 
In order to prove the theorem, we must show that  
\begin{eqnarray*}
&&\sum_{k=1}^K n_k (-\log\det(\bThetak) + \langle \bThetak, \tS^k \rangle) ) + \lambda_1 \sum_{k=1}^k \| \bThetak \|_1 + \lambda_2 h(\bThetaone,\ldots,\bThetaK) \nonumber \\
&>& \sum_{k=1}^K n_k (-\log\det(\bThetak_T) + \langle \bThetak_T, \tS^k \rangle) ) + \lambda_1 \sum_{k=1}^k \| \bThetak_T \|_1 + \lambda_2 h(\bThetaone_T,\ldots,\bThetaK_T). \label{thmshow}
\end{eqnarray*}

By assumption, 
\begin{equation}
h(\bThetaone,\ldots,\bThetaK)  > h(\bThetaone_T,\ldots,\bThetaK_T).\label{assump}
\end{equation}
We will now show that 
\begin{equation}
n_k \langle \bThetak, \tS^k \rangle +  \lambda_1 \| \bThetak \|_1 \geq n_k \langle \bThetak_T, \tS^k \rangle +  \lambda_1 \| \bThetak_T \|_1, \label{need}
\end{equation}
or equivalently,  that 
\begin{equation}
- n_k \langle \bThetak_{T^c}, \tS^k \rangle \leq   \lambda_1  \| \bThetak_{T^c} \|_1. \label{need2} 
\end{equation}

Note that $\langle \bThetak_{T^c}, \tS^k \rangle  = \langle \bThetak_{T^c}, \tS_{T^c}^k \rangle$. By the sufficient condition, $n_k \| \tS_{T^c}^k \|_{\infty} \leq \lambda_1$.
So
\begin{eqnarray*}
- n_k \langle \bThetak_{T^c}, \tS^k \rangle &=& - n_k \langle \bThetak_{T^c}, \tS_{T^c}^k \rangle  \nonumber \\
& \leq & \| n_k  \tS_{T^c}^k \|_{\infty} \|  \bThetak_{T^c} \|_1 \nonumber \\
& \leq & \lambda_1  \|  \bThetak_{T^c} \|_1.
\end{eqnarray*}
 So (\ref{need2}) holds, and hence (\ref{need}) holds.
 
 Finally, we apply Fischer's inequality, which states that $\det(\bThetak) \leq \det(\bThetak_T)$, and so \begin{equation} -\log \det(\bThetak) \geq -\log \det(\bThetak_T).\label{fischer}\end{equation} 
 Combining (\ref{assump}), (\ref{need}), and (\ref{fischer}), the theorem holds.

\section{Connection Between RCON and  \cite{Obozinski-11}} \label{app:obozinski}
We now show that the RCON penalty with $q=2$  {can be derived from} 
the overlap norm of \cite{Obozinski-11}.
For simplicity, here we restrict ourselves to the RCON with $K=1$. The general case of $K \geq 1$ can be shown via a simple extension of this argument.

Given any symmetric $p \times p$
matrix $\Theta$, let $\Vard$ be the $p \times p$ upper-triangular matrix such that $\Theta = \Vard+\Vard^T$. That is,

\begin{eqnarray}\label{eq:Theta_delta}
\begin{array}{rcl}
(\Vard)_{kl} &=& \left\{\begin{array}{rc} \var_{kl} & \mbox{if } k < l \\
\var_{kk}/2 & \mbox{if } k=l \\
		    0 & \mbox{if } k>l. \end{array} \right.		
\end{array}		
\end{eqnarray}

\begin{figure}

\begin{center}
\begin{subfigure}[$g_1$]
{
\includegraphics[width = 0.15\linewidth,clip]{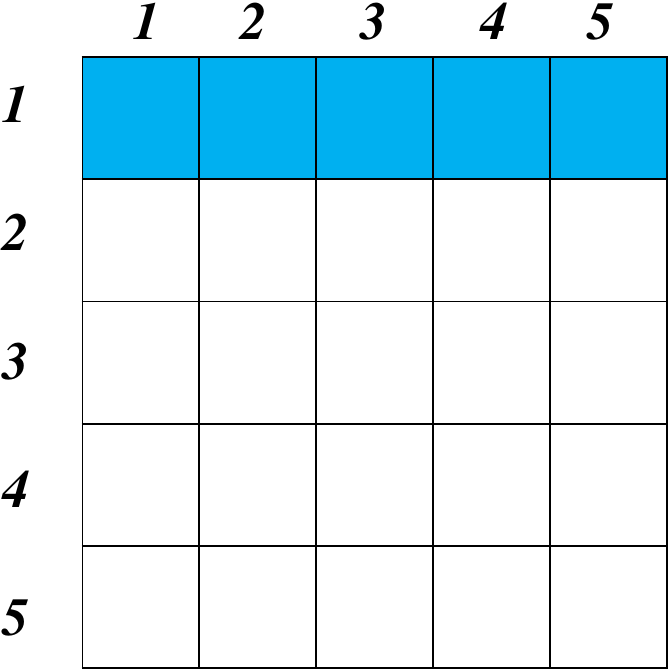}
}
\end{subfigure}
\quad
\begin{subfigure}[$g_2$]
{
\includegraphics[width = 0.15\linewidth,clip]{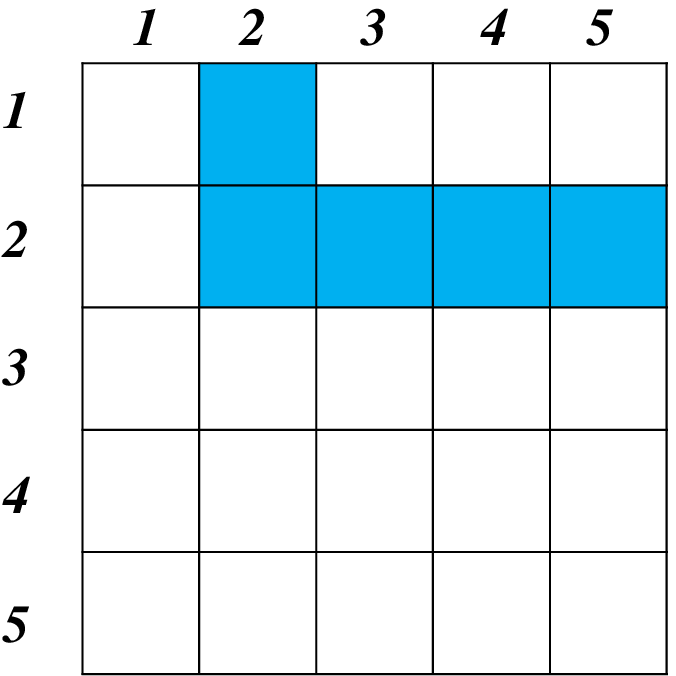}
}
\end{subfigure}
\quad
\begin{subfigure}[$g_3$]
{
\includegraphics[width = 0.15\linewidth,clip]{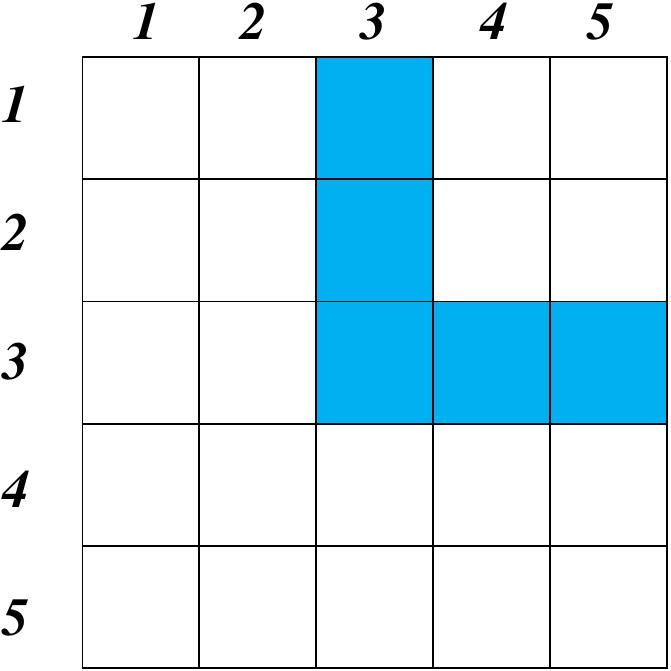}
}
\end{subfigure}\quad
\begin{subfigure}[$g_4$]
{
\includegraphics[width = 0.15\linewidth,clip]{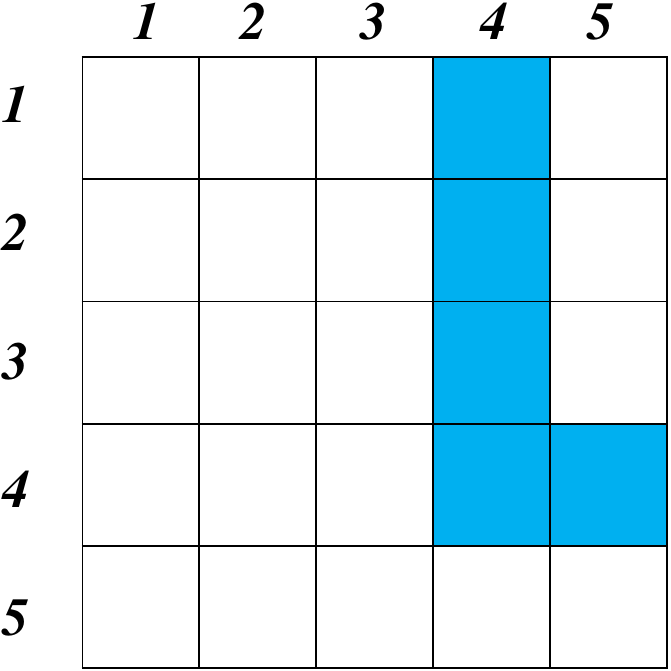}
}
\end{subfigure}\quad
\begin{subfigure}[$g_5$]
{
\includegraphics[width = 0.15\linewidth,clip]{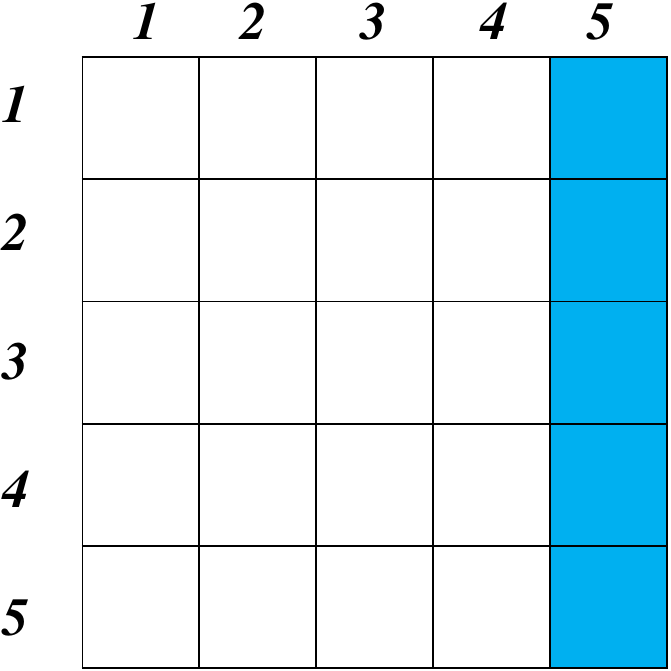}
}
\end{subfigure}
\end{center}
\caption{\label{fig:groups-overlap} Depiction of groups $g_1,\ldots,g_5$ for a $5 \times 5$ matrix. Each group's elements are shown in blue. }
\end{figure}

Now define $p$ groups, $g_1,\ldots,g_p$, each of which contains $p$ variables,
as displayed in   Figure \ref{fig:groups-overlap}.
Note that these groups overlap: if $k \leq l$, then the $(k,l)$ element of a matrix is contained in both the $k$th and $l$th groups.

The overlap norm corresponding to these groups is given by
\begin{eqnarray*} \label{eq:overlap-upper}
\begin{array}{rccl}
\Omega^O(\var) &=& \MIN_{\tV^1,\ldots,\tV^p \in \R^{p \times p}} & \SUM_{j=1}^p \|\tV^j\|_F \\
		   && \mbox{subject to } & \Vard = \SUM_{j=1}^p \tV^j, \; \mbox{supp}(\tV^j) \subseteq g_j,
\end{array}
\end{eqnarray*}
where the relation between $\var$ and $\Vard$ is as in Equation \eqref{eq:Theta_delta}.
We can rewrite this  as
 \begin{eqnarray}
 \begin{array}{rccl}
\Omega^O(\var) &=& \MIN_{\tV^1,\ldots,\tV^p \in \R^{p \times p}} & \SUM_{j=1}^p \|\tV^j\|_F \\
		   && \mbox{subject to } & \var = \SUM_{j=1}^p \tV^j + ( \SUM_{j=1}^p \tV^j)^T, \; \mbox{supp}(\tV^j) \subseteq g_j.
\end{array}	
\label{ugly}	
\end{eqnarray}

Now, define a $p \times p$ matrix $A$ such that
$$A_{ij} = \begin{cases}
(V^j)_{ij}& \mbox{ if }\;\; i<j \\
(V^j)_{ji} &\mbox{ if }\;\; i>j \\
(V^j)_{jj} &\mbox {if }\;\; i=j
\end{cases}.$$
Note that $A + A^T = \SUM_{j=1}^p \tV^j + ( \SUM_{j=1}^p \tV^j)^T$. Furthermore, $\| V^j \|_F = \|A_j \|_2$, where $A_j$ denotes the $j$th column of $A$. So we can rewrite
(\ref{ugly}) as
 \begin{eqnarray*}
 \begin{array}{rccl}
\Omega^O(\var) &=& \MIN_{\tV^1,\ldots,\tV^p \in \R^{p \times p}} & \SUM_{j=1}^p \|A_j\|_2 \\
		   && \mbox{subject to } & \var = A+A^T.
\end{array}	
\label{ugly2}	
\end{eqnarray*}
This is exactly the RCON penalty with $K=1$ {and $q=2$}.
Thus, with a bit of work, we have derived the RCON from the overlap norm \citep{Obozinski-11}.
 {Our penalty is useful because it accommodates groups given by the rows and columns of a symmetric matrix in an elegant and convenient way.}

\section{Derivation of Updates for ADMM Algorithms  }
\label{sec:update_derv_ADMM}

We derive the updates for ADMM algorithm when applied to \PNJGL and \CNJGL formulations respectively. We first begin with the \PNJGL formulation.

\subsection{Updates for ADMM Algorithm for \PNJGL}
Let $\auglag(\varone, \vartwo,\tZ^1,\tZ^2, \tV,\tW,\tF,\tG,\tQ^1,\tQ^2)$ denote the augmented Lagrangian (\ref{eq:PNJGL_aug_lag}).
In each iteration of the ADMM algorithm, each primal variable is updated while holding the other variables fixed. The dual variables are updated using a simple dual-ascent update rule.
Below, we derive the update rules for the primal variables.

\subsubsection{$\varone$ Update}
Note that
{\small
\begin{eqnarray*}
\begin{array}{rcll}
\varone &=& \argmini_{\var} & \auglag(\var,\vartwo,\Zone,\Ztwo,\tV,\tW,\tF,\tG,\Qone,\Qtwo) \\
        &=& \argmini_{\var} & n_1(-\log\det \var) + \rho \left\|\var - \frac{1}{2} \left((\vartwo + \tV + \tW + \Zone) - \frac{1}{\rho}(\tF + \Qone + n_1 \tS^1) \right)\right\|_F^2.
\end{array}
\end{eqnarray*}}
\noindent
Now it follows from the definition of the Expand operator that
\begin{eqnarray*}
\bThetaone \leftarrow \mbox{Expand}\left(\frac{1}{2}(\vartwo + \tV + \tW + \Zone) - \frac{1}{2\rho}(\Qone + n_1\tS^1 + \tF), \rho, n_1 \right).
\end{eqnarray*}
The update for $\bThetatwo$ can be derived in a similar fashion.

\subsubsection{$\Zone$ Update}
\begin{eqnarray*}
\begin{array}{rcll}
\Zone &=& \argmini_{\tZ} & \auglag(\varone,\vartwo,\tZ,\Ztwo,\tV,\tW,\tF,\tG,\Qone,\Qtwo) \\
      &=& \argmini_{\tZ} & \frac{1}{2} \left\|\Zone - (\varone + \frac{\Qone}{\rho})\right\|_F^2 + \frac{\lambda_1}{\rho}\|\Zone \|_1.
\end{array}
\end{eqnarray*}

By the definition of the soft-thresholding operator $\thresh_1$, it follows that
\begin{eqnarray*}
\Zone = \thresh_1\left(\varone + \frac{\Qone}{\rho}, \frac{\lambda_1}{\rho} \right).
\end{eqnarray*}
The update for $\Ztwo$ is similarly derived.

\subsubsection{$\tV$ Update}
\begin{eqnarray*}
\begin{array}{rcll}
\tV &=& \argmini_{\tX} & \auglag(\varone,\vartwo,\Zone,\Ztwo,\tX,\tW,\tF,\tG,\Qone,\Qtwo) \\
    &=& \argmini_{\tX} & \frac{\lambda_2}{2\rho} \SUM_{j=1}^p \|\tX_j\|_q + \frac{1}{2}\left\|\tX - \frac{1}{2}\left((\tW^T + \varone - \vartwo - \tW) + \frac{1}{\rho}(\tF - \tG)\right) \right\|_F^2.
\end{array}
\end{eqnarray*}
By the definition of the soft-scaling operator $\thresh_2$, it follows that
\begin{eqnarray*}
\tV = \thresh_2\left(\frac{1}{2}(\tW^T - \tW + \varone - \vartwo) + \frac{1}{2\rho}(\tF - \tG), \frac{\lambda_2}{2\rho}\right).
\end{eqnarray*}
The update for $\tW$ is easy to derive and we therefore skip it.

\subsection{Updates for ADMM Algorithm for \CNJGL}
Let $\auglag(\{\vari\},\{\tZ^i\},\{\Vic\},  \{\tW^i\},  \{\tF^i\},\{\tG^i\},\{\tQ^i\})$ denote the augmented Lagrangian  (\ref{eq:CNJGL_aug_lag}). 
Below, we derive the update rules for the primal variables $\{\Vic\}$. 
The update rules for the other primal variables are similar to the derivations discussed for PNJGL, 
and hence we omit their derivations.

The update rules for $\Vonec,\Vtwoc,\ldots,\VKc$ are coupled, so we derive them simultaneously.
 Note that
\begin{eqnarray*}\label{eq:Vupdate_CNJGL}
\begin{array}{rcll}
\{\Vic\}_{i=1}^K &=& \argmini_{A^1,\ldots,A^K} & \auglag\left(\{\bThetai\}_{i=1}^K,\{\tZ^i\}_{i=1}^K, \{ A^i\}_{i=1}^K, \{\tW^i\}_{i=1}^K, \{\tF^i\}_{i=1}^K, \{\tG^i\}_{i=1}^K, \{\tQ^i\}_{i=1}^K\right) \\
                         &=& 
                         \argmini_{A^1,\ldots,A^K} & \lambda_2 \SUM_{j=1}^p \left\| \left[  \begin{array}{c} A^1 - \diag(A^1) \\ \vdots \\ A^K - \diag(A^K) \end{array}  \right]_j \right\|_q  + \\
                         && &\rho\SUM_{i=1}^K \left\| A^i - \frac{1}{2} \left( (\Wi)^T + \vari - \Wi  + \frac{1}{\rho}(\Fi - \Gi) \right) \right\|_F^2.
\end{array}
\end{eqnarray*}
Let $\Ci = \frac{1}{2}((\Wi)^T + \vari - \Wi  + \frac{1}{\rho}(\Fi - \Gi))$. Then the update
$$\left[\begin{array}{c}\Vonec  \\ \vdots \\ \VKc \end{array} \right] \leftarrow \thresh_q \left( \left[\begin{array}{c} \Cone - \diag(\Cone) \\ \vdots \\ \CK - \diag(\CK) \end{array} \right],\frac{\lambda_2}{2\rho} \right) 
+ \left[\begin{array}{c} \diag(\Cone)  \\ \vdots \\ \diag(\CK) \end{array} \right]$$
follows by inspection. 

\section{Additional Simulation Results}
\label{sec:Norm_Plots}

\begin{figure}[!htbp]
\begin{center}
\fontsize{9}{12}\selectfont 
\includegraphics[width= 0.15\linewidth,clip]{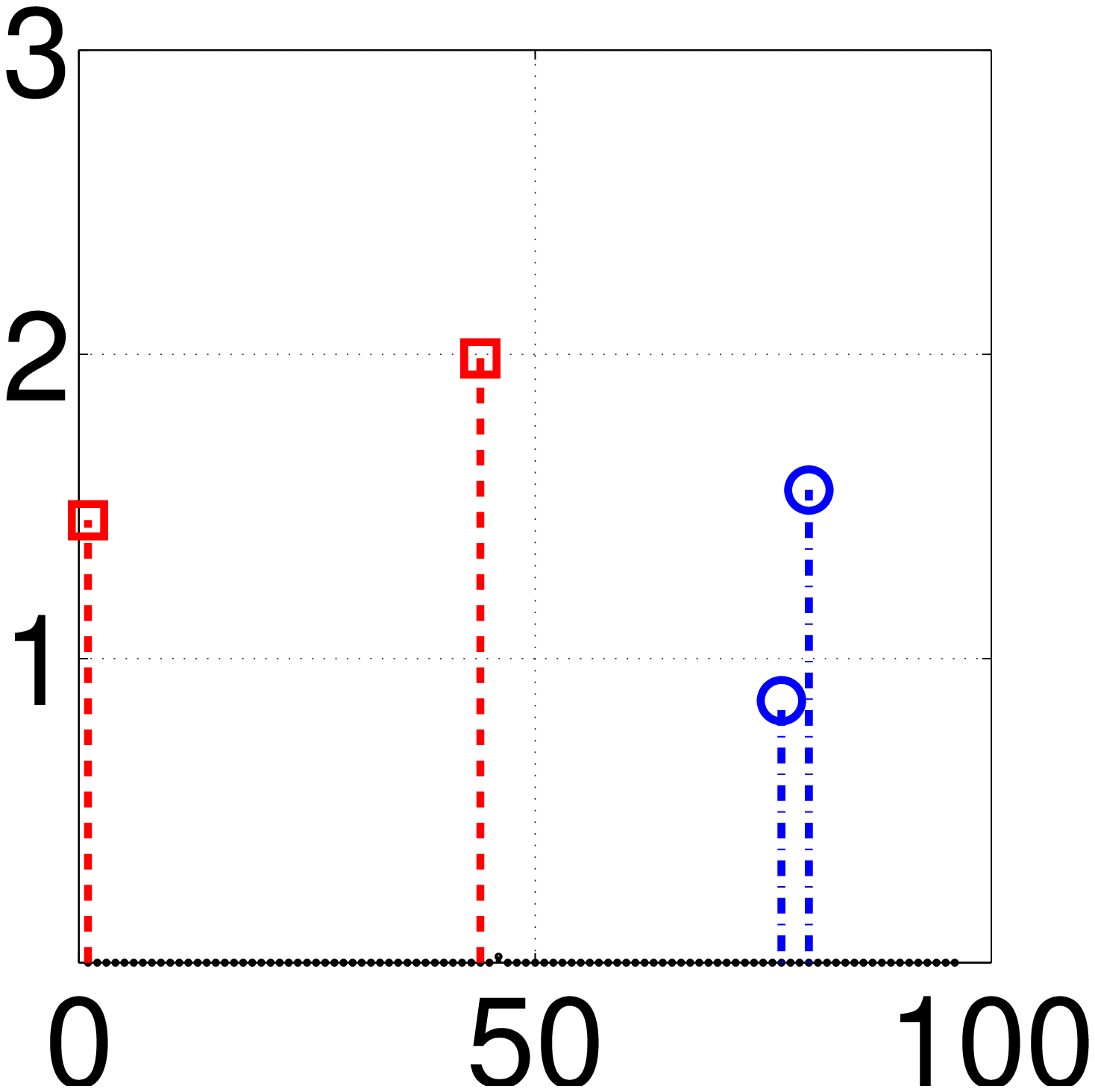}    \hspace{5mm}
\includegraphics[width= 0.15\linewidth,clip]{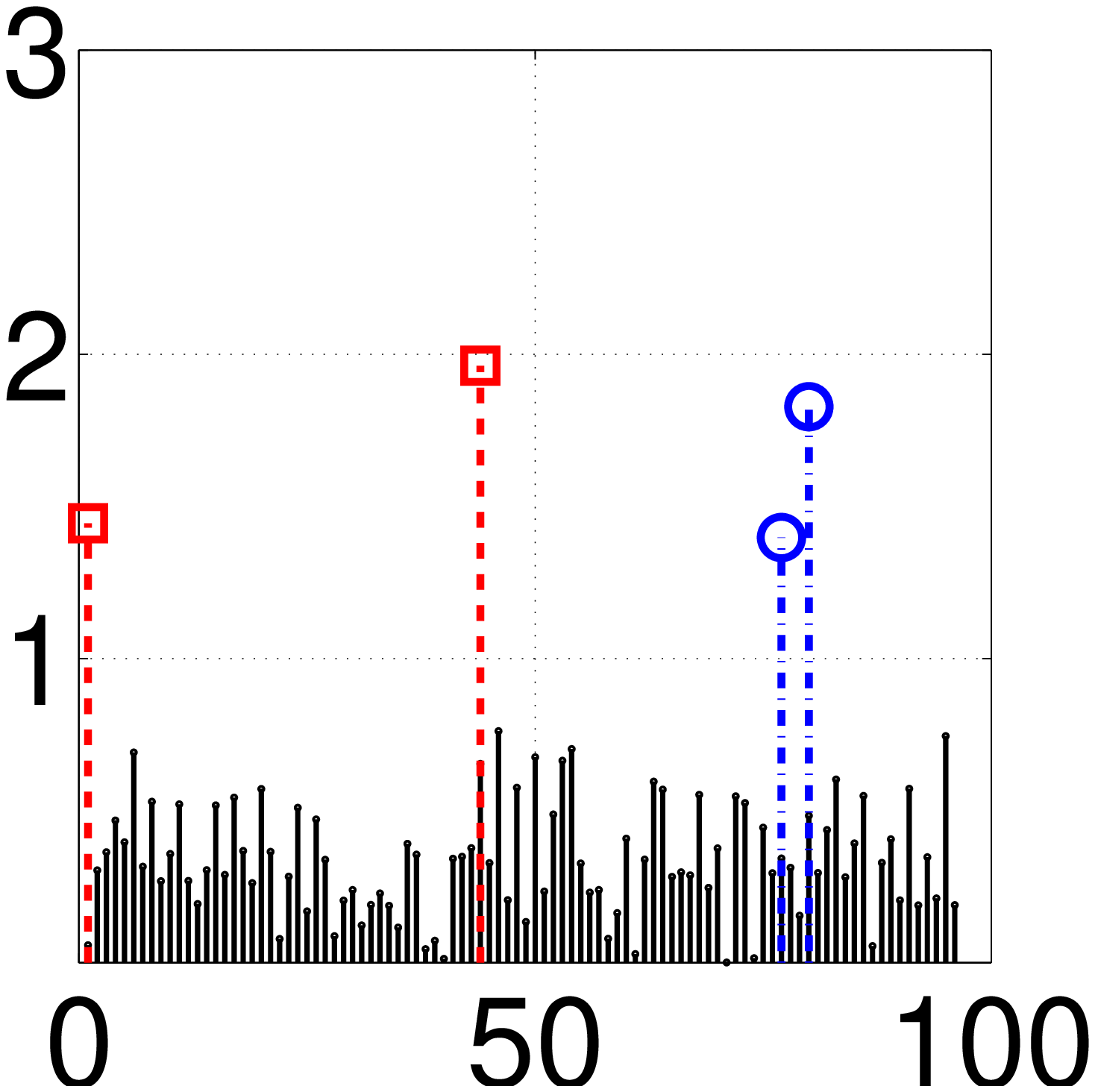}    \hspace{5mm}
\includegraphics[width= 0.15\linewidth,clip]{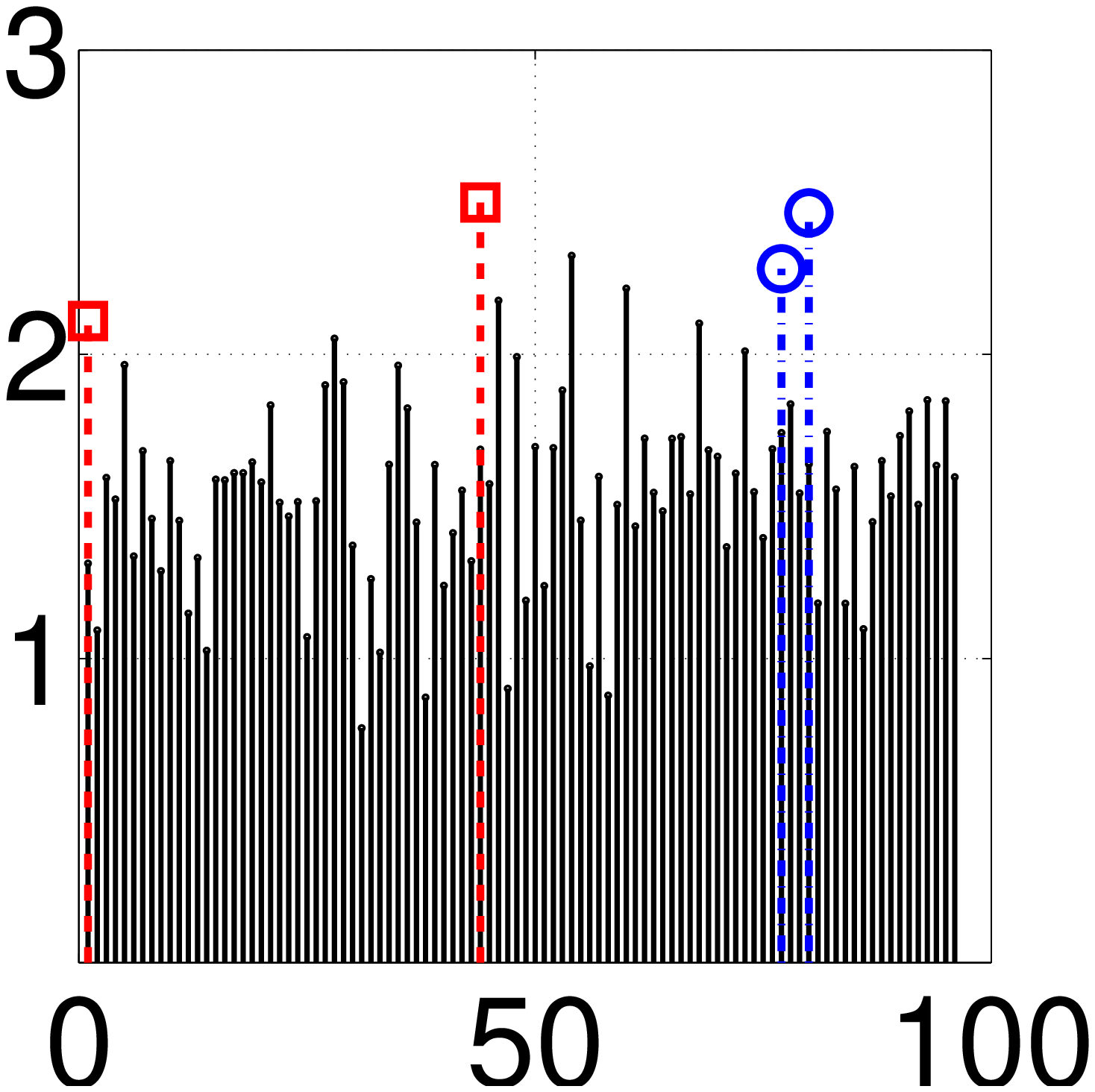}   \\
\hspace{3mm}
(a) $\hat{V}$: PNJGL \hspace{5mm} 
(b) ${(\hatvarone - \hatvartwo)}$: FGL \hspace{5mm} 
(c) ${(\hatvarone - \hatvartwo)}$: GL \\

$ $

\includegraphics[width= 0.15\linewidth,clip]{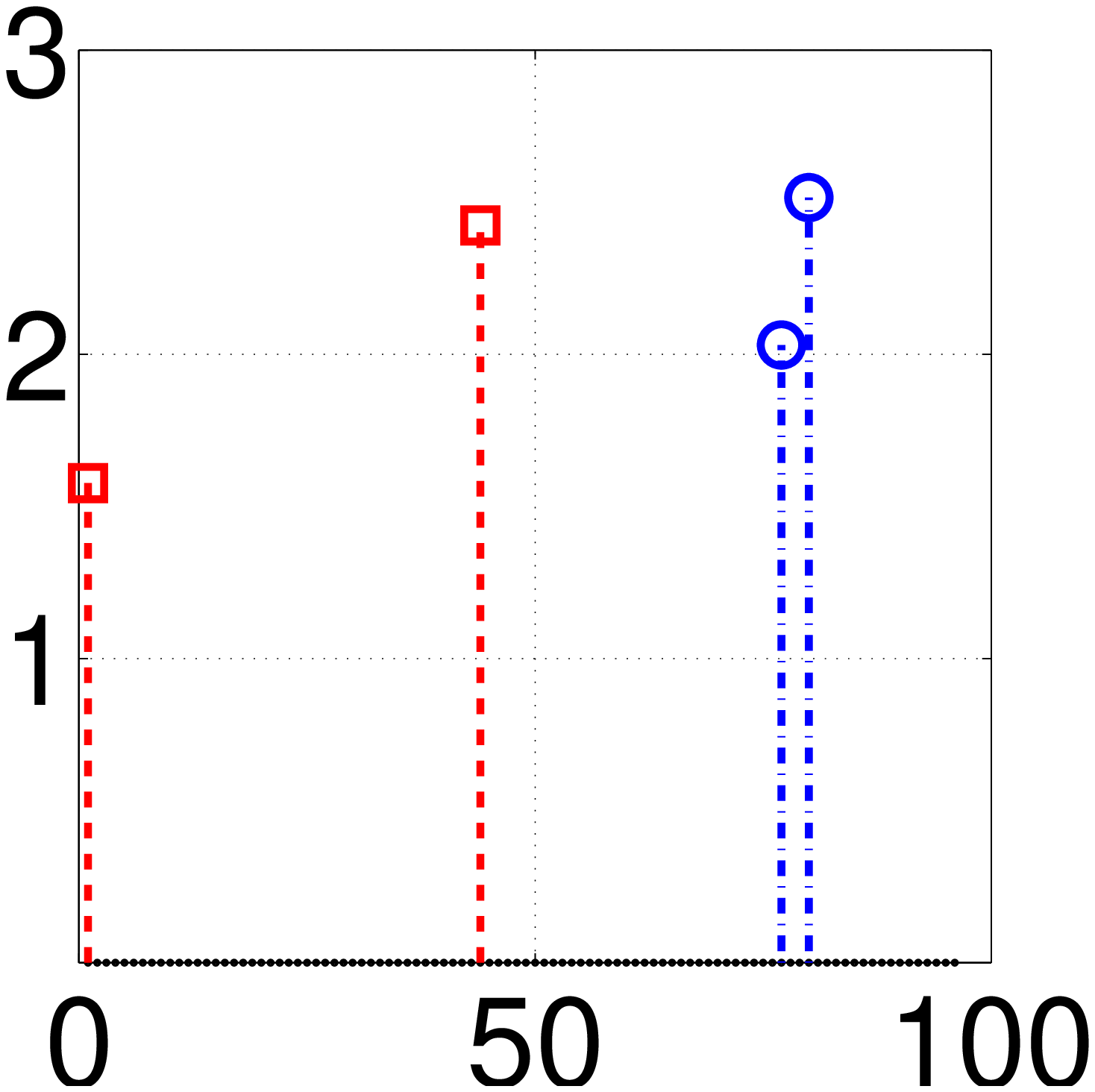}    \hspace{5mm}
\includegraphics[width= 0.15\linewidth,clip]{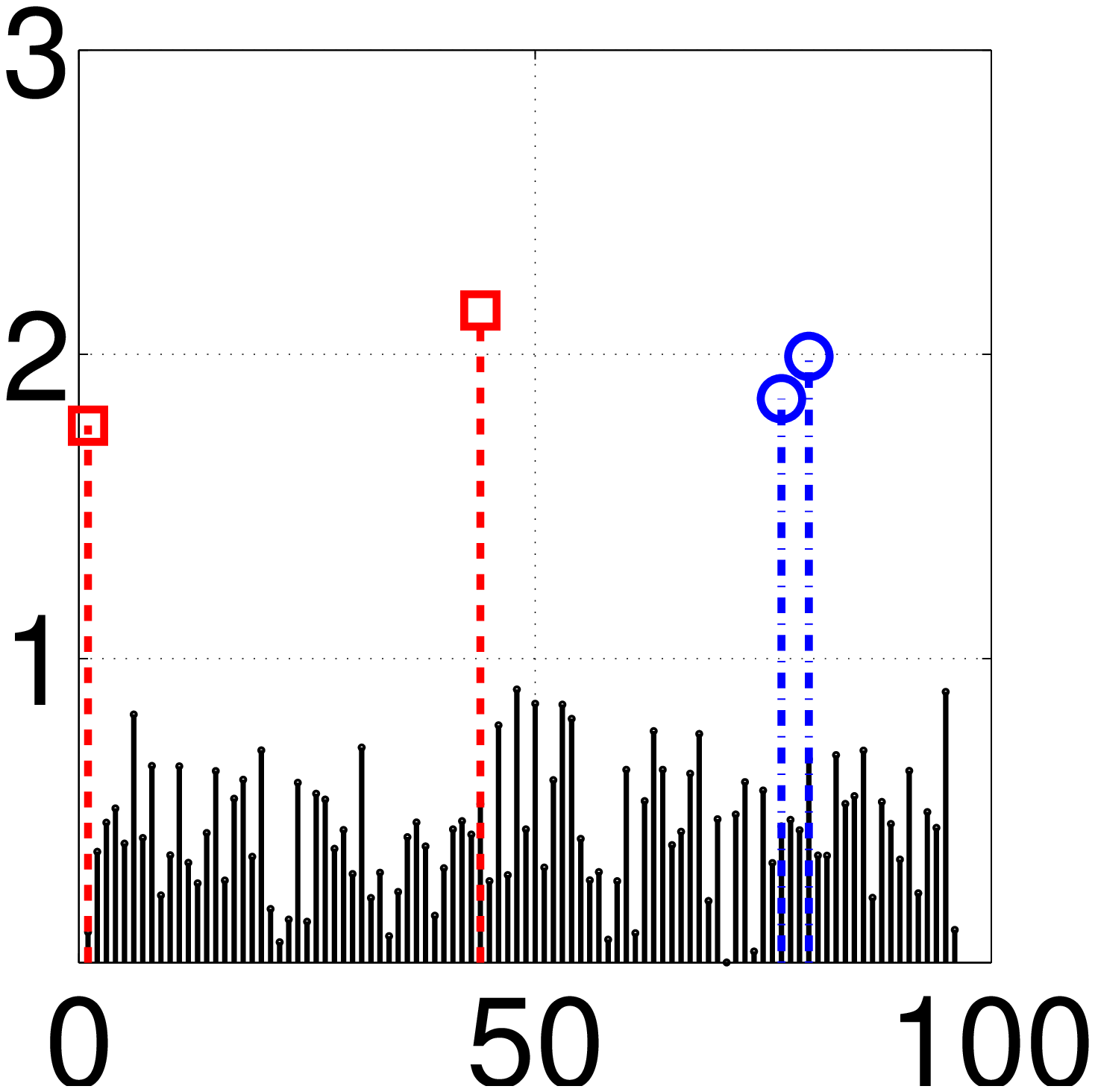}    \hspace{5mm}
\includegraphics[width= 0.15\linewidth,clip]{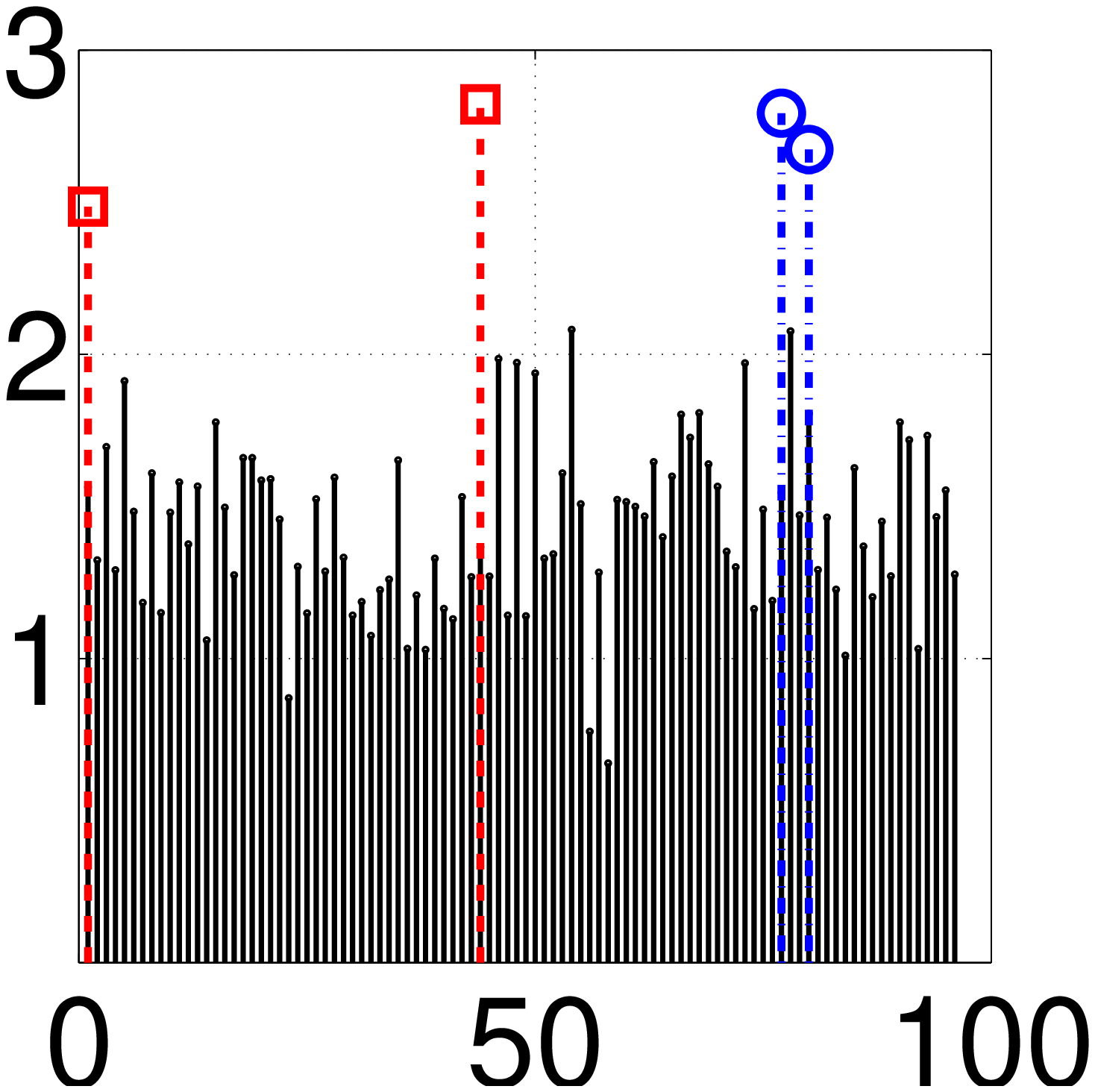}   \\
\hspace{-5mm}
(d) $\hat{V}^1$: CNJGL \hspace{10mm} 
(e) $\hatvarone$: GGL \hspace{10mm} 
(f) $\hatvarone$: GL \hspace{15mm} \\

$ $

\includegraphics[width= 0.15\linewidth,clip]{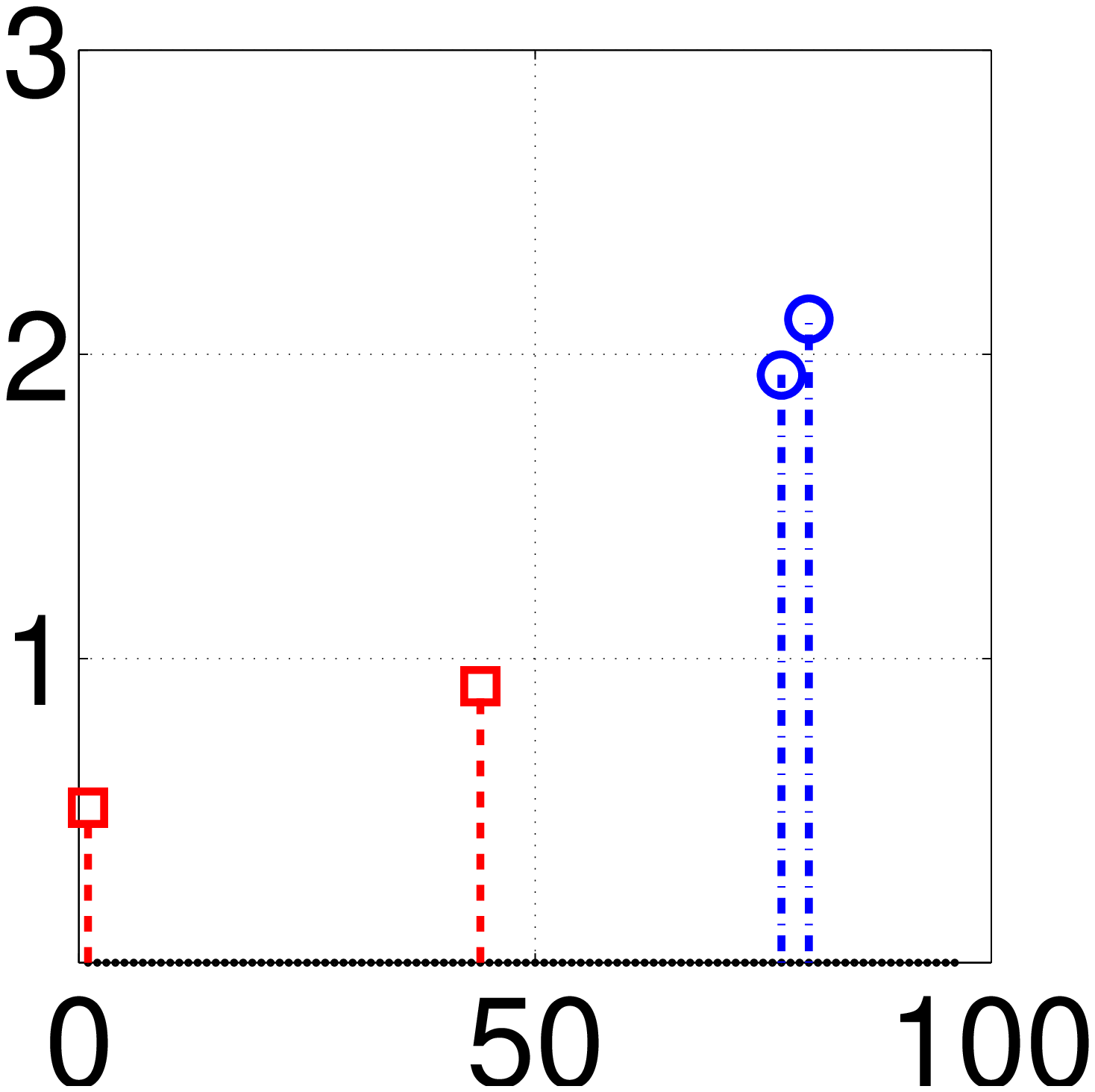}    \hspace{5mm}
\includegraphics[width= 0.15\linewidth,clip]{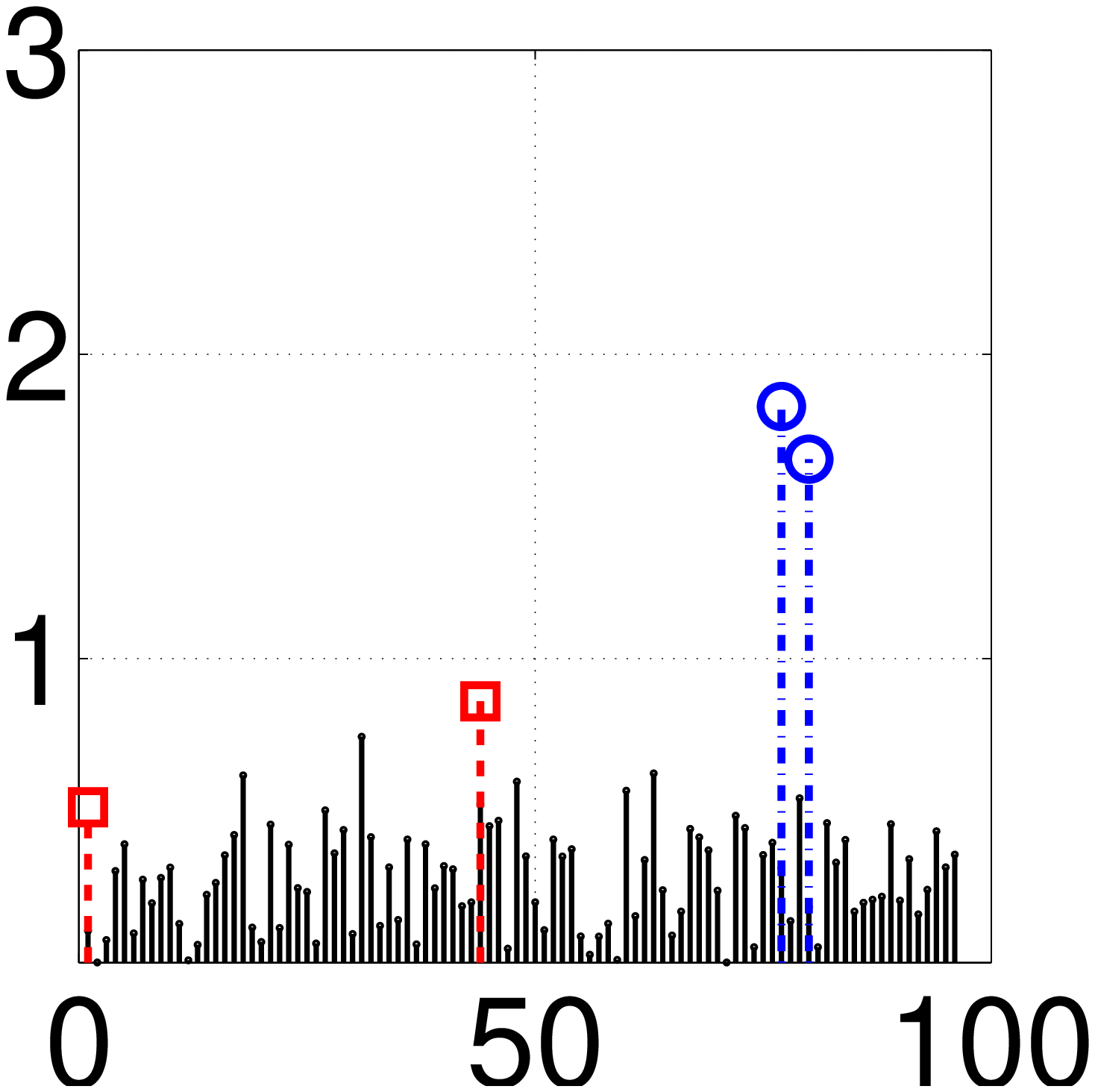}    \hspace{5mm}
\includegraphics[width= 0.15\linewidth,clip]{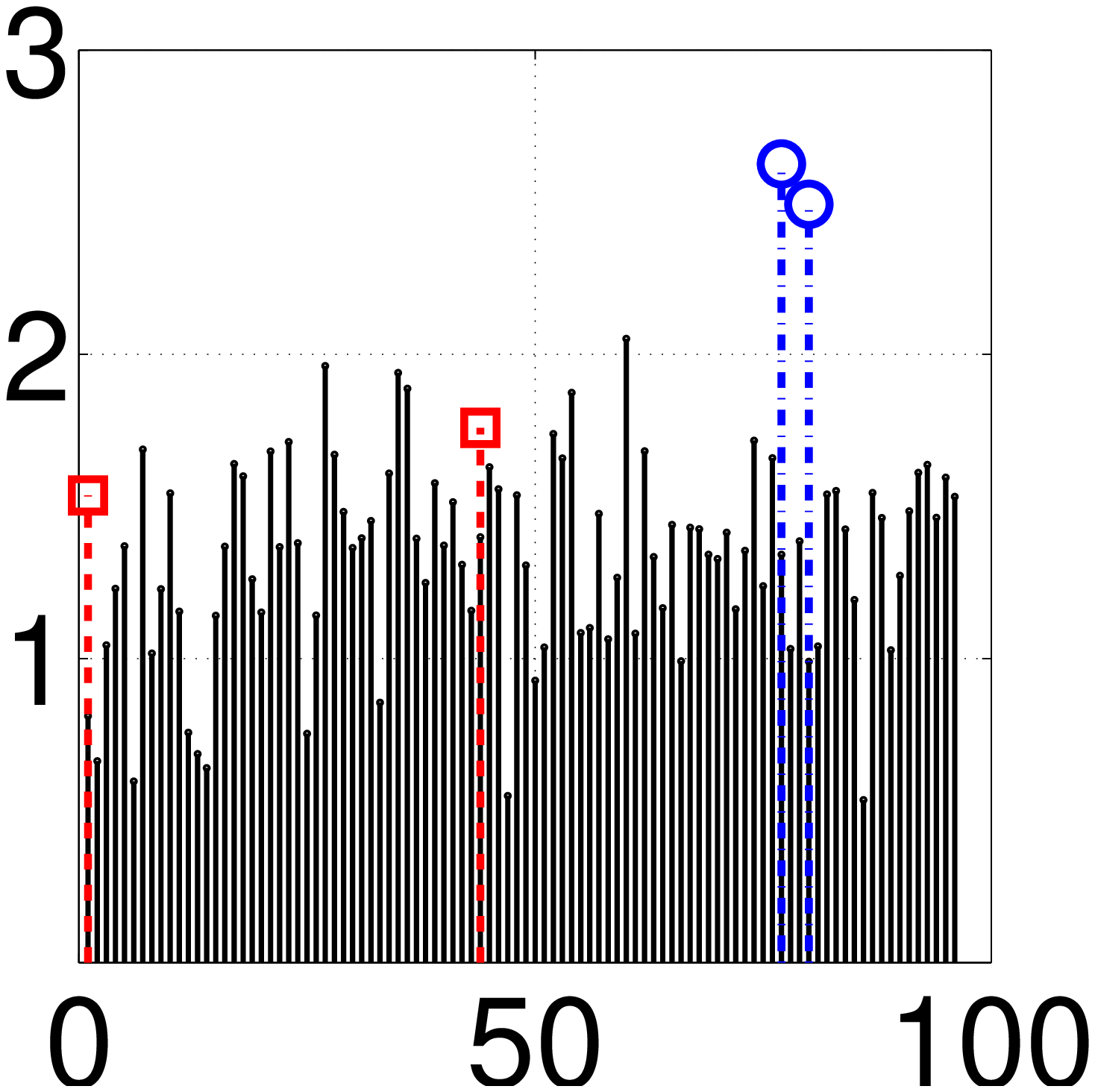}   \\
\hspace{-5mm} (g) $\hat{V}^2$: CNJGL \hspace{10mm} 
(h) $\hatvartwo$: GGL \hspace{10mm} 
(i) $\hatvartwo$: GL \hspace{15mm} \\

\fontsize{10}{12}\selectfont 
\end{center}

\caption{\label{fig:Norm_Plots} In all plots, the $x$-axis indexes the columns of the indicated matrix, and the $y$-axis displays the $\ell_2$ norms of the columns of the indicated matrix, with diagonal elements removed. The sample size is $n = 25$.   Perturbed nodes are indicated in red (with square markers), and cohub nodes are indicated in blue (with circle markers). 
\emph{(a)-(c):} Detection of perturbed nodes by PNJGL with $q=2$, FGL, and GL. \emph{(d)-(i):} Detection of cohub nodes by CNJGL with $q=2$, GGL, and GL. \emph{(a):} PNJGL with $q=2$ was performed with $\lambda_1=2.5$ and $\lambda_2=12.5$.
\emph{(b):} FGL was performed with $\lambda_1=2.5$ and $\lambda_2=0.75$. \emph{(c):} GL was performed with $\lambda=1.5$. \emph{(d), (g):} CNJGL was performed with $q=2$ and $\lambda_1=0.5$, $\lambda_2=37.5$. \emph{(e), (h):} GGL was performed with $\lambda_1=0.5$ and $\lambda_2=2.5$. \emph{(f), (i):} GL was performed with $\lambda=0.75$.}
\end{figure}

Here we present more detailed results for an instance of the simulation study described in Section \ref{sec:data}, for the case $n = 25$. 
Figure \ref{fig:Norm_Plots} illustrates how the PPC, TPPC, PCC and TPCC metrics are computed. As described in Table \ref{tbl:TableMetrics}, for PNJGL, PPC is given by  the number of columns of $\hat{ V }$ whose $\ell_2$ norms exceed the threshold $t_s$. Figure \ref{fig:Norm_Plots}(a) indicates that the two perturbed nodes in the data are identified as perturbed by PNJGL. Furthermore, given the large gap between the perturbed and non-perturbed columns, PPC is relatively insensitive to the choice of $t_s$. Similar results apply to the TPPC, PCC and TPCC metrics.

In order to generate Figure \ref{fig:Norm_Plots}, PNJGL, FGL, CNJGL, GGL, and GL were performed using tuning parameter values that led to the best identification of perturbed and cohub nodes. However, the results displayed in Figure~\ref{fig:Norm_Plots} were quite robust to the choice of tuning parameter. 

\bibliography{biblio}
\end{document}